%% file: main.tex
\pdfoutput=1

\ifdefined\useaosstyle
\documentclass[aos,preprint]{imsart}
\usepackage[numbers]{natbib}
\usepackage[colorlinks=true,citecolor=blue,allcolors=blue]{hyperref}
\usepackage{xr}
\arxiv{arXiv:1810.08750}

\else

\documentclass[11pt]{article}
\usepackage[numbers]{natbib}
\usepackage{fullpage}
\usepackage[colorlinks=true,allcolors=mydarkblue]{hyperref}
\fi

\usepackage{statistics-macros}

\ifdefined\useaosstyle
\else
\renewcommand{\red}[1]{#1}
\fi

\newcommand{\ones}{\mathbf{1}}
\newcommand{\zeros}{\mathbf{0}}

\usepackage{pgfplotstable}
\usepackage{graphicx}
\usepackage{subcaption}
\usepackage{float}

\usepackage{overpic}
\usepackage{tikz}
\usepackage{rotating}
\usepackage{psfrag}

\renewcommand{\bindic}[1]{\mathbf{I}\left({#1}\right)}

\providecommand{\comment}[1]{}

\makeatletter
\long\def\@makecaption#1#2{
  \vskip 0.8ex
  \setbox\@tempboxa\hbox{\small {\bf #1:} #2}
  \parindent 1.5em  
  \dimen0=\hsize
  \advance\dimen0 by -3em
  \ifdim \wd\@tempboxa >\dimen0
  \hbox to \hsize{
    \parindent 0em
    \hfil 
    \parbox{\dimen0}{\def\baselinestretch{0.96}\small
      {\bf #1.} #2
    } 
    \hfil}
  \else \hbox to \hsize{\hfil \box\@tempboxa \hfil}
  \fi
}
\makeatother

\begin{document}

\ifdefined\useaosstyle
\begin{frontmatter}
  \title{Learning Models with Uniform Performance via Distributionally Robust
    Optimization} \runtitle{Learning Models with Uniform Performance via DRO}

\begin{aug}
\author[A]{\fnms{John C.} \snm{Duchi} \thanksref{t1,t2}
  \ead[label=e1]{jduchi@stanford.edu}}
\and \author[B]{\fnms{Hongseok} \snm{Namkoong} \thanksref{t1,t3}
  \ead[label=e2]{hn2369@columbia.edu}}

\runauthor{J.\ C.\ Duchi and H.\ Namkoong}

\affiliation{Stanford University and Columbia University}

\address[A]{Departments of Statistics and Electrical Engineering,
  Stanford University,
  \href{mailto:jduchi@stanford.edu}{jduchi@stanford.edu}
}

\address[B]{Decision, Risk, and Operations Division,
  Columbia Business School,
  \href{mailto:hn2369@columbia.edu}{hn2369@columbia.edu}
}

\thankstext{t1}{Supported by SAIL-Toyota Center for AI Research.}
\thankstext{t2}{Supported by National Science Foundation award
  NSF-CAREER-1553086 and Office of Naval Research YIP award
  N00014-19-2288.}
\thankstext{t3}{Supported by Samsung Fellowship.}
\end{aug}

\else
\begin{center}
  {\LARGE Learning Models with Uniform Performance \\via Distributionally Robust
    Optimization} \\
  \vspace{.5cm}
  {\Large John C.\ Duchi$^1$ ~~~~ Hongseok Namkoong$^2$} \\
  \vspace{.2cm}
  {\large Departments of Statistics and
  Electrical Engineering, Stanford University$^1$} \\
  \vspace{.1cm}
  {\large Decision, Risk, and Operations Division, Columbia Business School$^2$} \\
  \vspace{.2cm}
  \texttt{jduchi@stanford.edu, hn2369@columbia.edu}
\end{center}
\fi 
\begin{abstract}
  A common goal in statistics and machine learning is to learn models that
  can perform well against distributional shifts, such as latent
  heterogeneous subpopulations, unknown covariate shifts, or unmodeled
  temporal effects.  We develop and analyze a distributionally robust
  stochastic optimization (DRO) framework that learns a model providing
  good performance against perturbations to the data-generating
  distribution.  We give a convex formulation for the problem,
  providing several convergence guarantees. We prove finite-sample minimax
  upper and lower bounds, showing that distributional robustness sometimes
  comes at a cost in convergence rates. We give limit theorems for the
  learned parameters, where we fully specify the limiting distribution so
  that confidence intervals can be computed.  On real tasks including
  generalizing to unknown subpopulations, fine-grained recognition, and
  providing good tail performance, the distributionally robust approach
  often exhibits improved performance.
\end{abstract}

\ifdefined\useaosstyle

\end{frontmatter}
\fi

\renewcommand{\theassumption}{\Alph{assumption}}


\input{introduction}
\input{formulation}

\input{experiments}
\input{upper}
\input{lower}
\input{asymptotics}

\input{asymptotics-limit-lam}
\input{discussion}

\ifdefined\useaosstyle
\else
\subsection*{Acknowledgments}

JCD and HN were partially supported by the SAIL-Toyota Center for AI
Research and HN was partially supported Samsung Fellowship. JCD was also
partially supported by the National Science Foundation award
NSF-CAREER-1553086 and the Office of Naval Research Young Investigator award
N00014-19-2288.
\fi

\ifdefined\useaosstyle




\begin{supplement}[id=suppA]
  \stitle{Proofs of Results}
  \slink[doi]{COMPLETED BY THE TYPESETTER}
  \sdatatype{.pdf}
  \sdescription{The supplementary material contains proofs of our results}
\end{supplement}

\bibliographystyle{imsart-nameyear}
\bibliography{bib}

\end{document}

\else

\bibliographystyle{abbrvnat}
\setlength{\bibsep}{.2em}
\bibliography{bib}


\appendix

\input{formulation-proof}

\input{upper-proof}

\input{lower-proof}

\input{consistency-proof}

\input{asymptotics-proof-lam}

\end{document}

\fi


%% file: introduction.tex
\section{Introduction}
\label{section:introduction}

In many applications of statistics and machine learning, we wish to learn
models that achieve uniformly good performance over almost all input values.
This is important for safety- and fairness-critical systems such as medical
diagnosis, autonomous vehicles, criminal justice and credit evaluations, where
poor performance on the tails of the inputs leads to high-cost system
failures. Methods that optimize average performance, however, often produce
models that suffer low performance on the ``hard'' instances of the
population. For example, standard regressors obtained from maximum likelihood
estimation can lose predictive power on certain regions of
covariates~\cite{MeinshausenBu15}, and high average performance comes at the
expense of low performance on minority subpopulations. In this work, we
study a procedure that explicitly optimizes performance on tail inputs that
suffer high loss.


Modern datasets incorporate 
heterogeneous (but often latent) subpopulations,
and a natural goal is to perform well across all
of these~\cite{MeinshausenBu15, RothenhauslerMeBu16,
  BuhlmannMe16}. While many statistical models show strong average
performance, their performance often deteriorates on minority
groups underrepresented in the dataset. For example, speech
recognition systems are inaccurate for people with minority
accents~\cite{AmodeiAnAnBaBaCaCaCaChCh16}. In numerous other
applications---such as facial recognition, automatic video captioning,
language identification, academic recommender systems---performance
varies significantly over different demographic
groupings, such as race, gender, or age~\cite{GrotherQuPh10, HovySo15,
  BlodgettGrOc16, SapiezynskiKaWi17, Tatman17}.

In addition to latent heterogeneity in the population, distributional shifts
in covariates~\cite{Shimodaira00, Ben-DavidBlCrPe07} or unobserved confounding
variables (e.g.\ unmodeled temporal effects~\cite{Hand06}) can contribute to
changes in the data generating distribution.
Performance of machine learning models degrades significantly on domains
that are different from what the model was trained on~\cite{Hand06,
  BlitzerMcPe06, DaumeMa06, SaenkoKuFrDa10, TorralbaEf11} and
even when new test data are constructed following identical data
construction procedures~\cite{RechtRoScSh19}. Domain
adaptation~\cite{Shimodaira00, Ben-DavidBlCrPe07, Ben-DavidBlCrPeVa10} and
multi-task learning methods~\cite{Caruana98} can be effective in
situations where (potentially unlabeled) data points from the target domain
are available. The reliance on \emph{a priori} fixed target domains,
however, is restrictive, as the shifted target distributions are usually
unknown before test time and it is impossible to collect
data from the targets.

To mitigate these challenges, we
consider unknown distributional shifts,
developing and analyzing a loss minimization framework that is
explicitly robust to local changes in the data-generating
distribution. Concretely, let $\Theta \subseteq \R^d$ be the parameter (model)
space, $P_0$ be the data generating distribution on the measure space
$(\statdomain, \mc{A})$, $X$ be a random element of $\statdomain$, and
$\loss: \paramdomain \times \statdomain \to \R$ be a loss function. Rather
than minimizing the average loss
$\E_{P_0}[\loss(\theta; X)]$, we study the \emph{distributionally
  robust}  problem
\begin{equation}
  \label{eqn:objective}
  \minimize_{\param \in \Theta} ~
  \left\{
    \risk_f(\theta; P_0) \defeq
    \sup_{Q \ll P_0} \left\{
    \E_Q[\loss(\param; \statrv)]
    : \fdiv{Q}{P_0} \le \tol
  \right\}
 \right\},
\end{equation}
where the hyperparameter $\rho > 0$ modulates the distributional
shift. Here
\begin{equation*}
  \fdiv{Q}{P_0} \defeq
  \int f\left(\frac{dQ}{dP_0}\right)dP_0
\end{equation*}
is the $f$-divergence~\cite{AliSi66,Csiszar67} between $Q$ and $P_0$, where
$f : \R \to \overline{\R}_+ = \R_+ \cup \{\infty\}$ is a convex function
satisfying $f(1) = 0$ and $f(t) = +\infty$ for any $t < 0$.


The worst-case risk~\eqref{eqn:objective} upweights regions of $\mc{X}$ with
high losses $\loss(\param; \statrv)$, and thus
formulation~\eqref{eqn:objective} optimizes performance on the tails, as
measured by the loss on ``hard'' examples. In our motivating scenarios of
distribution shift or latent subpopulations, \red{as long as the alternative
  distribution $Q$ remains $\tol$-close to the data-generating distribution
  $P_0$,} the model $\theta\opt \in \Theta$ that minimizes the worst-case
formulation~\eqref{eqn:objective} evidently guarantees that
$\E_Q[\loss(\theta\opt; \statrv)] \le \risk_f(\theta\opt; P_0)$ and provides the
smallest such bound; as we show shortly, this is equivalent to controlling
the tail-performance under $P_0$. \red{In our subsequent discussion, we refer to
this behavior as \emph{uniform performance}.}
Letting $\emp$ denote the
empirical measure on $X_i \simiid P_0$, our approach to minimizing
objective~\eqref{eqn:objective} is via the plug-in estimator
\begin{equation}
  \label{eqn:plug-in}
  \what{\theta}_n
  \in \argmin_{\theta \in \Theta}
  \left\{
    \risk_f(\theta; \emp) \defeq
    \sup_{Q \ll \emp} \left\{
      \E_Q[\loss(\param; \statrv)]
      : \fdivs{Q}{\emp} \le \tol
    \right\}
  \right\}.
\end{equation}




To build intuition for the worst-case formulation~\eqref{eqn:objective}, we
begin our discussion (in Section~\ref{section:formulation}) by showing that
protection against distributional shifts is equivalent to controlling the
tail-performance of a model. The modeler's choice of $f$ determines the tail
performance she wants to control, and this dual interpretation provides
intuition for the appropriate choice of $f$ and $\rho$.  To concretely
understand the types of distributional shifts the worst-case
formulation~\eqref{eqn:objective} protects against, we provide (in
Section~\ref{subsection:formulation-examples}) explicit calculations
suggesting appropriate choices of $f$ in some situations. Given nontrivial
modeling freedom in choosing $f$ and $\rho$, we begin our study in
Section~\ref{section:experiments} with experiments that substantiate our
intuitive explanations.  \red{Our experimental and theoretical work
  demonstrates that the distributionally robust estimator $\what{\theta}_n$
  trades performance on the tails of the data-generating distribution with
  average-case performance---which empirical risk minimization
  optimizes. Empirically, we observe in a number of scenarios that such gains
  in tail-performance (e.g.\ hard inputs) come at moderate degradation to the
  average-case performance, so that the robust estimator~\eqref{eqn:plug-in}
  achieves fairly low loss uniformly across the input space $\mc{X}$.  For non
  worst-case distribution shifts, the worst-case
  formulation~\eqref{eqn:objective} \emph{prima-facie} does not guarantee
  better performance than empirical risk minimization; the duality between it
  and tail losses to come suggests that for light-tailed data, distributional
  robustness comes at little cost to typical-case performance.  While work in
  finance and operations research~\cite{BertsimasGuKa18} highlights the
  benefits of robustness, it is important to investigate the typical shifts
  one might expect in statistical learning scenarios.  To this end, we see
  in our experiments that the robust estimator~\eqref{eqn:plug-in} sacrifices
  some average-case performance (which empirical risk minimization optimizes)
  for lower losses on difficult subpopulations, covariate shift, and other
  latent confounding.  }


Although we view a general theoretical characterization of the ``right''
choice of $f$ and $\rho$ as an important open question, we provide two
heuristics for this choice and evaluate their performance on simulation
experiments in Section~\ref{section:experiments}. First, as a general
approach, we \red{advocate splitting training data non-exchangeably into
  multiple validation sets}, then using these to validate choices $f$ and
$\rho$; we will expand on this later in the paper with concrete examples and
experiments. As brief examples, we may group data by its loss or, in
supervised learning scenarios with outcome/label $Y$, by values of $Y$; when
an auxiliary dataset on worse-than-average subpopulations is available, we
could use this.  The intuition is to use variability within the available data
as a proxy for potential departures from the data-generating distribution.

Motivated by our empirical findings in Section~\ref{section:experiments}, the
main theoretical component of this work is to study finite sample and
asymptotic properties of the plug-in estimator~\eqref{eqn:plug-in}. We first
provide an efficiently minimizable (finite-dimensional) dual formulation which
also forms the basis of our above tail-performance interpretation of
distributional robustness (Section~\ref{section:formulation}). We give
convergence guarantees for the plug-in estimator~\eqref{eqn:plug-in}
(Section~\ref{section:upper}), and prove that it is rate optimal
(Section~\ref{section:lower}), thereby providing finite-sample minimax bounds
on the optimization problem~\eqref{eqn:objective}.  Because the
formulation~\eqref{eqn:objective} protects against gross departures from the
average loss, we observe a degradation in minimax convergence rates that is
effectively a consequence of needing to estimate high moments of random
variables.  More quantitatively, our convergence guarantees show that for
$f$-divergences with $f(t) \asymp t^k$ as $t \to \infty$, where
$k \in (1, \infty)$, the empirical minimizer $\what{\theta}_n$ satisfies
\begin{align*}
  \risk_f(\what{\theta}_n; P_0)
  -\inf_{\theta \in \Theta} \risk_f(\theta; P_0)
  =
  O_P \left( n^{-\frac{1}{k_* \vee 2}} \log n\right),
\end{align*}
where $k_* = \frac{k}{k-1}$ (Section~\ref{section:upper}). We provide minimax
lower bounds matching these rates in $n$ up to log factors. These
results quantify fundamental \emph{statistical costs} for protecting against
large distributional shifts (the worst-case region
$\{Q: \fdiv{Q}{P_0} \le \rho\}$ becomes larger as $k \to 1$, or
$k_* \to \infty$).

Since these minimax guarantees do not necessarily reflect the
typical behavior of the estimators, we complete our theoretical analysis in
Section~\ref{section:asymptotics} with an asymptotic analysis.  The estimator
$\what{\theta}_n$ is consistent under mild (and standard) regularity
conditions (Section~\ref{section:consistency}). Under suitable
differentiability conditions on $\risk_f$, $\what{\theta}_n$ is asymptotically
normal at the typical $\sqrt{n}$-rate, allowing us to obtain calibrated
confidence intervals (Section~\ref{section:limit-theorems}).




\ifdefined\useaosstyle
\subsection*{Related Work}
\else
\paragraph{Related Work}
\fi


Distributional shift arise in many guises across statistics, machine
learning, applied probability, simulation, and optimization; we give a
necessarily abridged survey of the many strains of work and their respective
foci.  Work in domain adaptation seeks models that receive data from one
domain and are tested on a specified target; typical approach is to reweight
the distribution $P_0$ to make it ``closer'' to the known target distribution
$P_{\rm target}$~\cite{Shimodaira00, HuangGrBoKaScSm07, BickelBrSc07,
  SugiyamaKrMu07, SugiyamaNaKaBuKa08, TsuboiKaHiBiSu09}. In this vein, one
interpretation of the worst-case formulation~\eqref{eqn:objective} is as
importance-weighted loss minimization without a known target domain---that is,
without assuming even unlabeled data from the target domain. The
formulation~\eqref{eqn:objective} is more conservative than most domain
adaptation methods, as it considers shifts in the joint distribution of
predictors $X$ and target variable $Y$ instead of covariate shifts.

Other scenarios naturally give rise to structural distributional
changes. Time-varying effects are a frequent culprit~\cite{Hand06}, and
time-varying-coefficient models are effective when time indices are
available~\cite{FanZh99, CaiFaLi00}.  When one believes there may be latent
subpopulations, mixture model approaches can model latent membership
directly~\cite{AitkinRu85, FigueiredoJa02, McLachlanPe04, CappeMoRy05}.  In
contrast, our worst-case approach~\eqref{eqn:objective} does not directly
represent (or require) such latent information, and---especially in the case
of mixture models---can maintain convexity because of the focus on uniform
performance guarantees.

When we know and can identify heterogeneous populations within the data,
B\"{u}hlmann, Meinshausen, and colleagues connect methods that achieve good
performance on all subpopulations with causal interventions. In this vein,
they study maximin effects on heterogeneous datasets and learn linear models
that maximize relative performance over the worst (observed)
subgroup~\cite{MeinshausenBu15}, which connects to minimax regret in
linear models~\cite{EldarBeNe04, Ben-TalGhNe09, RothenhauslerMeBu16,
  BuhlmannMe16, RothenhauslerBuMePe18}.
Without access to information about particular subpopulations, the
worst-case formulation~\eqref{eqn:objective} is more conservative than
their approaches, but can still achieve good performance, as we see in our
experimental evaluation.

The idea to build predictors robust to perturbation of an underlying
data-generating distribution has a long history across multiple fields.  In
dynamical systems and control, \citet{PetersenJaDu00} build worst-case optimal
controllers for systems whose uncertain dynamics are described by
Kullback-Leibler (KL) divergence balls. In econometrics, \citet{HansenSa08}
study systems in which rational agents dynamically make decisions assuming
worst-case (dynamics) model misspecification, where the misspecification is
bounded by an evolving KL-divergence quantity.  There is also substantial work
in characterizing worst-case sensitivity of risk measures to distributional
misspecification~\cite{GlassermanXu13, AtarChDu15, Lam16, DupuisKaPaPl16,
  Lam17, GhoshLa19}. A common goal in such sensitivity calculations is an
asymptotic expansion of a risk measure as the radius $\tol$ of the region of
misspecification decreases to 0.  In contrast, we study statistical properties
of the worst-case formulation~\eqref{eqn:objective} given observations drawn
from the data generating distribution $P_0$, so that we must both address
statistical uncertainty and challenges of robustness.


In the optimization literature, a body of work studies distributionally robust
optimization problems. Several authors investigate worst-case regions arising
out of moment conditions on the data vector $X$~\cite{DelageYe10, JiangGu16,
  BertsimasGuKa18}.  Other work~\cite{Ben-TalHeWaMeRe13, BertsimasGuKa18,
  DuchiGlNa16, NamkoongDu17, Lam16, LamZh17} studies a scenario similar to our
$f$-divergence formulation~\eqref{eqn:objective}. In this line of research,
the empirical plug-in procedure~\eqref{eqn:plug-in} with radius $\tol / n$
provides a finite sample confidence set for the \emph{population objective}
$\E_{P_0}[\loss(\theta; X)]$; the focus there is on the true distribution
$P_0$ and does not consider distributional shifts. \citet{DuchiGlNa16}
and~\citet{LamZh17} show how such approximations correspond to generalized
empirical likelihood~\cite{Owen90} confidence bounds on
$\E_{P_0}[\loss(\theta; X)]$. These procedures are identical to the
plug-in~\eqref{eqn:plug-in} except that the radius decreases as $\tol/n$.
Thus, the magnitude of this radius depends on whether the modeler's goal is
good performance with respect to $\E_{P_0}[\loss(\theta; X)]$ (radius shrinks
as $\rho / n)$, or---as is the case here---robustness under distributional
shifts (radius $\tol$ is fixed).

An alternative to our $f$-divergence based sets
$\{Q : \fdivs{Q}{P_0} \le \tol\}$ are Wasserstein
balls~\cite{Wald45, PflugWo07, Wozabal12, Shafieezadeh-AbadehEsKu15,
  BlanchetMu19, BlanchetKaMu19, GaoKl16, EsfahaniKu18, SinhaNaDu17, LeeRa17}.
Such approaches are satisfying, as Wasserstein balls allow worst-case
distributions with different support from the data-generating distribution
$P_0$. This power, however, means that tractable reformulations are only
available under restrictive scenarios~\cite{Shafieezadeh-AbadehEsKu15,
  EsfahaniKu18, SinhaNaDu17}, and they remain computationally challenging.
Furthermore, most guarantees~\cite{BlanchetKaMu19, EsfahaniKu18,
  Shafieezadeh-AbadehEsKu15} for these problems also consider approximation
only of the canonical (population) loss $\E_{P_0}[\loss(\theta; X)]$ using
shrinking radius $\tol_n \to 0$. In comparison, our $f$-divergence formulation is
computationally efficient to solve, even in large-scale learning
scenarios~\cite{NamkoongDu16, NamkoongDu17}.

\ifdefined\useaosstyle
\subsection*{Notation}
\else
\paragraph{Notation}
\fi

For a sequence of random variables $Z_1, Z_2, \ldots$ in a metric space
$\mathcal{Z}$, we say $Z_n \cd Z$ if $\E[h(Z_n)] \to \E[h(Z)]$ for all bounded
continuous functions $h$, and $Z_n \cp Z$ for convergence in probability.
We let $\ell^{\infty}(\mc{Z})$ the space of bounded
real-valued functions on $\mc{Z}$ equipped with the supremum norm.  We let
$\dchi{P}{Q} = \half \int \left(dP/dQ - 1\right)^2 dQ$ be the
$\chi^2$-divergence. For $Z \sim P$,
$\esssup_{P} Z$ is its essential supremum. We make the dependence on the
underlying measure explicit when we write expectations (e.g. $\E_P[X]$),
except for when $P = P_0$. For $k \in (1, \infty)$, we
let $k_* \defeq k / (k-1)$.  
By $\nabla \loss(\theta; X)$ we mean
differentiation with respect to the parameter vector $\theta \in \R^d$.



%% file: formulation.tex

\section{Formulation}
\label{section:formulation}




We begin our discussion by presenting dual reformulations for the worst-case
objective $\risk_f(\theta; P_0)$, deferring formulation in terms of
worst subpopulations to Example~\ref{example:cvar} to come.
The dual form gives a single convex
minimization problem for computing the empirical plug-in
estimator~\eqref{eqn:plug-in} in place of the minimax formulation, and it
makes explicit the role that $t \mapsto f(t)$ plays in defining such a
\emph{risk-averse} version of the usual average loss
$\E_{P_0}[\loss(\theta; X)]$.  This provides an equivalence
between distributional robustness and tail-performance, which we draw on
subsequently both statistical and computational reasons. Defining the
\emph{uncertainty region}
\begin{equation*}
  \uncertainty_P \defeq \left\{Q : \fdivs{Q}{P} \le \tol\right\},
\end{equation*}
we may use the likelihood ratio
$\likerat(\statval) \defeq dQ(\statval) / dP_0(\statval)$ to reformulate our
distributionally robust problem~\eqref{eqn:objective} via
\begin{align}
  \nonumber
  \risk_f(\theta; P_0)
  & =
  \sup_P\left\{\E_P[\loss(\param; \statrv)] : P \in \uncertainty_{P_0}
  \right\} \\
  & = \sup_{\likerat \ge 0} \left\{
  \E_{P_0}[\likerat(\statrv) \loss(\param; \statrv) ]
  \mid \E_{P_0}[f(\likerat(\statrv))]
  \le \tol,
  \E_{P_0}[\likerat(\statrv)] = 1
  \right\},
  \label{eqn:sup-ratio-version}
\end{align}
where the supremum is over measurable functions.
We now recall \citet{Ben-TalHeWaMeRe13} and
\citeauthor{Shapiro17}'s dual reformulation of the
quantity~\eqref{eqn:sup-ratio-version}, where
$f^*(s) \defeq \sup_t \{st - f(t) \}$ is
the usual Fenchel conjugate.
\begin{proposition}[{\citet[Section 3.2]{Shapiro17}}]
  \label{proposition:duality}
  Let $P$ be a
  probability measure on $(\statdomain, \sigalg)$ and $\tol > 0$.
  Then
  \begin{align}
    \label{eqn:dual}
    \risk_f(\theta; P)
    =  \inf_{\lambda \ge 0, \eta \in \R}
    \left\{
    \E_P \left[ \lambda f^*\left(\frac{\loss(\param; X) - \eta}{\lambda}\right)
    \right] + \lambda \tol + \eta \right\}
  \end{align}
  for all $\theta$.
  Moreover, if the supremum on the left hand side
  is finite, there are finite $\lambda(\param) \ge 0$ and
  $\eta(\param) \in \R$ attaining the infimum on the right hand side.
\end{proposition}
\noindent For convex losses $\theta \mapsto \loss(\theta; X)$, the dual
form~\eqref{eqn:dual} is jointly convex in $(\theta, \eta, \lambda)$. While
interior point methods~\cite{BoydVa04} are powerful tools for solving such
problems, they may be slow in settings where $n$, the sample size, and $d$,
the dimension of $\theta \in \Theta$, are large. More direct methods can
directly solve the primal form, including gradient descent or stochastic
gradient algorithms~\cite{NamkoongDu16, NamkoongDu17}.

\ifdefined\useaosstyle
\subsection*{Divergence families}
\else
\vspace{-10pt}
\paragraph{Divergence families}
\fi

Much of our development centers on two families of divergences. The
\emph{R\'{e}nyi $\alpha$-divergence}~\cite{ErvenHa14} between distributions
$P$ and $Q$ is
\begin{equation}
  \label{eqn:renyi-def}
  D_\alpha(P |\!| Q)
  \defeq \frac{1}{\alpha - 1} \log \int \left(\frac{dP}{dQ}\right)^\alpha dQ,
\end{equation}
where the limit as $\alpha \to 1$ satisfies $D_1(P |\!| Q) =
\dkl{P}{Q}$. For analytical reasons, we use the equivalent
Cressie-Read family of $f$-divergences~\cite{CressieRe84}. These are
parameterized by $k \in (-\infty, \infty) \setminus \{0, 1\}$,
$k_* = \frac{k}{k-1}$, with
\begin{equation}
  \label{eqn:cressie-read}
  f_k(t) \defeq \frac{t^{k} - kt + k - 1}{k(k-1)}
  ~~~ \mbox{so} ~~~
  f_k^*(s) \defeq \frac{1}{k}\left[ \hinge{(k-1)s + 1}^{k_*} - 1\right].
\end{equation}
We let $f_k(t) = +\infty$ for $t < 0$, and we
define $f_1$ and $f_0$ as their respective limits as $k \to 0, 1$. The
family of divergences~\eqref{eqn:cressie-read} includes $\chi^2$-divergence
($k=2$), empirical likelihood $f_0(t) = -\log t + t - 1$, and KL-divergence
$f_1(t) = t \log t - t + 1$, and we frequently use the shorthand
\begin{equation}
  \label{eqn:cr-risk}
  \risk_k(\theta; P)
  \defeq \sup_{Q \ll P} \left\{
    \E_Q[\loss(\param; \statrv)]
    : D_{f_k}\left( Q |\!| P\right) \le \tol
  \right\}.
\end{equation}

While most of our results generalize to other values of $k$, we focus
temporarily on $k \in (1, \infty)$ for ease of exposition (only our
finite-sample guarantees in Section~\ref{section:upper} require
$k \in (1, \infty)$).  By minimizing out $\lambda \ge 0$ in the dual
form~\eqref{eqn:dual}, we obtain a simplified formulation for the
Cressie-Read family~\eqref{eqn:cressie-read}.
\begin{lemma}
  \label{lemma:cressie-read-risk}
  For any probability $P$ on $(\statdomain, \sigalg)$,
  $k \in (1, \infty)$, $k_* = k / (k-1)$, any
  $\tol > 0$, and $c_k(\tol) \defeq (1 + k(k - 1) \tol)^\frac{1}{k}$,
  we have for all $\param \in \Theta$
  \begin{equation}
    \risk_k(\theta; P) = \inf_{\eta \in \R}
    \left\{ 
    c_k(\tol)
    \E_P\left[\hinge{\loss(\theta; X) - \eta}^{k_*}\right]^\frac{1}{k_*}
      + \eta \right\}.
    \label{eqn:cressie-read-risk}
  \end{equation}
\end{lemma}
\noindent See Section~\ref{section:proof-of-cressie-read-risk} for the
proof. The simplified dual form~\eqref{eqn:cressie-read-risk} shows that
protecting against worst-case distributional shifts is equivalent to
optimizing the tail-performance of a model; the worst-case objective
$\risk_k(\theta; P)$ only penalizes losses above the optimal dual
variable $\eta\opt(\theta)$.
The $L^{k_*}(P)$-norm upweights these tail values of $\loss(\theta; x)$,
giving a worst-case objective that focuses on ``hard'' regions of
$\statdomain$. Eq.~\eqref{eqn:cressie-read-risk} also makes explicit the
relationship between the growth $f_k$ and the worst-case objective
$\risk_k(\theta; P)$: as growth of $f_k(t)$ for large $t$ becomes steeper
($k \uparrow \infty$), the $f$-divergence ball
$\{Q: D_{f_k}\left(Q |\!| P \right) \le \rho\}$ shrinks, and the risk measure
$\risk_k(\theta; P)$ becomes less conservative (smaller). Since the dual
form~\eqref{eqn:cressie-read-risk} quantifies this with the $L^{k_*}(P)$-norm
of the loss above the quantile $\eta$, we see that $f_k$ with
$k \in (1, \infty)$ is a possible choice if the loss has finite $k_*$-moments
under the nominal distribution $P_0$. In contrast, the worst-case
formulation~\eqref{eqn:objective} corresponding to the KL-divergence ($k = 1$)
is finite only when the moment generating function of the loss
exists~\cite{Ahmadi12}.\footnote{This correspondence
  between higher moments and divergences holds in more generality in
  that if $f(t)$ grows asymptotically as $t^k$ as $t \to \infty$, then
  the dual exhibits similar $k_*$th moment behavior; see
  supplementary Appendix~\ref{sec:moments-duality}.}

An extensive literature on coherent risk measures defines utility functions
that exhibit ``sensible'' tail risk preference~\cite{ArtznerDeEbHe99,
  RockafellarUr00, Krokhmal07, ShapiroDeRu09}; there is a duality between
distributionally robust optimization and coherent risk
measures~\cite[e.g.][Thm.~6.4]{ShapiroDeRu09}.  In this sense the
distributionally robust problem~\eqref{eqn:objective} is a risk-averse
formulation of the canonical stochastic optimization problem of minimizing
$\E_{P_0}[\loss(\theta; X)]$.  Indeed, \citet{Krokhmal07} proposes the dual
form~\eqref{eqn:cressie-read-risk} as a higher order generalization of the
classical conditional value-at-risk~\cite{RockafellarUr00}, which
corresponds to $\risk_k(\theta; P)$ defined with $k = \infty$ (or $k_* = 1$)
in our notation.


\subsection{Examples}
\label{subsection:formulation-examples}

While---as we note in the introduction---we do not provide precise
recommendations for the choice of $f$-divergence, it is instructive to
consider a few examples for motivation and to connect to our worst-case
subpopulation considerations
(Examples~\ref{example:cvar}--\ref{example:mixture-linear-regression}).  We
begin with a generic description and specialize subsequently, deferring
heuristic procedures for choosing $f$ and $\rho$ (and empirical efficacy
evaluations) to the next section.

\begin{example}[Generic distributional shift]
  \label{example:generic-shift}
  Consider data in pairs $(X, Y)$, where $X$ is a feature (covariate) vector
  and $Y$ is a dependent variable (e.g.\ label) we wish to model from $X$. Let
  $U$ be a latent (unobserved) confounding variable, and assume that the pair
  $(X, Y)$ jointly follows $P_0(\cdot \mid U = u)$.  For a marginal
  distribution $\mu$ on $U$, let
  $P_\mu((X, Y) \in A) \defeq \int P_0((X, Y) \in A \mid U = u) d\mu(u)$.  We
  have the essentially tautological correspondence
  \begin{align*}
    \left\{P \mid \fdiv{P}{P_0} \le \tol\right\}
    & = \left\{P_\mu \mid
    \int f\left(\frac{dP_\mu(x, y)}{dP_0(x, y)}\right) dP_0(x, y)
    \le \tol \right\}.
  \end{align*}
  The robustness set is a family of distributional interventions on $U$.  We
  leave characterizing the precise form of such interventions as an open
  question.
\end{example}


For well-specified linear models, it is frequently the case that the robust
parameter $\theta\dro \in \argmin_\theta \risk_f(\theta; P)$ minimizing the
objective~\eqref{eqn:objective} coincides with the true parameter, though its
plug-in estimator may be less efficient than standard ordinary least-squares
estimators (we do not discuss this efficiency here).

\begin{example}[Regression and stochastic domination]
  \label{example:regression-domination}
  To make things precise, recall stochastic orders~\cite{ShakedSh07}: for two
  $\R$-valued random variables $U$ and $V$, we say that $V$ stochastically
  dominates $U$ if $\P(U \ge t) \le \P(V \ge t)$ for all $t \in \R$, written
  $U \preceq V$; this is equivalent to the condition that
  $\E[g(U)] \le \E[g(V)]$ for all nondecreasing $g$.  For any problem with
  data in pairs $(X, Y)$ and a loss $\loss(\theta; X, Y)$, if there exists a
  parameter $\theta\subopt$ such that
  $\loss(\theta\subopt; X, Y) \preceq \loss(\theta; X, Y)$ for all $\theta$,
  we then have $\theta\subopt \in \argmin_\theta \risk_f(\theta; P)$ for all
  $f$-divergences, as $\risk_f$ is a coherent risk
  measure~\cite[cf.][Ch.~6.3]{ShapiroDeRu09}.  Existence of such
  $\theta\subopt$ is a strong condition, but holds in a few important cases.

  For concreteness consider linear regression,
  where $(x, y) \in \R^d \times \R$ and $\loss(\theta;
  x, y) = \half (\theta^T x - y)^2$. First, we consider
  the case that the model is well-specified, so that $Y = X^T \theta\subopt
  + \noise$, where $\E[\noise \mid X] = 0$.  If the distribution of
  $\noise$ given $X = x$ is symmetric and log quasiconcave (unimodal),
  then Anderson's theorem~\cite[Thm.~11.1]{Anderson55, Gardner02} implies
  that
  \begin{equation*}
    \P(|x^T \theta - Y| \ge t \mid X = x)
    = \P(|x^T(\theta - \theta\subopt) - \noise| \ge t \mid X = x)
    \ge \P(|\noise| \ge t \mid X = x),
  \end{equation*}
  for all $t \in \R$, and so $\loss(\theta\subopt; X, Y) \preceq
  \loss(\theta; X, Y)$ for all $\theta$, and $\theta\subopt \in \argmin_\theta
  \risk_f(\theta; P)$. 

  In a different vein, we can consider the case that
  $X, Y$ are jointly Gaussian and mean zero,
  \begin{equation*}
    (X, Y) \sim \normal\left(\zeros, \left[\begin{matrix} \Sigma & \gamma \\
        \gamma^T & \sigma^2\end{matrix}\right]
    \right).
  \end{equation*}
  Then for any $\theta$ we have $(X^T \theta - Y) \sim \normal(0,
  \theta^T \Sigma \theta - 2 \theta^T \gamma + \sigma^2)$, and the ordinary
  least-squares solution $\theta\ols = \Sigma^{-1} \gamma = \E[XX^T]^{-1}
  \E[XY]$ evidently uniformly minimizes the variance of $(X^T \theta -
  Y)$. Once again, we thus have the stochastic dominance $\loss(\theta\ols;
  X, Y) \preceq \loss(\theta; X, Y)$ for all $\theta$, and so
  the robust solutions coincide with standard estimators.
\end{example}

\ifdefined\useaosstyle
\vspace{-10pt}
\fi
\begin{example}[Worst-case minority performance and CVaR]
  \label{example:cvar}
  For $0 < \alpha \le 1$,
  the conditional value-at-risk~\cite{RockafellarUr00} (CVaR) is
  \begin{equation*}
    \cvar{\alpha}(\theta; P_0) \defeq \inf_{\eta \in \R}
    \left\{ 
    \alpha^{-1} \E_{P_0}\left[\hinge{\loss(\theta; X) - \eta}\right]
    + \eta \right\}. 
  \end{equation*}
  This corresponds to an uncertainty set arising
  out of limiting $f$- or R\'{e}nyi divergences.
  Recalling the
  R\'{e}nyi divergence~\eqref{eqn:renyi-def}, we have
  $D_\infty(P |\!| Q) \defeq \lim_{\alpha \to \infty} D_\alpha(P |\!| Q)
  = \esssup \log \frac{dP}{dQ}$,
  and if we define $f_{\infty, c}(t) = 0$ for $0 \le t \le c$ and
  $+\infty$ otherwise,
  then the uncertainty region
  \begin{align*}
    \uncertainty_{P_0}
    & \defeq \left\{P \mid D_{\infty}(P |\!| P_0) \le \log \frac{1}{\alpha}
    \right\}
    = \left\{P \mid D_{f_{\infty,\alpha^{-1}}}(P |\!| P_0) \le 1 \right\} \\
    & ~ = \left\{P \mid
    ~ \mbox{there~exists~} Q, ~ \beta \in [\alpha, 1]
    ~ \mbox{s.t.}~ P_0 = \beta P + (1 - \beta) Q
    \right\}
  \end{align*}
  by a calculation~\cite[Example 6.19]{ShapiroDeRu09}.
  The uncertainty set corresponds to
  distributions with minority sub-populations of size at least
  $\alpha$, and
  $\cvar{\alpha}(\theta; P_0) = \sup_{P \in \uncertainty_{P_0}}
  \E_P[\loss(\theta; X)]$
  is the expected loss
  of the worst $\alpha$-sized subpopulation.
\end{example}

\ifdefined\useaosstyle
\vspace{-5pt}
\fi

\noindent
\red{The Kusuoka representation~\cite{Shapiro13k, Kusuoka01} of risk
  measures shows that the robust formulations~\eqref{eqn:objective}
  are worst-case CVaR mixtures,
  $\risk_f(\theta; P_0)
  = \sup_{\mu \in \mc{M}_f} \int_0^1 \cvar{\alpha}(\theta; P_0)
  d\mu(\alpha)$ for a set $\mc{M}_f$ of probability measures on $[0, 1]$.
  They thus correspond to drawing a random sub-population size
  $\alpha$ and measuring the loss of the worst subpopulation of
  $P_0$ mass at least $\alpha$. Precisely
  connecting the subpopulation size and robustness set
  $\{P : \fdiv{P}{P_0} \le \tol\}$ is challenging.  }

We now consider two examples in which data comes from latent \emph{mixtures}
of populations, where within each subpopulation a model is well-specified,
though it is not globally.  In both of these cases---mean estimation and a
linear regression problem---we see that as the robustness parameter $\tol
\uparrow \infty$ in the DRO formulation~\eqref{eqn:objective}, the robust
estimator converges to the minimax estimator minimizing the worst-case loss
across all sub-populations. This recalls \citet{MeinshausenBu15}, who
consider min/max effects in heterogeneous regression problems with known
group identities, but here the DRO estimator recovers a minimax estimator
\emph{without} such knowledge.  The examples are stylized to give explicit
limits, though they convey the intuition that the robust estimators seek to
do well on unknown sub-populations in a reasonably precise way.  In each
example, we consider the conditional value at risk (Ex.~\ref{example:cvar})
for simplicity; the results for higher-order robustness measures are similar
but tedious.

\begin{example}[Mixtures in mean estimation]
  \label{example:mean-estimation}
  Consider a finite number of distinct populations on $\R^d$ indexed by
  $v \in V$, each appearing with probability $p_v > 0$, where under population
  $v$, we observe
  \begin{equation*}
    Y = \theta_v + \noise,
    ~~ \noise \simiid \normal(0, I_d).
  \end{equation*}
  Letting the loss
  $\loss(\theta; y) = \half \ltwo{\theta - y}^2$,
  we define the minimax estimator
  \begin{equation*}
    \theta\maximin \defeq \argmin_\theta
    \max_{v \in V} \ltwo{\theta - \theta_v}^2
    = \argmin_\theta \max_{v \in V} \E_v[\ltwo{\theta - Y}^2].
  \end{equation*}
  The unique vector $\theta\maximin$ coincides with the Chebyshev
  center of the vectors $\{\theta_v\}$~\cite[Ch.~8.5]{BoydVa04};
  it also requires knowledge of the groups $v \in V$.
  In Appendix~\ref{sec:proof-example-mean},
  we show that if
  $\theta_\alpha = \argmin_\theta \cvar{\alpha}(\loss(\theta; Y))$,
  then
  \begin{equation*}
    \theta_1 = \sum_v p_v \theta_v
    ~~ \mbox{and} ~~
    \lim_{\alpha \downarrow 0} \theta_\alpha = \theta\maximin.
  \end{equation*}
  \red{Recalling from Ex.~\ref{example:cvar} that the parameter $\alpha$ is
    inversely proportional to the robustness in the DRO formulation, we see
    the expected behavior: as robustness increases, the DRO estimator
    converges to an estimator minimizing the worst sub-population expected
    loss.}
\end{example}

\begin{example}[Mixtures in linear regression]
  \label{example:mixture-linear-regression}
  We expand the previous example to allow covariates and potentially
  infinite subgroups.
  For groups indexed by $v \in V$, we draw $v \in V$ according to a
  probability measure $\mu$ on $V$, and then conditional on $v$ draw
  \begin{equation}
    \label{eqn:mixture-regression-model}
    X \sim \normal(0, \Sigma_v),
    ~~
    \noise_v \sim \normal(0, \sigma^2_v),
    ~~ Y = X^T \theta_v + \noise_v,
  \end{equation}
  assuming implicitly that all parameters are $v$-measurable.  (To show the
  result in the most straightforward way, we make the simplifying assumptions
  that $0 < \inf_v \sigma_v^2 \le \sup_v \sigma_v^2 < \infty$, that the
  eigenvalues of $\Sigma_v$ are finite and bounded away from $0$ uniformly in
  $v$,
  that $\sup_v \norm{\theta_v} <
  \infty$, and we also assume that for each $\theta \in
  \R^d$, we have $\esssup_v (\theta - \theta_v)^T \Sigma_v (\theta - \theta_v)
  + \sigma_v^2 = \sup_v (\theta - \theta_v)^T \Sigma_v (\theta - \theta_v) +
  \sigma_v^2$. Each of these assumptions is trivial when there are a finite
  number of groups.)
  
  Letting $\E_v$ denote expectation according to the
  model~\eqref{eqn:mixture-regression-model}, let $\loss(\theta; x, y) =
  \half (x^T \theta - y)^2$ be the standard squared error and consider the
  conditional value at risk
  \begin{align*}
    \cvar{\alpha}(\loss(\theta; X, Y))
    & = \inf_{\eta}
    \left\{\frac{1}{\alpha}
    \int \E_v\left[\hinge{\loss(\theta; X, Y) - \eta} \right] d\mu(v)
    + \eta \right\}.
  \end{align*}
  We define the minimax estimator to minimize the
  worst sub-population risk
  \begin{equation*}
    \theta\maximin = \argmin_\theta
    \sup_{v \in V} \left\{
    \E_v[(\theta^T X - Y)^2]
    = (\theta - \theta_v)^T \Sigma_v (\theta - \theta_v)
    + \sigma_v^2
    \right\}.
  \end{equation*}
  In this case, for
  the distributionally robust parameter $\theta_\alpha \defeq \argmin_\theta
  \cvar{\alpha}(\loss(\theta; X, Y))$ and
  ordinary least squares solution
  $\theta\ols = \argmin_\theta \E[\loss(\theta; X, Y)]$, we show in
  Appendix~\ref{sec:proof-mixture-linear-regression} that
  \begin{equation*}
    \theta\ols = \theta_1 = \int \theta_v d\mu(v)
    ~~~\mbox{and}~~~
    \lim_{\alpha \downarrow 0} \theta_\alpha = \theta\maximin.
  \end{equation*}
  \red{We again see the interpolation from an average parameter to one that
    minimizes the worst-case subpopulation risk as the robustness increases
    (i.e.\ $\alpha \downarrow 0$).}
\end{example}

%% file: experiments.tex
\section{Empirical analysis, validation, and choice of uncertainty set}
\label{section:experiments}


As this paper proposes and argues for alternatives to empirical risk
minimization and standard
M-estimation---workhorses of much of
machine learning and statistics~\cite{VanDerVaart98, VanDerVaartWe96,
  HuberRo09}---it is important that we justify our approach.
To that end, we first provide a
number of experiments that illustrate the empirical properties of the
distributionally robust formulation~\eqref{eqn:objective}.  We test our
plug-in estimator~\eqref{eqn:plug-in} on a variety of tasks involving real
and simulated data, and compare its performance with the standard empirical
risk minimizer
\begin{equation*}
  \ermsol \in \argmin_{\theta \in \Theta} \E_{\emp}[\loss(\theta; X)].
\end{equation*}
For concreteness, we focus on the Cressie-Read
(equivalently R\'{e}nyi)
divergence family~\eqref{eqn:cressie-read} with $k \in (1, \infty)$,
experimenting on three related challenges:
\begin{enumerate}[1.]
\item Domain adaptation and distributional shifts, in which we fit
  predictors on a training distribution differing from the test distribution
\item \label{item:tail-performance} Performance on tail losses,
  where we measure quantiles of a model's loss rather than its expected
  losses
\item \label{item:subpopulations} Data coming from multiple heterogeneous
  subpopulations, where we study performance on each subpopulation
  (or worst-case subpopulations).
\end{enumerate}
If our intuition on the distributionally robust risk is accurate, we expect
results of roughly the following form: as we decrease $k$ in the Cressie-Read
divergence~\eqref{eqn:cressie-read}, $f_k(t) \propto t^k - 1$, the solutions
should exhibit more robustness \red{while trading against average-case
  empirical performance,} as the set $\{Q : \fdivs{Q}{P_0} \le \tol\}$ gets
larger. \red{Thus, such models should have better tail behavior or
  generalization on rare or difficult subpopulations compared to standard
  average-case procedures.} We expect increasing $\tol$ to exhibit similar
effects, and we shall see the ways this intuition bears out in our
experiments.

\red{ Since the choice of $f$ and $\rho$ governs the trade-off between
  average and tail performance, we propose two heuristics for choosing
  $\rho$ and $k$, evaluating their performance on simulated examples. Our
  heuristics aim to provide uniform performance over difficult inputs by
  considering proxy subpopulations constructed from the training data,
  though to be clear, the only formal guarantees on robustness they provide
  is robustness to shifts contained in specified by $f_k$ for the chosen and
  $k$ (the duality relationships~\eqref{eqn:dual}
  and~\eqref{eqn:cressie-read-risk} makes the robustness less sensitive to
  $\rho$). Our first heuristic splits the training dataset into $s$ equi-sized
  groups based on the values of the response variable $Y$, where $Y$ has
  highest values in the first group, and the lowest values in the last $s$th
  group.  We split each of the $s$ groups into 80\%/20\% training/validation
  splits, and re-unify all of the 80\% splits to give a new training dataset
  with 80\% of the original data.  We train our robust
  models~\eqref{eqn:plug-in} (varying $\rho$ and $k$) on the new training
  dataset, evaluating these models on the unused data from each group
  (20\%), giving $s$ different empirical losses for a given model. A model's
  score is then its empirical loss on the \emph{worst} of the $s$ held-out
  sets.  We use $s = 5$ groups since this consistently gives a good
  selection procedure across different settings. As our second heuristic, we
  consider scenarios where more is known about the problem. If a small
  auxiliary dataset collected from a worse-than-average subpopulation is
  available, we tune $\rho$ and $k$ on this auxiliary dataset so that
  heuristically, the resulting model performs uniformly well against all
  subpopulations of a \emph{similar size} (the worst-case
  formulation~\eqref{eqn:plug-in} optimizes performance only over large
  enough subpopulations e.g.\ Example~\ref{example:cvar}). Empirically, we
  observe that the second heuristic performs well even on rare subgroups
  that are far from the subpopulation generating the auxiliary dataset.  On
  simulation examples, we observe good worst-case subpopulation performance
  for both procedures, with moderate degradation in the average-case
  performance.}

We begin with simulation experiments that touch on all three of above
challenges in Section~\ref{section:simulation}. To investigate these
challenges on different real-world datasets, in
Section~\ref{section:domain-generalization} we study domain adaptation in the
context of predictors trained to recognize handwritten digits, then test them
to recognize typewritten digits.  In Section~\ref{section:tail-performance}, we study
tail prediction performance in a crime prediction problem. In our final
experiment, in Section~\ref{section:fine-grained}, we study a fine-grained
recognition problem, where a classifier must label images as one of 120
different dog breeds; this highlights a combination of
items~\ref{item:tail-performance} and~\ref{item:subpopulations} on tail
performance and subpopulation performance.

To efficiently solve the empirical worst-case problem~\eqref{eqn:plug-in} for
the Cressie-Read family~\eqref{eqn:cressie-read}, we employ two approaches.
For small datasets (small $n$ and $d$), we solve the dual
form~\eqref{eqn:cressie-read-risk} directly using a conic interior point
solver; we extended the open-source Julia package \texttt{convex.jl} to
implement power cone solvers~\cite{UdellMoZeHoDiBo14} (the package now
contains our implementation).  For larger datasets (e.g.\ $n \approx
10^3-10^5$ and $d \approx 10^2-10^4$), we apply gradient descent with
backtracking Armijo line-searches~\cite{BoydVa04}. The probability
vector $Q^* = \{q_i^*\}_{i = 1}^n \in \R^n_+$ achieving the supremum in the
definition~\eqref{eqn:cr-risk} is unique as long as the loss
vector $[\loss(\theta; \statrv_i)]_{i=1}^n$ is non-constant, which it is in
all of our applications, so $\risk_k$ is differentiable~\cite[Theorem
  VI.4.4.2]{HiriartUrrutyLe93ab} with
\begin{equation*}
  \nabla \risk_k(\theta, \emp) = \sum_{i = 1}^n q_i^* \nabla
  \loss(\theta; \statrv_i)
  ~~ \mbox{and} ~~
  Q^* = \!\! \argmax_{Q: D_{f_k}\left( Q |\!| \emp \right) \le \tol}
  \bigg\{\sum_{i = 1}^n q_i \loss(\theta; \statrv_i) \bigg\}.
\end{equation*}
We use a fast bisection method~\cite{NamkoongDu17} to compute $Q^*$ at every
iteration of our first-order method; see
\url{https://github.com/hsnamkoong/robustopt} for the implementation.

\subsection{Simulation}
\label{section:simulation}
\begin{figure}[t]
  \centering
        \ifdefined\useaosstyle
  \vspace{-10pt}
  \fi
  \begin{tabular}{cc}
    \includegraphics[width=.49\columnwidth]{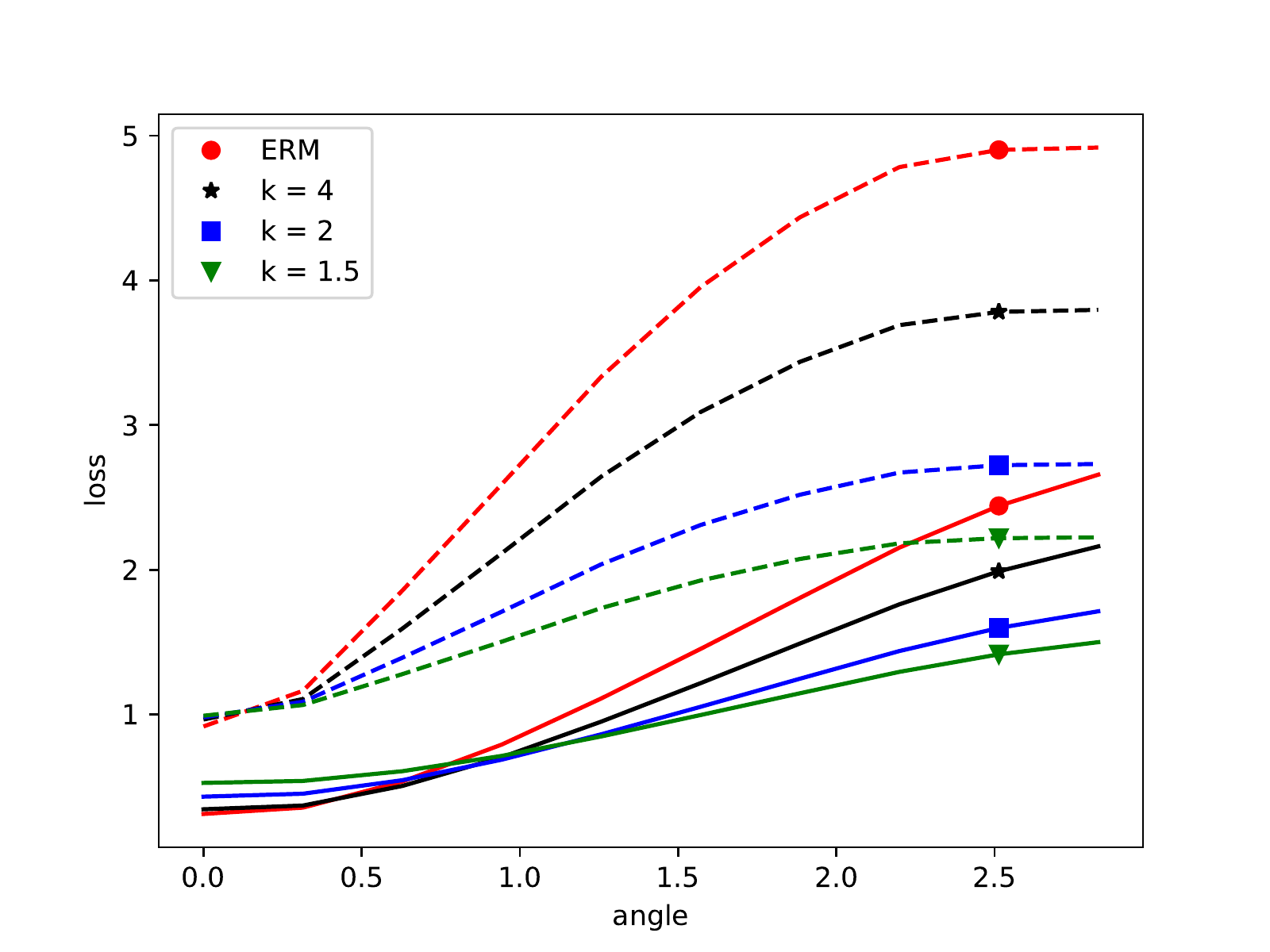}
    &
    \includegraphics[width=.49\columnwidth]{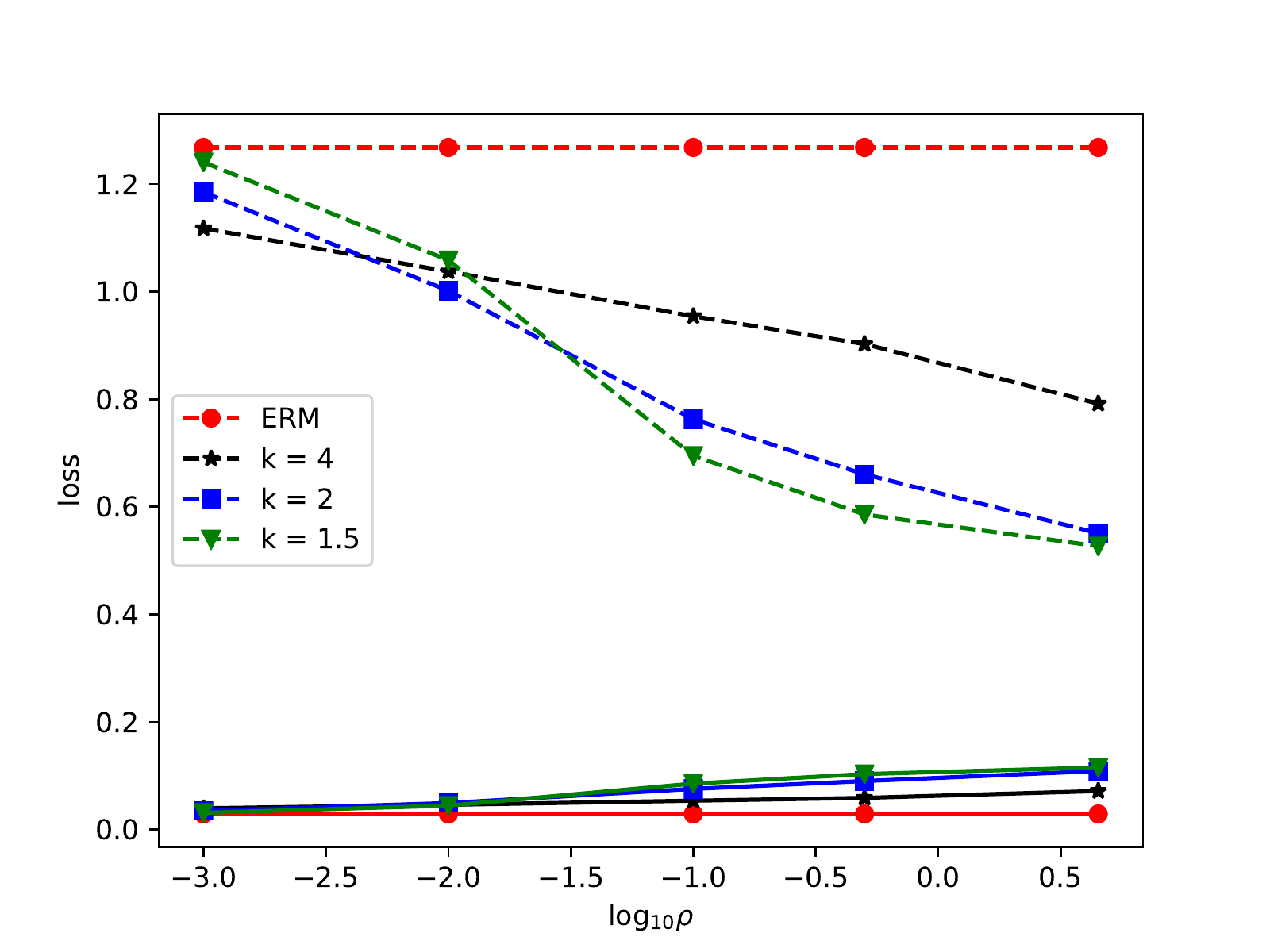} \\
    {\small (a) Classification } & {\small (b) Regression}
  \end{tabular}
  \caption{\label{fig:sim-one} (a) Hinge losses (average and 90th percentile
    in solid and dashed lines, respectively) under distributional shifts
    from $\theta_0\opt$ to $\theta_t\opt = \theta_0\opt \cdot \cos t + v
    \cdot \sin t$. The horizontal axis indexes perturbation $t$.
    (b) Losses on minority group (solid-line) and
    majority group (dotted-line) under the distribution~\eqref{eqn:thresh}. We
    define the minority group as those with $X^1 \le z_{.95}$.}
    \ifdefined\useaosstyle
  \vspace{-10pt}
  \fi
\end{figure}

Our first experiments use simulated data, where we fit linear models for
binary classification and prediction of a real-valued
signal.  We train our models with different values of $f$-divergence
power $k$ and tolerance $\tol$, testing them on
perturbations of the data-generating distribution.


\subsubsection{Domain adaptation and distributional shifts}

We investigate distributional shifts via a binary classification experiment
using the hinge loss $\loss(\theta; (x, y)) = \hinge{1 - y x^\top \theta}$,
where $y \in \{\pm 1\}$ and $x \in \R^d$ with $d = 5$. We
choose a vector $\theta_0\opt \in \R^5$ uniformly on the unit sphere
and generate data
\begin{equation}
  \label{eqn:scheme}
  X \simiid \normal(0, I_d)
  ~~~ \mbox{and} ~~
  Y \mid X = \begin{cases}
    \sign(X^\top \theta_0\opt) & ~~\mbox{w.p.}~~ 0.9 \\
    - \sign(X^\top \theta_0\opt) & ~~\mbox{w.p.}~~ 0.1.
  \end{cases}
\end{equation}
(Our below observations still hold when varying these probabilities.) We train
our models on $n_{\rm train} = 100$ training data points, where we use
$\tol = .5$ and vary values of $k \in \{1.5, 2, 4 \}$ for our distributionally
robust procedure~\eqref{eqn:plug-in}. To simulate distributional shift, we
take a uniformly random vector $v \perp \theta\opt_0$, $v \in \sphere^{d-1}$,
and for $s \in [0, \pi]$ define
$\theta_s\opt = \theta\opt_0 \cdot \cos s + v \cdot \sin s$, so that
$\theta_\pi\opt = -\theta_0\opt$.  For each perturbation, we generate
$n_{\rm test} = 100,000$ test examples using the same
scheme~\eqref{eqn:scheme} with $\theta_t\opt$ replacing $\theta_0\opt$.

We measure both average and 90\%-quantile losses for our problems.  Based on
our intuition, we expect that the lower $k$ is (recall that
$f_k(t) \propto t^k$), \red{the better the fitted model should perform on
high quantiles of the loss, with potentially worse average
performance.} Moreover, for $s = 0$, we should see that ERM and large $k$
solutions exhibit the best average performance, with growing $s$ reversing
this behavior. In
Figure~\ref{fig:sim-one}(a), we plot the average loss (solid line) and the
$90\%$-quantile of the losses (dotted line) on the shifted test sets, where
the horizontal axis displays the rotation $s \in [0,\pi]$.  The plot
bears out our intuition: \red{the distributionally robust solution $\robsol$ has
worse \emph{mean} loss on the original distribution than empirical risk
minimization (ERM) while achieving significantly smaller loss on the
distributional shifts.}  The ordering of the mean performance of the different
solutions inverts as the perturbation grows: under no perturbation ($s = 0$),
the least robust method (ERM) has the best performance, while the most robust
method (corresponding to $k = \frac{3}{2}$) performs the best
under large distributional perturbations ($s$ large).

\subsubsection{Tail performance}

\providecommand{\noise}{\varepsilon}

We transition now to regression, investigating performance on rare
examples, where the goal is to predict $y \in \R$ from $x \in \R^d$ and we use
loss $\loss(\theta; (x, y)) = \half (y - x^\top \theta)^2$.  In this case, we
take $d = 5$ and generate data $X \simiid N(0, I_d)$,
$\noise \sim N(0, .01)$,
\begin{equation}
  \label{eqn:thresh}
  Y = \begin{cases}
    X^\top \param\opt + \noise &~~ \mbox{if}~~X^1 \le z_{.95} = 1.645 \\
    X^\top \param\opt + X^1 + \noise &~~\mbox{otherwise},
  \end{cases}
\end{equation}
where we choose $\param\opt$ uniformly on the unit sphere $\sphere^{d-1}$ and
$X^1$ denotes the first coordinate of $X$. (We use very small noise to
highlight the more precise transition between average-case and higher
percentiles.)  As the effect of $X^1$ changes only 5\% of the time (when it is
above $z_{.95}$), we expect ERM to have poor performance on rare events when
$X^1 \ge 1.645$, or in the tails generally. In addition, a fully robust
solution is $\theta^{\rm rob} = \theta\opt + \half e_1$, as this
minimizes worst-case expected loss across the two cases~\eqref{eqn:thresh}; we
expect that for high robustness parameters ($\tol$ large) the robust model
should have worse average performance but about half of the losses at higher
quantiles. We simulate $n_{\rm train} = 2000$ training data points, and train
the distributionally robust solution~\eqref{eqn:plug-in} with
$\rho \in \{.001, .01, .1, .5, 4.5\}$, and $k \in \{1.5, 2, 4\}$. In
Figure~\ref{fig:sim-one}(b), we plot the mean loss under the data generation
scheme~\eqref{eqn:thresh} as solid lines and the 90\%-quantile as a dotted
line. \red{We see once again that the robust solutions trade tail performance for
average-case performance.} The tail performance (90\%-quantile loss) improve
with increasing robustness level $\tol$, with slight degradation in average
case performance.

\begin{figure}[t]
  \centering
  \begin{tabular}{cc}
    \includegraphics[width=.49\columnwidth]{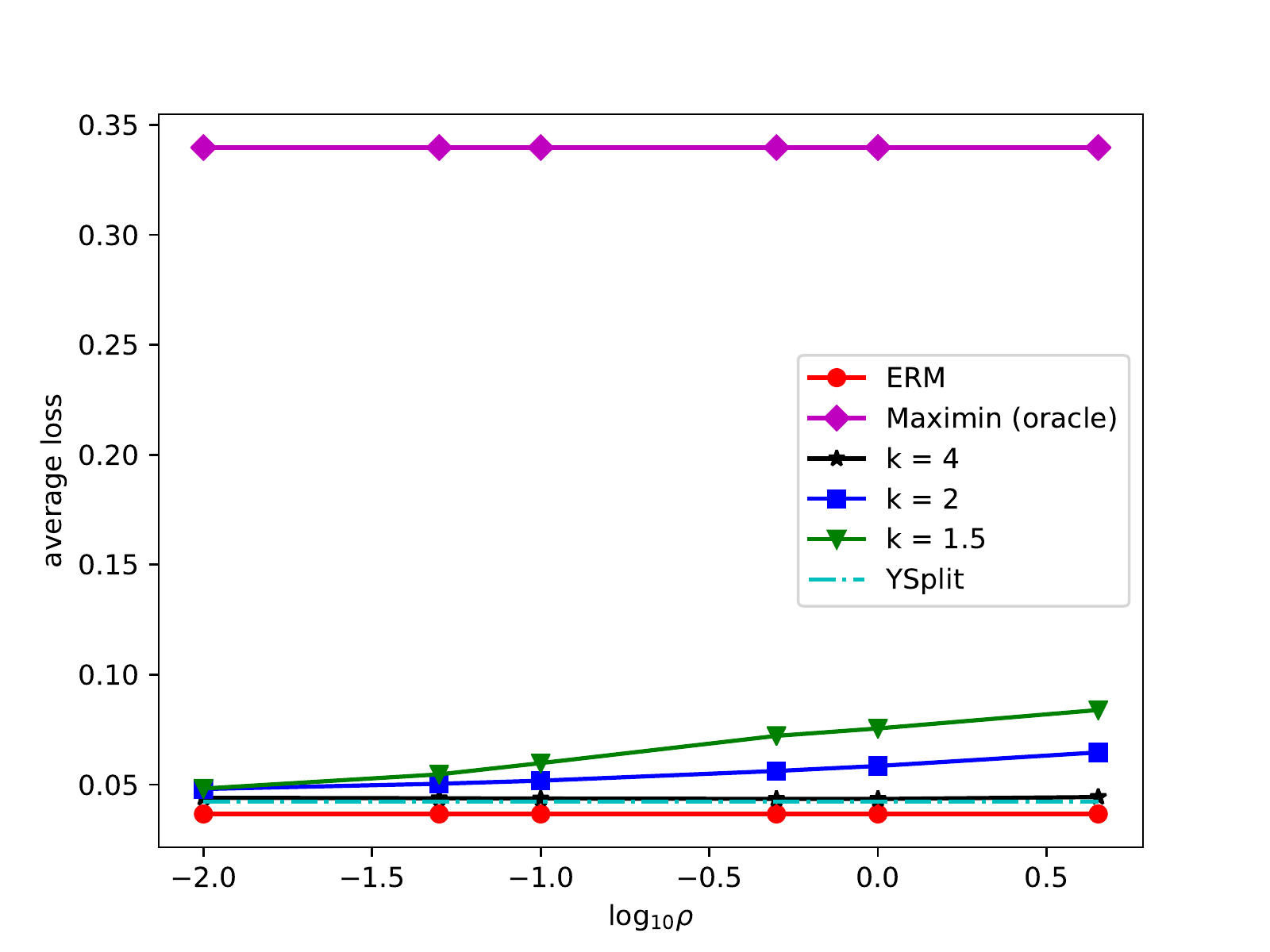}
    &
    \includegraphics[width=.49\columnwidth]{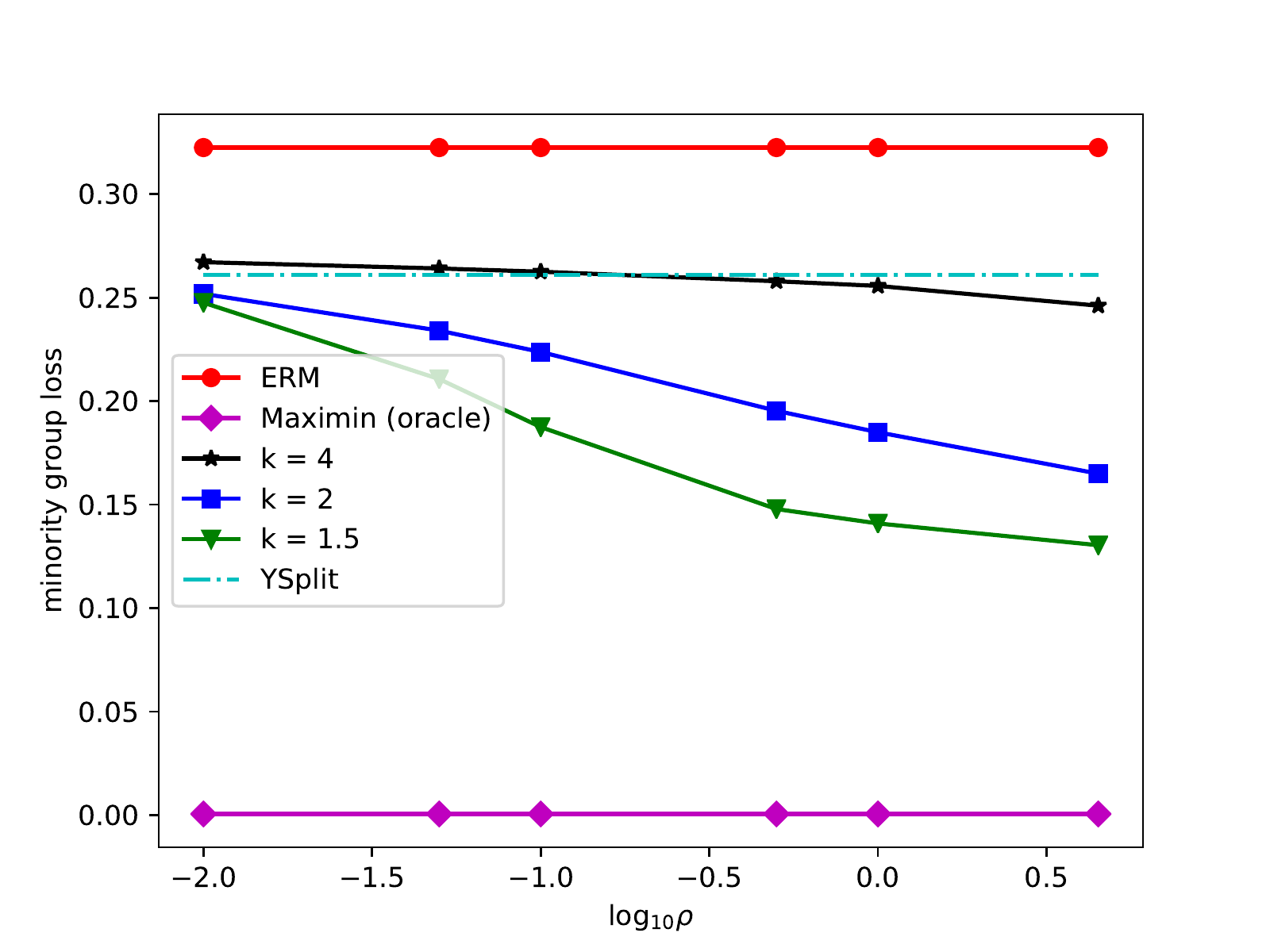}
    \\
    \small{(a) Average loss} & \small{(b) Loss on minority group}
  \end{tabular}
  \caption{ \label{fig:sim-two} Two groups: Figures (a) and (b) plots average
    and minority group losses under the
    distribution~\eqref{eqn:first-scenario}. ``YSplit'' is the performance of
    the model whose $\rho$ and $k$ was chosen based on groups formed by sorted
    values of $Y$.}
    \ifdefined\useaosstyle
  \vspace{-10pt}
  \fi
\end{figure}

\subsubsection{Performance on different subpopulations}
\label{section:subpopulation}


\begin{figure}[t]
  \centering
  \begin{tabular}{cc}
    \includegraphics[width=.49\columnwidth]{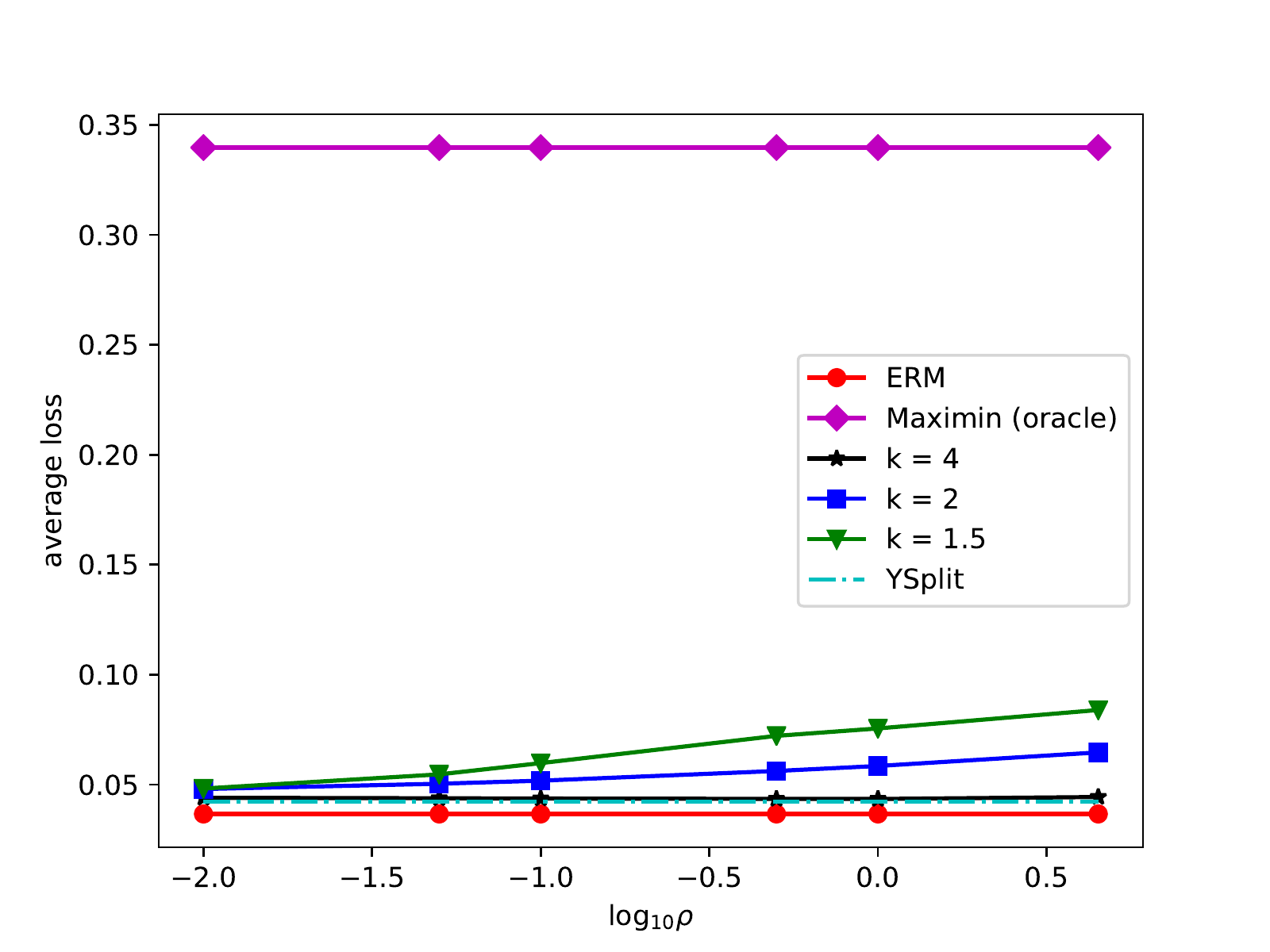}
    &
    \includegraphics[width=.49\columnwidth]{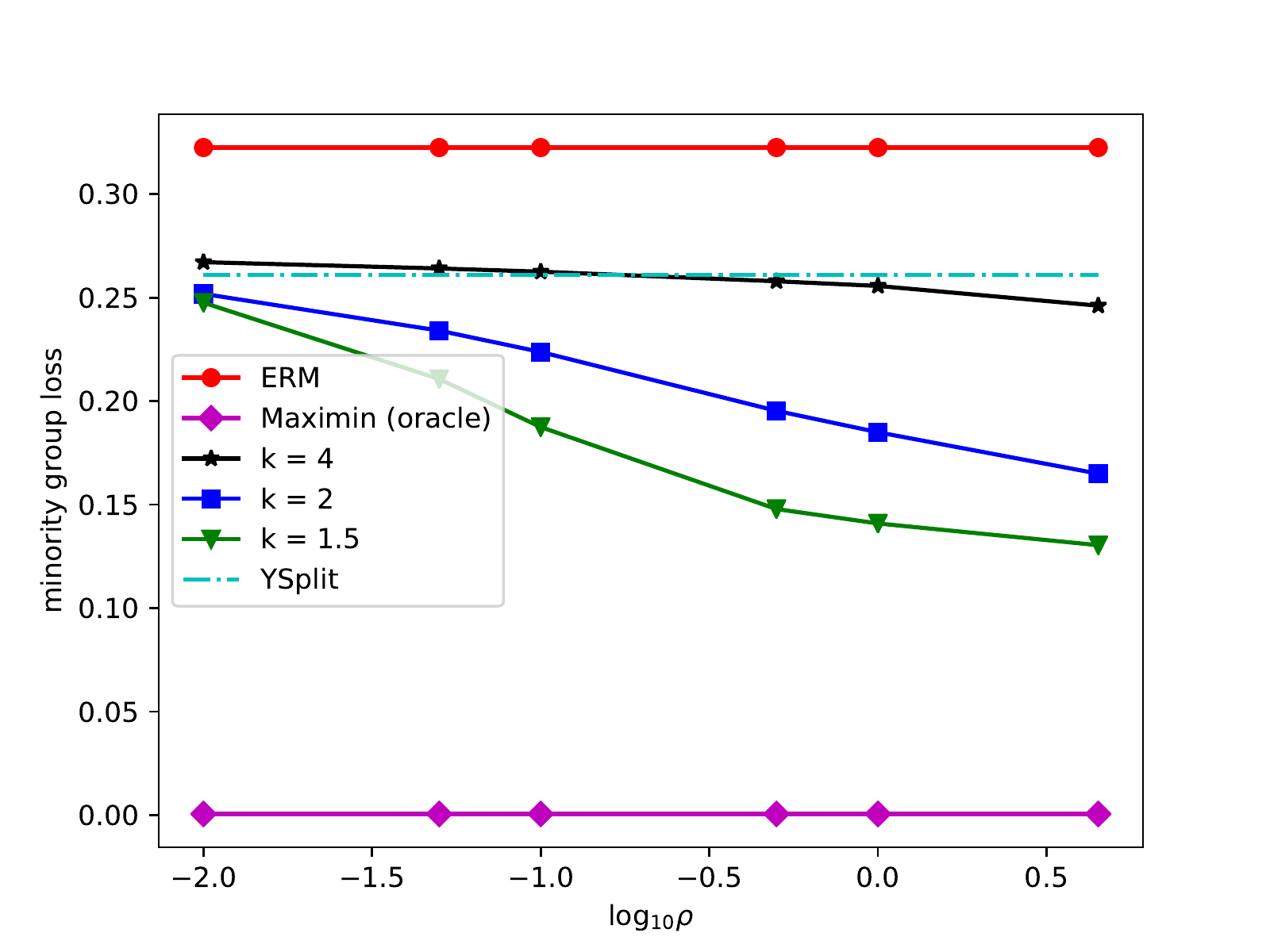}
    \\
    \small{(a) Average loss} & \small{(b) Loss on minority group}
  \end{tabular}
  \caption{ \label{fig:sim-inf} Infinite groups: Figures (a) and (b) plot average and minority
    group losses under the
    distribution~\eqref{eqn:infinite-scenario}. ``YSplit'' is the performance of
    the model whose $\rho$ and $k$ was chosen based on groups formed by sorted
    values of $Y$, and ``$G = .5$'' chose $k$ and $\rho$ based on auxiliary
    data with intervention $G = 0.5$.}
    \ifdefined\useaosstyle
  \vspace{-10pt}
  \fi
\end{figure}

For our final small-scale simulation, we study item~\ref{item:subpopulations}
(subpopulation performance) by considering a two-dimensional regression
problem with heterogeneous subpopulations. We consider two scenarios:
a two-group setting and
an infinite number of groups. In each scenario, we demonstrate the
performance of our heuristic procedure for choosing $\rho$ and $k$;
these subpopulation scenarios are appropriate for succinctly characterizing
the trade-off between average and tail subpopulations. Our tuning procedure
provides good performance on the (latent) worst-case subpopulation even when
the proxy subpopulation for tuning $\rho$ and $k$ is far from the rare
subpopulation. In what follows, we denote by ``YSplit'' our first
proposal that chooses $\rho$ and $k$ based  on sorted values of $Y$.


\paragraph{Two groups} In our first scenario, for
$\theta_0\opt = (1, .1),~\theta\opt_1 = (1, 1)$, we generate
\begin{equation}
  \label{eqn:regression-group}
  Y = X^\top ((1-G) \theta\opt_{0} + G \theta\opt_1) + \noise
\end{equation}
where $X \sim N(0, I_2)$, $\noise \sim \normal(0, .01)$, and $G \in [0, 1]$
indicates a random \emph{latent group}. We assume that $X$, $G$ and $\noise$
are mutually independent. Both the distributionally robust
procedure~\eqref{eqn:plug-in} and ERM are oblivious to the label $G$, where we
think of $G = 1$ as the \emph{majority} group, and $G = 0$ as the
\emph{minority} group. We simulate $n_{\rm train} = 1000$ training data
points, and train ERM and robust models~\eqref{eqn:plug-in} on varying values
of $k$ and $\rho$. We let
\begin{equation}
  \label{eqn:first-scenario}
  G = \begin{cases}
    0 ~& \mbox{with probability}~.1 ~~(\mbox{minority})\\
    1  ~& \mbox{with probability}~.9~~(\mbox{majority})
    \end{cases}
\end{equation}
In this two-group setting, we also consider the maximin
effects estimator~\cite{MeinshausenBu15}
\begin{equation*}
  \what{\theta}_{n}^{\rm maximin} = \argmax_{\theta} \min_{g = 0, 1}
  \left\{ 2\theta^\top \what{\Sigma}_{n, g} \theta\opt_{g}
    - \theta^\top \what{\Sigma}_{n, g} \theta \right\}
\end{equation*}
as a benchmark, where $\what{\Sigma}_{n, g}$ is the empirical covariance
matrix of the $X_i$ with $G_i = g$, which maximizes the explained
variance for each group~\cite{MeinshausenBu15}. The oracle estimator
$\what{\theta}^{\rm maximin}_n$ requires knowledge of the
labels $G_i$ and the group-specific regressors $\theta\opt_g$ for $g = 0, 1$.


In Figure~\ref{fig:sim-two}(a) and (b), we plot the average and minority group
losses for the different methods, respectively.  Here the robust methods
interpolate between the empirical risk minimizing (ERM) solution---which has
the best average loss and worst minority group loss---and the maximin
estimator $\what{\theta}_n^{\rm maximin}$, which sacrifices performance on the
average loss for strong minority group performance. \red{The distributionally
robust estimators $\what{\theta}_n$ exhibit tradeoffs between the two regimes,
improving performance on the minority population at smaller degradation in the
average loss.} The parameters $\tol$ and $k$ allow flexibility in achieving
these tradeoffs, though they of course must be set appropriately in
applications. Our first heuristic (``YSplit'') chooses $\rho$ and $k$ based on
groups formed by sorted values of $Y$, and improves minority performance while
sacrificing very little average-case performance.



\paragraph{Infinite groups} For our last scenario, we again generate $X$ and $Y$
following the equation~\eqref{eqn:regression-group}, but with
\begin{equation}
  \label{eqn:infinite-scenario}
  G \sim P_G~\mbox{with density}~p_G(g) \propto (1-g)^{-\frac{1}{3}},
\end{equation}
so small values of $G$ again correspond to rare minority
subpopulations. To study how $k$ and $\rho$ can be tuned if a small auxiliary
dataset is available, we generate a small auxiliary dataset from the
distribution~\eqref{eqn:regression-group} with group $G = .5$, which we
interpret as a particular group intervention; we simulate
$n_{\rm auxiliary} = 100$ observations from $G = .5$, which is small compared
to $n_{\rm train} = 1000$ training examples. We refer to choosing $k$ and
$\rho$ with the least prediction error on this auxiliary validation data as
the ``$G = 0.5$'' method.

As earlier, we plot in Figure~\ref{fig:sim-inf}(a) and (b) the average and
minority group ($G = 0$) losses for the different methods. The minority group
$G = 0$ now never appears in the training set, and small values of $G$ are
rare under the distribution~\eqref{eqn:infinite-scenario}. Our first
heuristic ``YSplit'' chooses a model that \red{balance average and minority
performance}, although it is somewhat conservative. Our second proposal, the
$G = 0.5$ method, achieves \red{good performance on the rare minority group while
sacrificing little average performance}, despite the fact that the auxiliary
data was collected from the group $G = 0.5$ that is far from the minority
group $G = 0$.

\subsection{Domain generalization for classification and digit recognition}
\label{section:domain-generalization}

\begin{figure}[h!]
  \begin{center}
    \begin{tabular}{cc}
      \hspace{-.4cm}
      \begin{overpic}[width=.5\columnwidth]{
          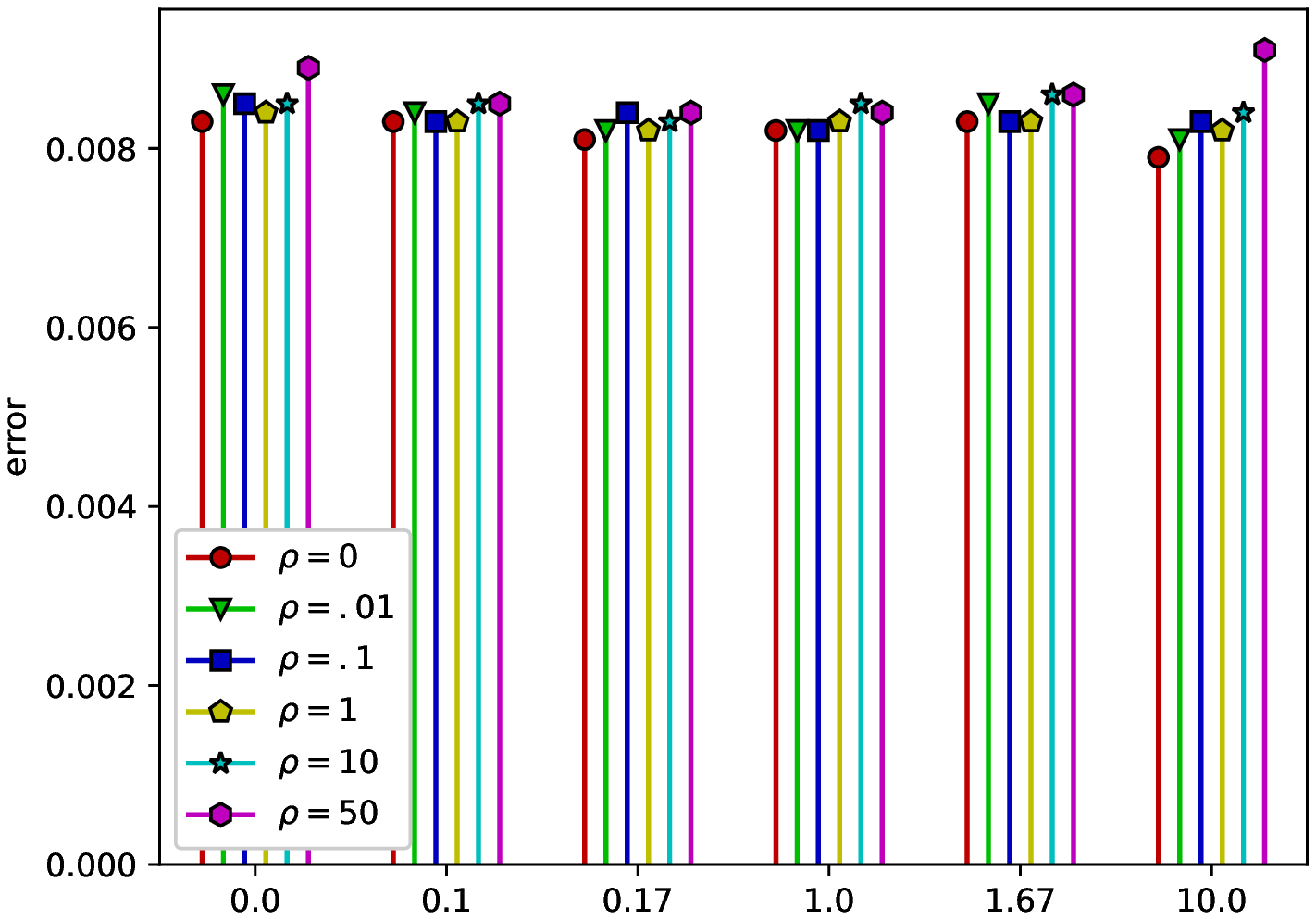}
        \put(35,1){\footnotesize{\% typewritten digits}}
      \end{overpic}
      & 
      \begin{overpic}[width=.5\columnwidth]{%
          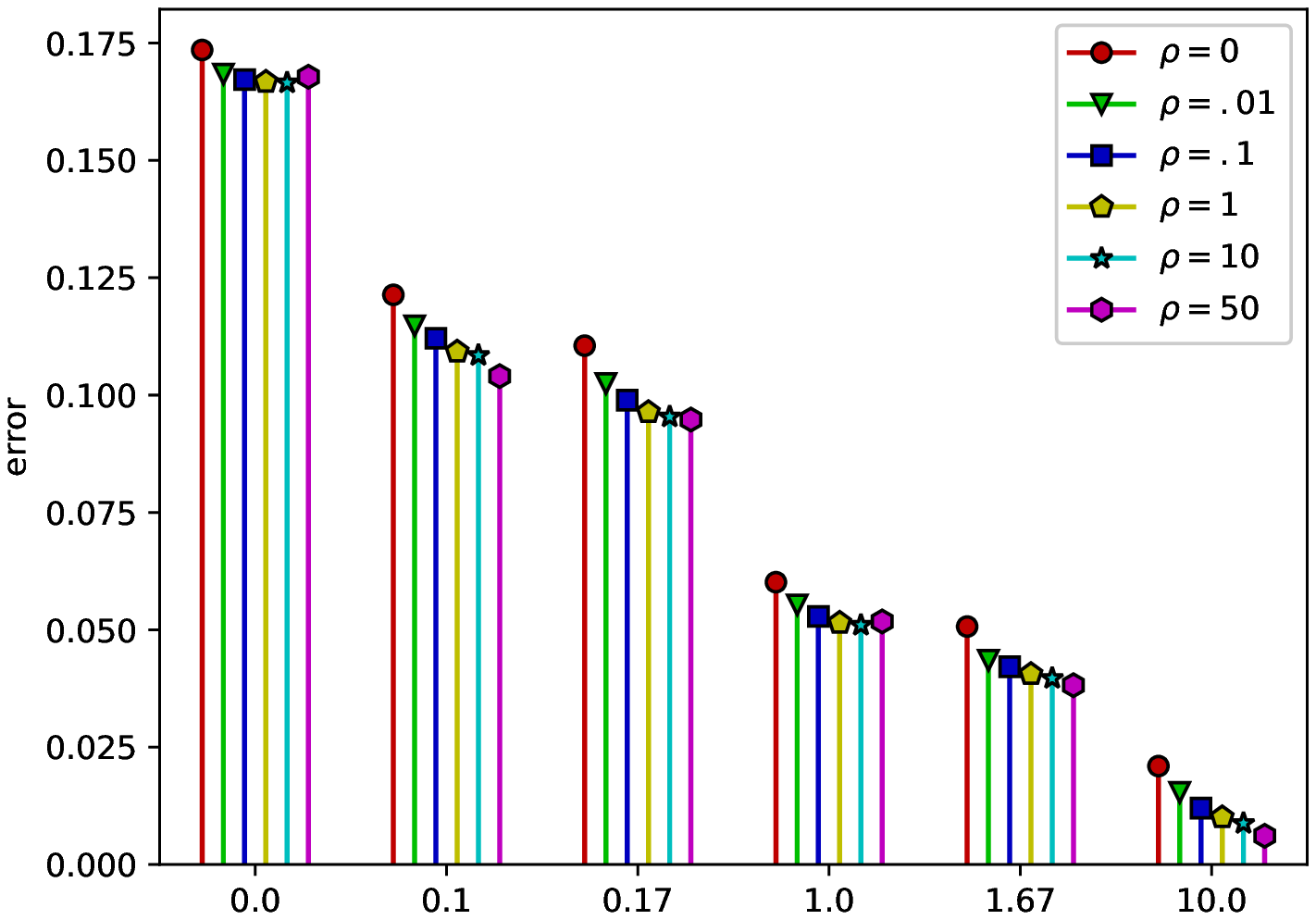}
        \put(35,1){\footnotesize{\% typewritten digits}}
      \end{overpic} \\
      \small{(a) MNIST hand-written Digits}
      & \small{(b) All type-written Digits} \\
      \hspace{-.4cm}
      \begin{overpic}[width=.5\columnwidth]{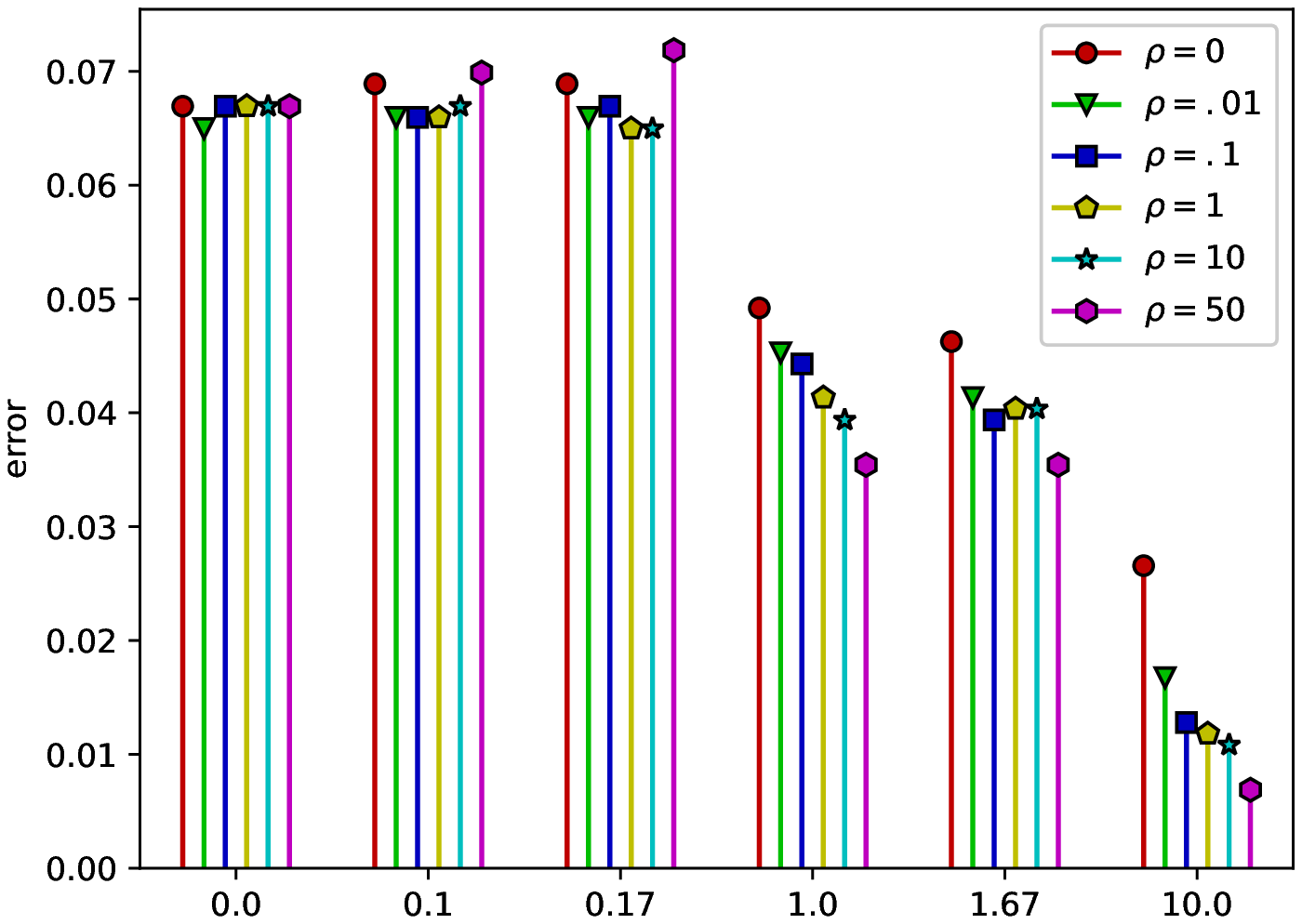}
        \put(35,1){\footnotesize{\% typewritten digits}}
      \end{overpic}
      & 
      \begin{overpic}[width=.5\columnwidth]{%
          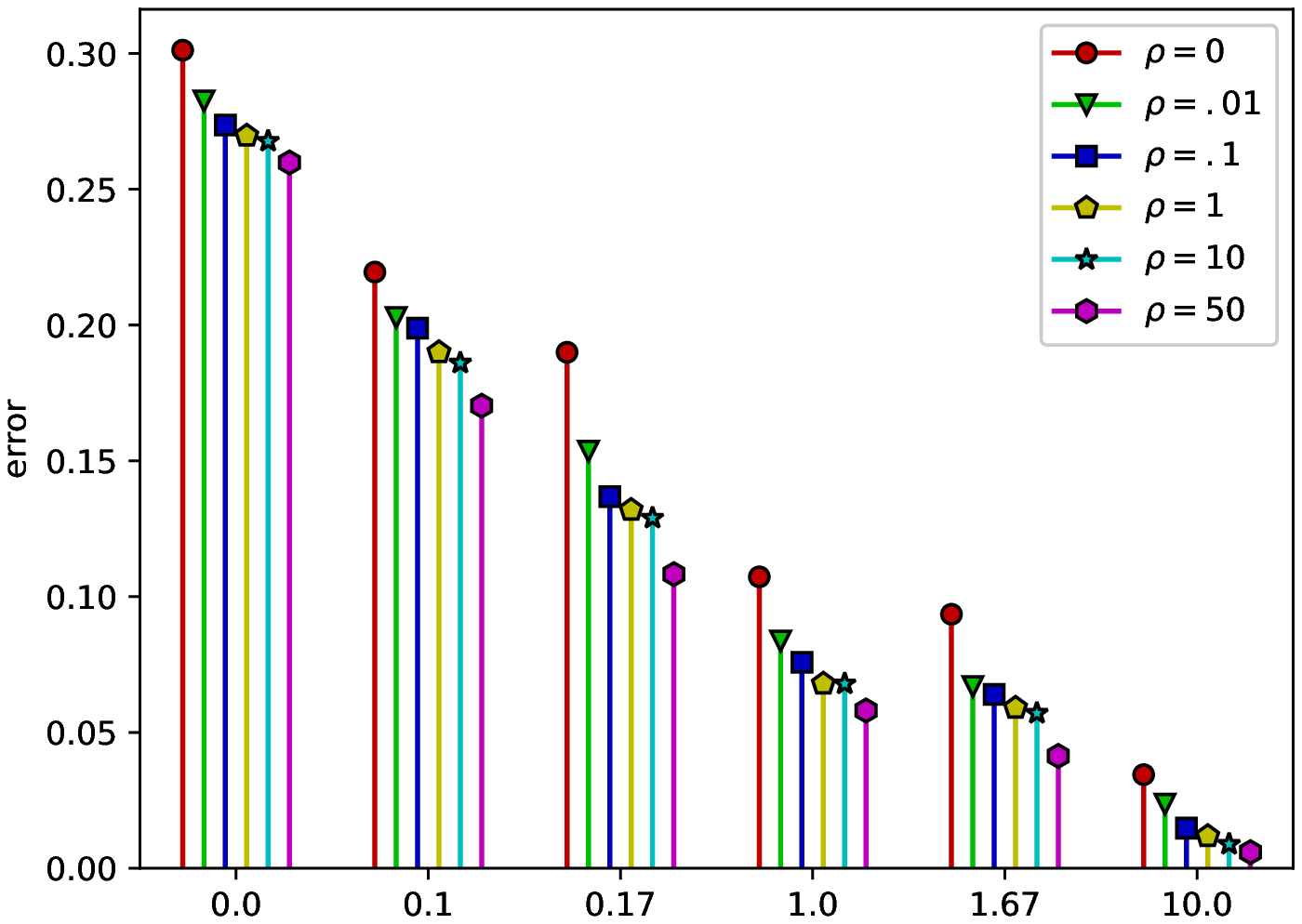}
        \put(35,1){\footnotesize{\% typewritten digits}}
      \end{overpic} \\
      \small{(c) Type-written Digit $3$ (easy class)}
      & \small{(d) Type-written Digit $9$ (hard class)}
    \end{tabular}
    \caption{ \label{fig:mnist-font} (a) Test error on the
      \emph{hand-written digits} (MNIST test dataset).  (b)--(d) Test errors
      on \emph{type-written digits}. Models were trained on data consisting
      of MNIST hand-written digits with 0--10\% replaced by type-written
      digits. The horizontal axis of each plot denotes percentage of
      type-written digits (relative to handwritten) in training. Each of the
      six lines represents a different value of $\tol$ used in training,
      where $\tol = 0$ corresponds to empirical risk minimization (ERM).
      (b) Classification error on entire test set of type-written digits.
      (c) Classification error on digit 3 of the type-written digits.
      (d) Classification errors for digit 9 of the type-written digits.}
  \end{center}
  \ifdefined\useaosstyle
  \vspace{-10pt}
  \else
  \vspace{-15pt}
  \fi
\end{figure}

\begin{table}[h]
  \centering
  \begin{filecontents*}{./figures/mnistfont/mix.csv}
ratio,erm,rob,ermnine,robnine,ermsix,robsix,ermthree,robthree
0,17.35,16.78,30.12,25.98,35.63,38.39,6.69,6.69
0.1,12.14,10.4,21.95,17.03,21.06,14.27,6.89,6.99
0.17,11.05,9.48,19,10.83,19.69,12.8,6.89,7.19
1,6.01,5.18,10.73,5.81,7.97,7.97,4.92,3.54
1.67,5.07,3.82,9.35,4.13,6.59,5.91,4.63,3.54
10,2.1,0.61,3.44,0.59,1.77,0.39,2.66,0.69
\end{filecontents*}
\pgfplotstabletypeset[
  col sep=comma,
  string type,
  every head row/.style={%
    before row={\hline
      Minority
      & \multicolumn{2}{c}{All Digits}
      & \multicolumn{2}{c}{Digit $9$ (hard)}
      & \multicolumn{2}{c}{Digit $6$ (hard)}
      & \multicolumn{2}{c}{Digit $3$ (easy) }
      \\
    },
    after row=\hline
  },
  every last row/.style={after row=\hline},
  columns/ratio/.style={column name=proportion, column type=l},
  columns/erm/.style={column name=ERM, column type=l},
  columns/rob/.style={column name=$\tol \equalto 50$, column type=c},
  columns/ermnine/.style={column name=ERM, column type=l},
  columns/robnine/.style={column name=$\tol \equalto 50$, column type=c},
  columns/ermsix/.style={column name=ERM, column type=l},
  columns/robsix/.style={column name=$\tol \equalto 50$, column type=c},
  columns/ermthree/.style={column name=ERM, column type=l},
  columns/robthree/.style={column name=$\tol \equalto 50$, column type=c}
  ]{./figures/mnistfont/mix.csv}
  \caption{Test error on type-written digits  ($\%$)}
    \label{table:mnistfont}
  \ifdefined\useaosstyle
  \vspace{-8pt}
  \fi
\end{table}

In this first of our real experiments, we consider a multi-class digit
classification example, investigating domain generalization, though we
conflate this with item~\ref{item:subpopulations} (multiple
subpopulations). We construct our training set as a mixture of MNIST
hand-written digits~\cite{DenkerGaGrHeHoHuJaBaGu88} (majority population) and
type-written digits consisting of different fonts~\cite{deCamposBaVa09}
(minority population).  We fix the number of training examples, and vary the
minority proportions of type-written digits from 0--10\% of the training
data. In the MNIST hand-written training dataset comprising of
$n_{\rm train} = 60,000$ digits, we replace
$n \in \{ 0, 6, 10, 60, 100, 600 \}$ images per digit by randomly drawn digits
from the type-written dataset (with the same label).

Our classifiers have no knowledge of whether an image is hand-written
or type-written, and our goal is to learn models that perform uniformly well
across both majority (hand-written) and minority (type-written)
subpopulations.  We compare our procedure~\eqref{eqn:plug-in} with $k = 2$
against the ERM solution $\ermsol$, where we vary $\tol$ and the latent
minority proportion.  We evaluate our classifiers on both hand- and
type-written digits on held-out test sets.

For $y \in \{0, \ldots, 9\}$ and $x \in \R^d$,
we use the multi-class logistic loss
  $\loss(\theta; (x, y))
  = \log (\sum_{i = 0}^k \exp((\theta_i - \theta_y)^\top x)
)$,
where $\theta_i \in \R^d$.
For our feature vector $X$, we use the $d = 4509$-dimensional
output of the final fully connected layer of
LeNet~\cite{LeCunBoDeHeHoHuJa89} after $10^4$ stochastic gradient steps
on the training dataset (see~\cite{JiaShDoKaLoGiGuDa14}
for detailed hyper-parameter settings). We constrain
our parameter matrix $[\theta_0, \ldots, \theta_9]$
to lie in the Frobenius norm ball of radius $r = 5$, chosen by cross
validation on ERM ($\tol = 0$). 

Returning to the justification for our development, \red{we expect our robust
  models to exhibit better performance on rare and difficult test data when
  compared against ERM models.  This prediction is mostly consistent with our
  observations, though the effects are not always strong. We suspect this is
  because the test data we construct is different from the
  worst-case scenario; the procedure~\eqref{eqn:plug-in} can be conservative
  as it guarantees uniform performance by optimizing the worst-case
  performance.} In Figure~\ref{fig:mnist-font}, we plot the classification
errors over the minority proportion as we vary $\tol$ (so that $\tol = 0$
corresponds to ERM), summarizing the classification errors in
Table~\ref{table:mnistfont}. In Figure~\ref{fig:mnist-font}(a), we observe
virtually the same performance on the hand-written test set (majority) across
different radii $\tol$ (error below 1\%, with a decrease in accuracy of at
most .1--.2\%). On a test set of all typed digits
(Figure~\ref{fig:mnist-font}(b)), the robust solutions exhibit a 1--2\%
improvement over the non-robust (ERM) solution in each mixture of typewritten
digits (minority proportions) into the training data, which is larger than the
persistent .1--.2\% degradation on handwritten recognition.  The trend of
robust improvements on typewritten digits is more pronounced on the harder
classes: the gap between $\ermsol$ and $\robsol$ widens up to $9\%$ on the
digit $9$ (see Table~\ref{table:mnistfont} and
Fig.~\ref{fig:mnist-font}(d)). We observe that $\robsol$ consistently performs
well on the latent minority (type-written) subpopulation by virtue of
upweighting the hard instances in the training set.

\ifdefined\useaosstyle
\begin{figure}[t!]
  \begin{center}
    \begin{tabular}{cc}
      \hspace{-.4cm}
      \begin{overpic}[width=.5\columnwidth]{
          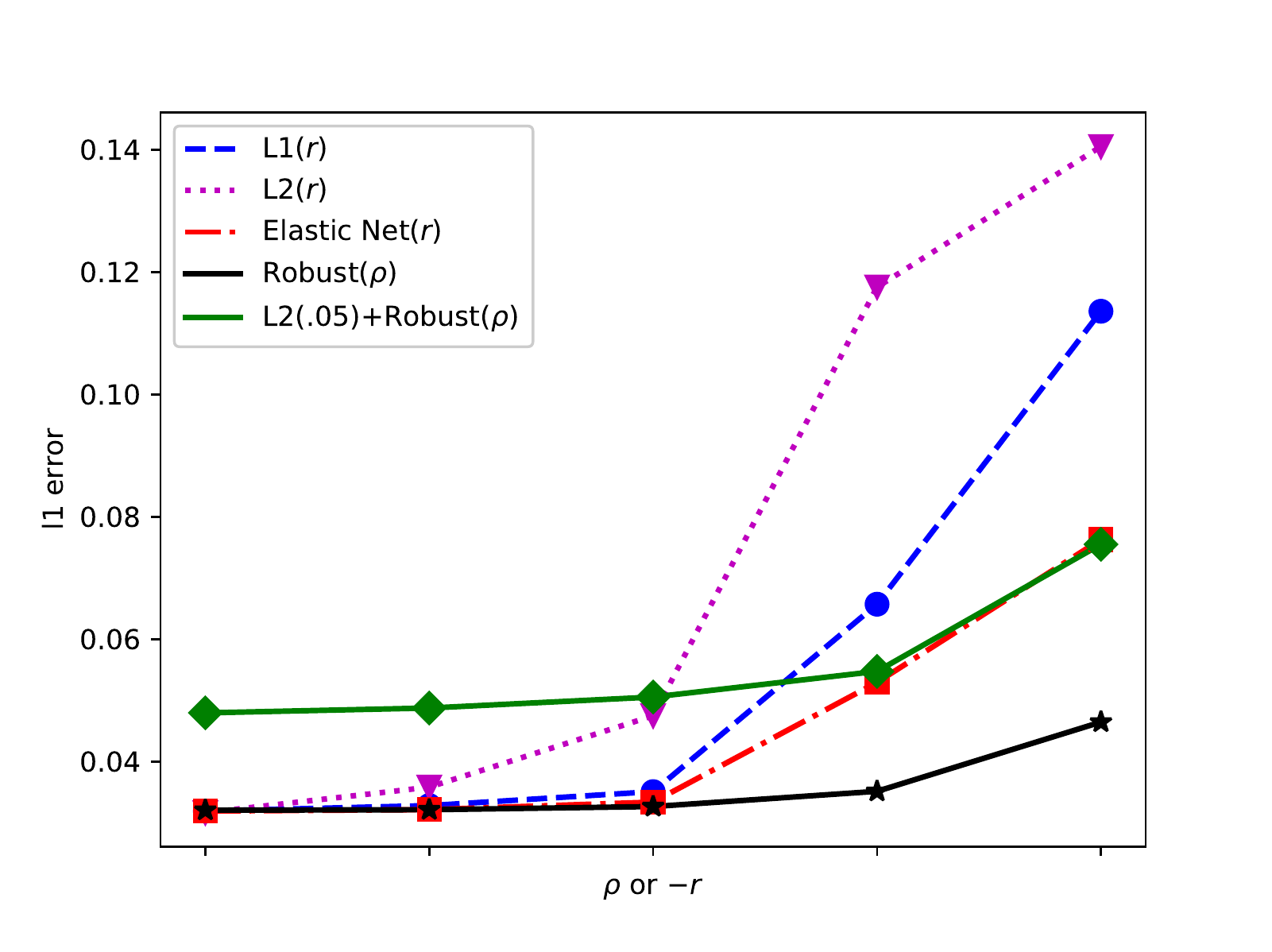}
        \put(45,3){
          \tikz{\path[draw=white,fill=white] (0, 0) rectangle (1cm,.25cm);}}
        \put(8,71){\scriptsize{$\begin{matrix} r_1 = 5 \\ r_2 = 50
            \end{matrix}$}}
        \put(26.5,71){\scriptsize{$\begin{matrix} r_1 = 1 \\ r_2 = 10
            \end{matrix}$}}
        \put(45,71){\scriptsize{$\begin{matrix} r_1 = .5 \\ r_2 = 5
            \end{matrix}$}}
        \put(63.5,71){\scriptsize{$\begin{matrix} r_1 = .1 \\ r_2 = 1
            \end{matrix}$}}
        \put(80,71){\scriptsize{$\begin{matrix} r_1 = .05 \\ r_2 = .5
            \end{matrix}$}}
        \put(11,3){\scriptsize{$\tol = .001$}}
        \put(29.5,3){\scriptsize{$\tol = .01$}}
        \put(48,3){\scriptsize{$\tol = .1$}}
        \put(66.5,3){\scriptsize{$\tol = 1$}}
        \put(83,3){\scriptsize{$\tol = 10$}}
      \end{overpic}
   & 
     \begin{overpic}[width=.5\columnwidth]{
         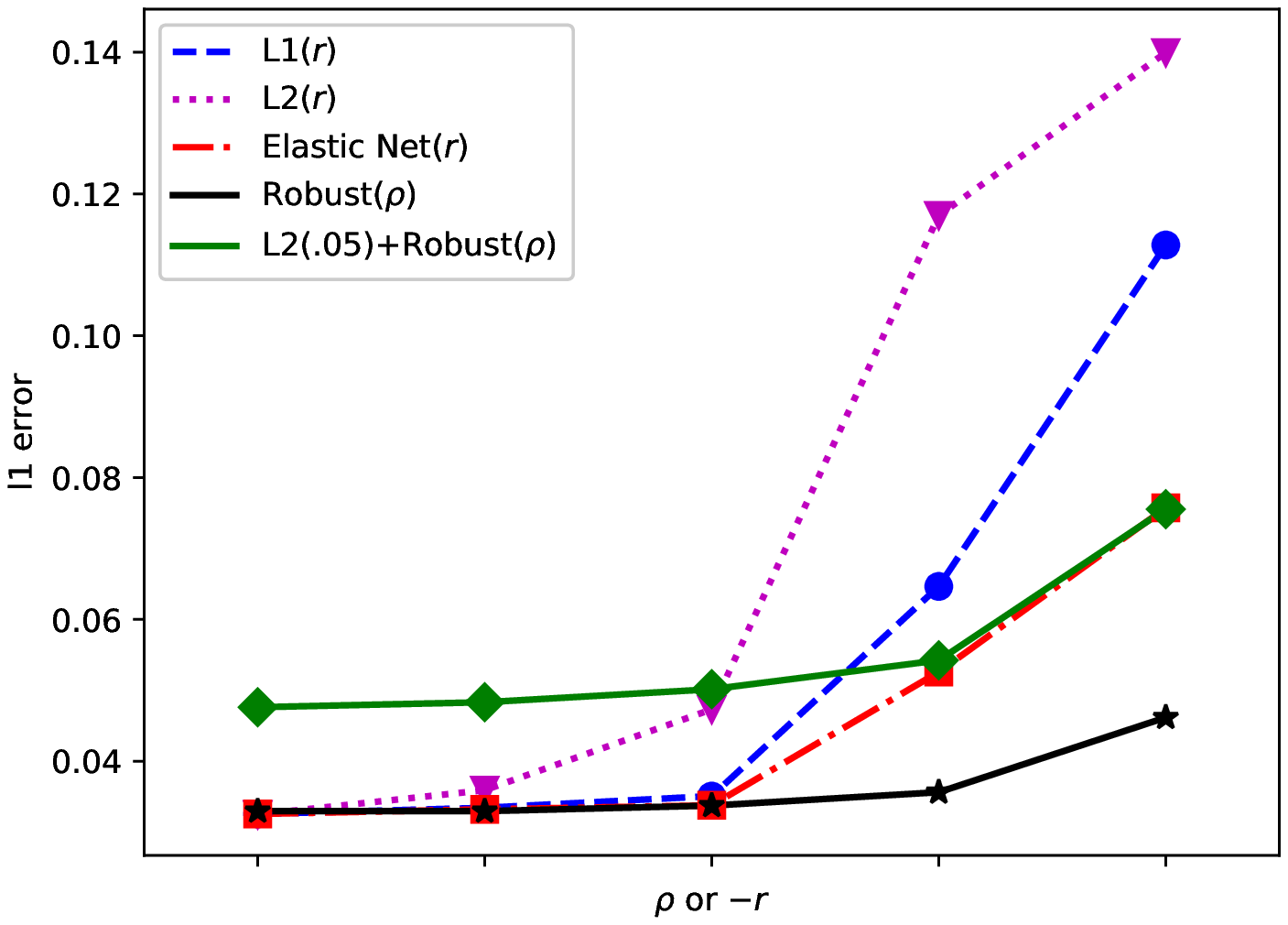}
        \put(45,3){
          \tikz{\path[draw=white,fill=white] (0, 0) rectangle (1cm,.25cm);}}
        \put(8,71){\scriptsize{$\begin{matrix} r_1 = 5 \\ r_2 = 50
            \end{matrix}$}}
        \put(26.5,71){\scriptsize{$\begin{matrix} r_1 = 1 \\ r_2 = 10
            \end{matrix}$}}
        \put(45,71){\scriptsize{$\begin{matrix} r_1 = .5 \\ r_2 = 5
            \end{matrix}$}}
        \put(63.5,71){\scriptsize{$\begin{matrix} r_1 = .1 \\ r_2 = 1
            \end{matrix}$}}
        \put(80,71){\scriptsize{$\begin{matrix} r_1 = .05 \\ r_2 = .5
            \end{matrix}$}}
        \put(11,3){\scriptsize{$\tol = .001$}}
        \put(29.5,3){\scriptsize{$\tol = .01$}}
        \put(48,3){\scriptsize{$\tol = .1$}}
        \put(66.5,3){\scriptsize{$\tol = 1$}}
        \put(83,3){\scriptsize{$\tol = 10$}}
     \end{overpic}
      \vspace{.3cm} \\
      (a) Median training loss
      & (b) Median test loss \vspace{.5cm} \\ 
      \hspace{-.4cm}
      \begin{overpic}[width=.5\columnwidth]{
          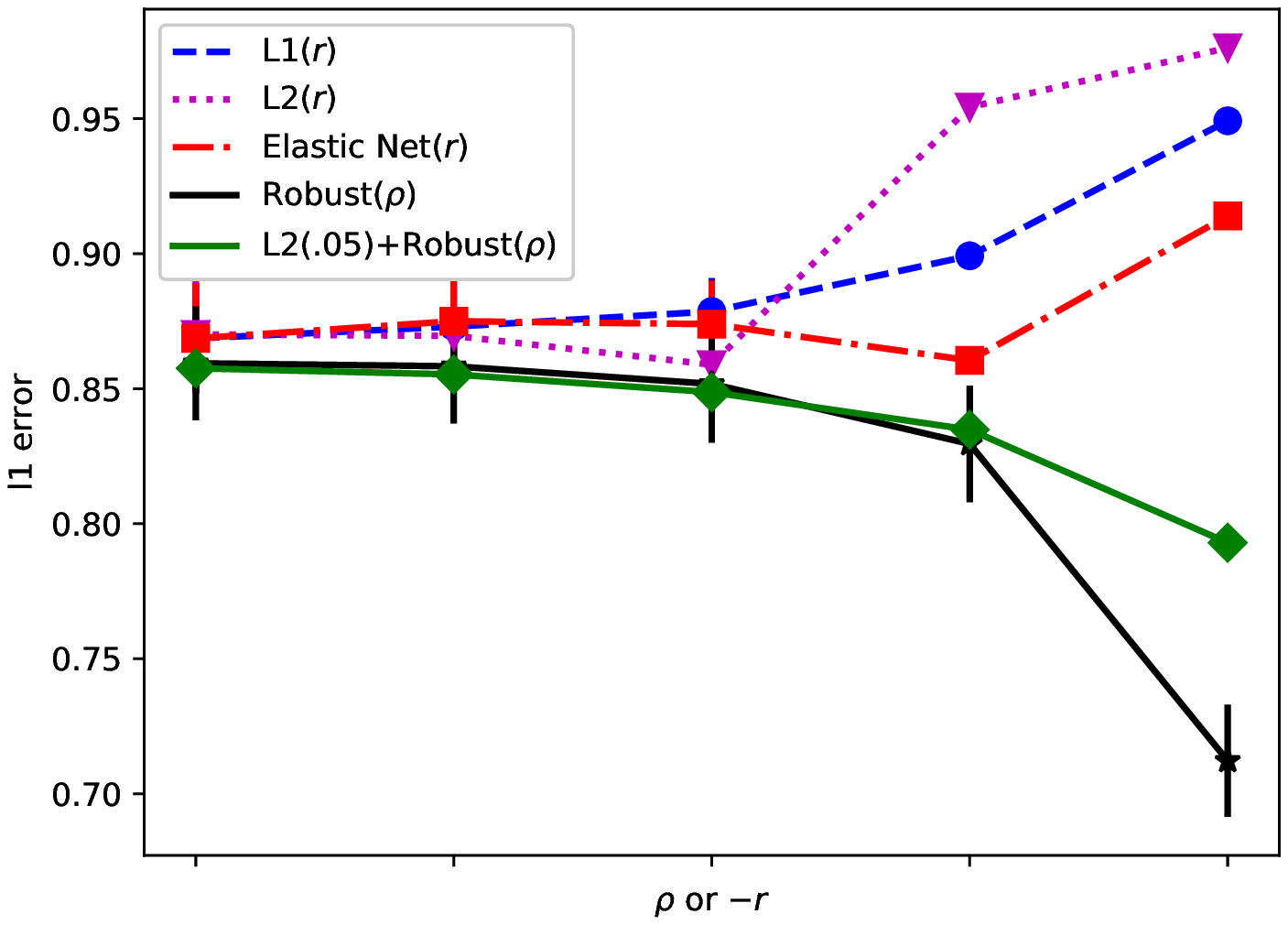}
        \put(45,3){
          \tikz{\path[draw=white,fill=white] (0, 0) rectangle (1cm,.25cm);}}
        \put(8,71){\scriptsize{$\begin{matrix} r_1 = 5 \\ r_2 = 50
            \end{matrix}$}}
        \put(26.5,71){\scriptsize{$\begin{matrix} r_1 = 1 \\ r_2 = 10
            \end{matrix}$}}
        \put(45,71){\scriptsize{$\begin{matrix} r_1 = .5 \\ r_2 = 5
            \end{matrix}$}}
        \put(63.5,71){\scriptsize{$\begin{matrix} r_1 = .1 \\ r_2 = 1
            \end{matrix}$}}
        \put(80,71){\scriptsize{$\begin{matrix} r_1 = .05 \\ r_2 = .5
            \end{matrix}$}}
        \put(11,3){\scriptsize{$\tol = .001$}}
        \put(29.5,3){\scriptsize{$\tol = .01$}}
        \put(48,3){\scriptsize{$\tol = .1$}}
        \put(66.5,3){\scriptsize{$\tol = 1$}}
        \put(83,3){\scriptsize{$\tol = 10$}}
      \end{overpic}
      & 
        \begin{overpic}[width=.5\columnwidth]{
          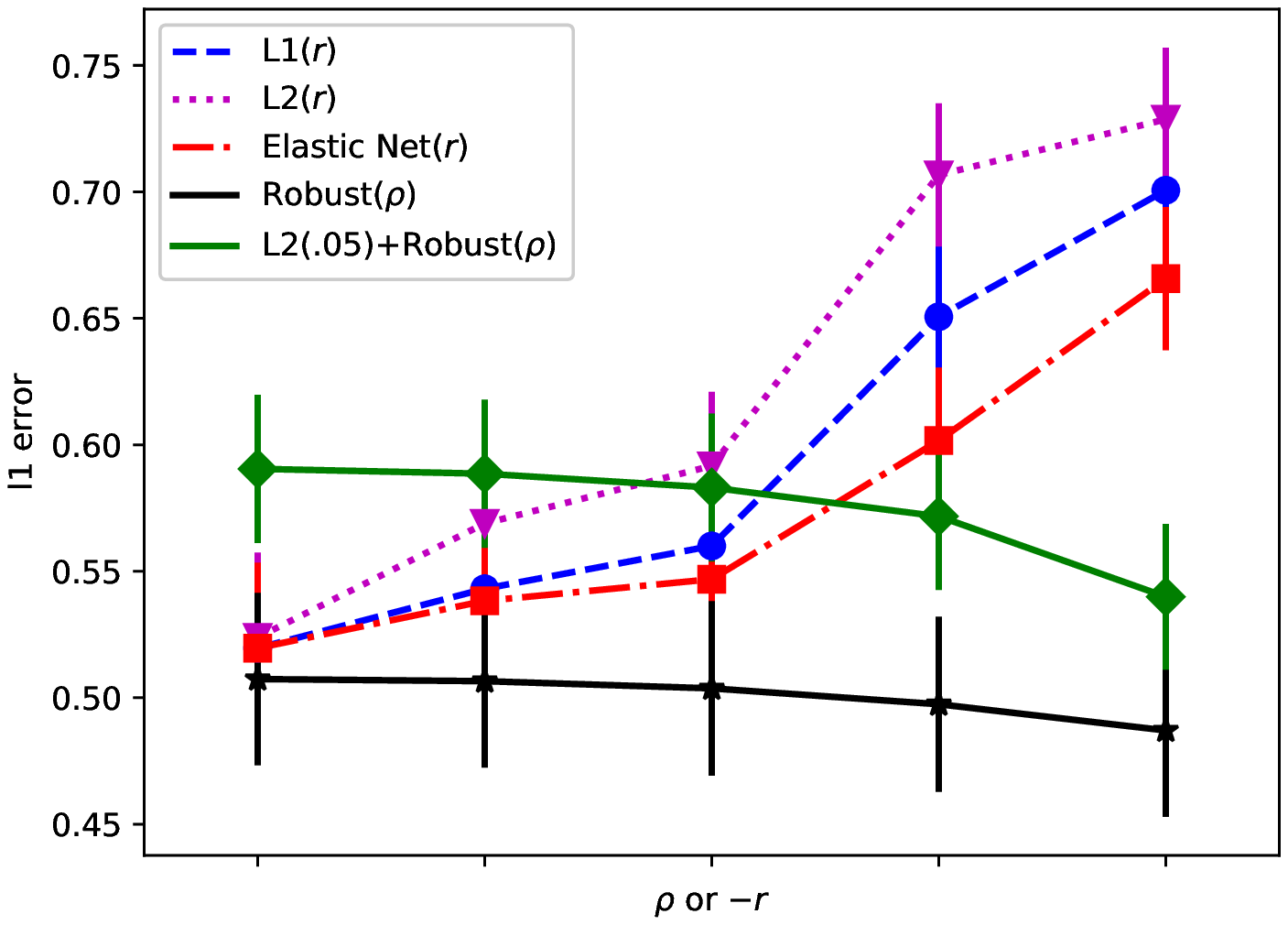}
        \put(45,3){
          \tikz{\path[draw=white,fill=white] (0, 0) rectangle (1cm,.25cm);}}
        \put(8,71){\scriptsize{$\begin{matrix} r_1 = 5 \\ r_2 = 50
            \end{matrix}$}}
        \put(26.5,71){\scriptsize{$\begin{matrix} r_1 = 1 \\ r_2 = 10
            \end{matrix}$}}
        \put(45,71){\scriptsize{$\begin{matrix} r_1 = .5 \\ r_2 = 5
            \end{matrix}$}}
        \put(63.5,71){\scriptsize{$\begin{matrix} r_1 = .1 \\ r_2 = 1
            \end{matrix}$}}
        \put(80,71){\scriptsize{$\begin{matrix} r_1 = .05 \\ r_2 = .5
            \end{matrix}$}}
        \put(11,3){\scriptsize{$\tol = .001$}}
        \put(29.5,3){\scriptsize{$\tol = .01$}}
        \put(48,3){\scriptsize{$\tol = .1$}}
        \put(66.5,3){\scriptsize{$\tol = 1$}}
        \put(83,3){\scriptsize{$\tol = 10$}}
     \end{overpic}
      \vspace{.3cm}
      \\
      (c) Maximal training loss
      & (d) Maximal test loss
    \end{tabular}
    \caption{ \label{fig:crime} Median and maximal loss $|Y - Z^\top \theta|$
      evaluated on training and test datasets. Values of the $x$-axis
      corresponds to different indices for the values of $\rho$ and $r$, so
      that ``$x$-axis = 1'' for the $\ell_1$-constrained problem corresponds
      to $r = 5$, and for the distributionally robust
      method~\eqref{eqn:plug-in} it corresponds to $\tol = .001$. Error bars
      correspond to standard error.}
  \end{center}
  \vspace{-10pt}
\end{figure}
\else
\begin{figure}[t!]
  \begin{center}
    \begin{tabular}{cc}
      \hspace{-.4cm}
      \begin{overpic}[width=.5\columnwidth]{
          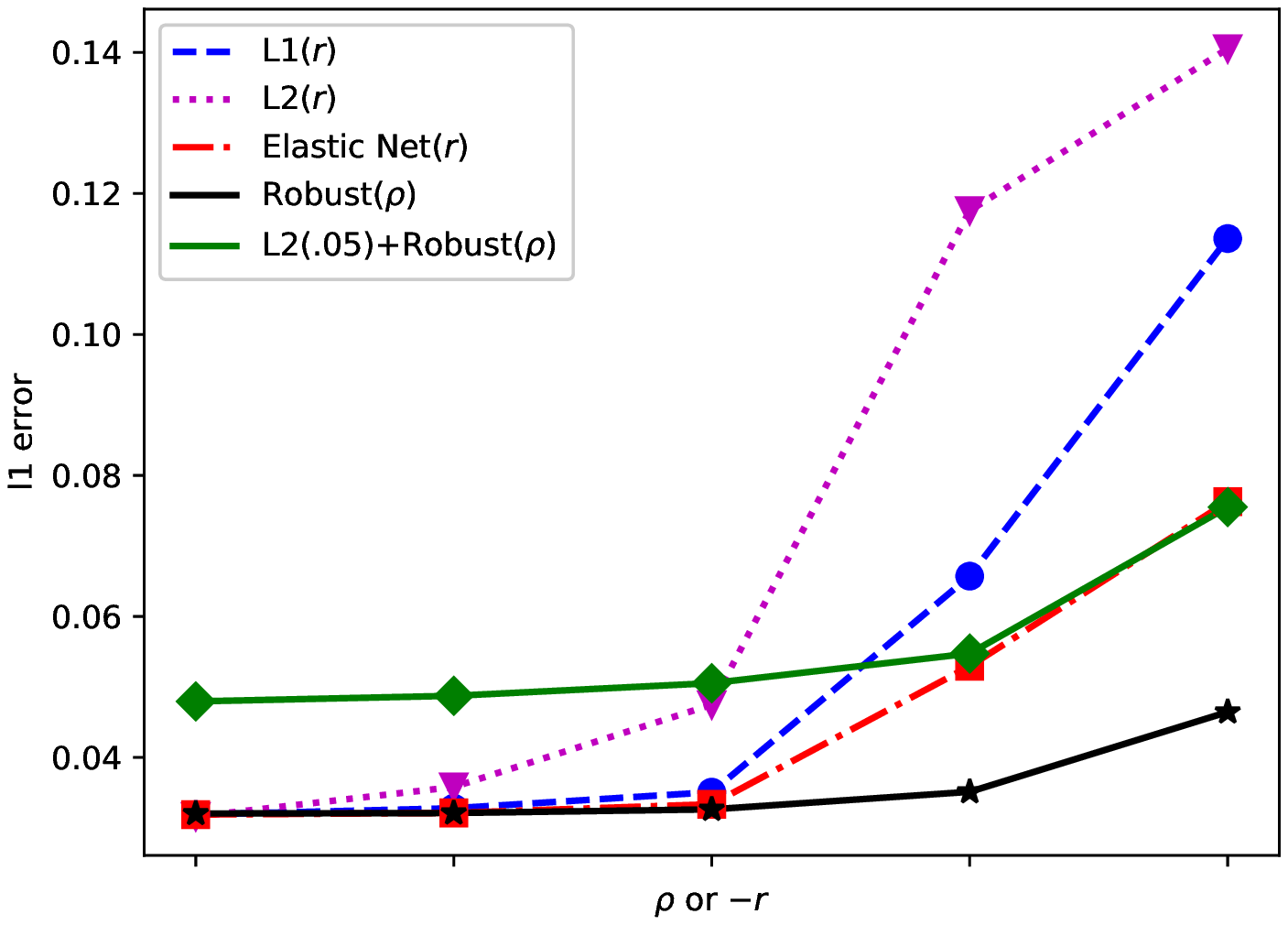}
        \put(45,3){
          \tikz{\path[draw=white,fill=white] (0, 0) rectangle (1cm,.35cm);}}
        \put(11,69){\scriptsize{$\begin{matrix} r_1 = 5 \\ r_2 = 50
            \end{matrix}$}}
        \put(29.5,69){\scriptsize{$\begin{matrix} r_1 = 1 \\ r_2 = 10
            \end{matrix}$}}
        \put(48,69){\scriptsize{$\begin{matrix} r_1 = .5 \\ r_2 = 5
            \end{matrix}$}}
        \put(66.5,69){\scriptsize{$\begin{matrix} r_1 = .1 \\ r_2 = 1
            \end{matrix}$}}
        \put(83,69){\scriptsize{$\begin{matrix} r_1 = .05 \\ r_2 = .5
            \end{matrix}$}}
        \put(11,3){\scriptsize{$\tol = .001$}}
        \put(29.5,3){\scriptsize{$\tol = .01$}}
        \put(48,3){\scriptsize{$\tol = .1$}}
        \put(66.5,3){\scriptsize{$\tol = 1$}}
        \put(83,3){\scriptsize{$\tol = 10$}}
      \end{overpic}
   & 
     \begin{overpic}[width=.5\columnwidth]{
         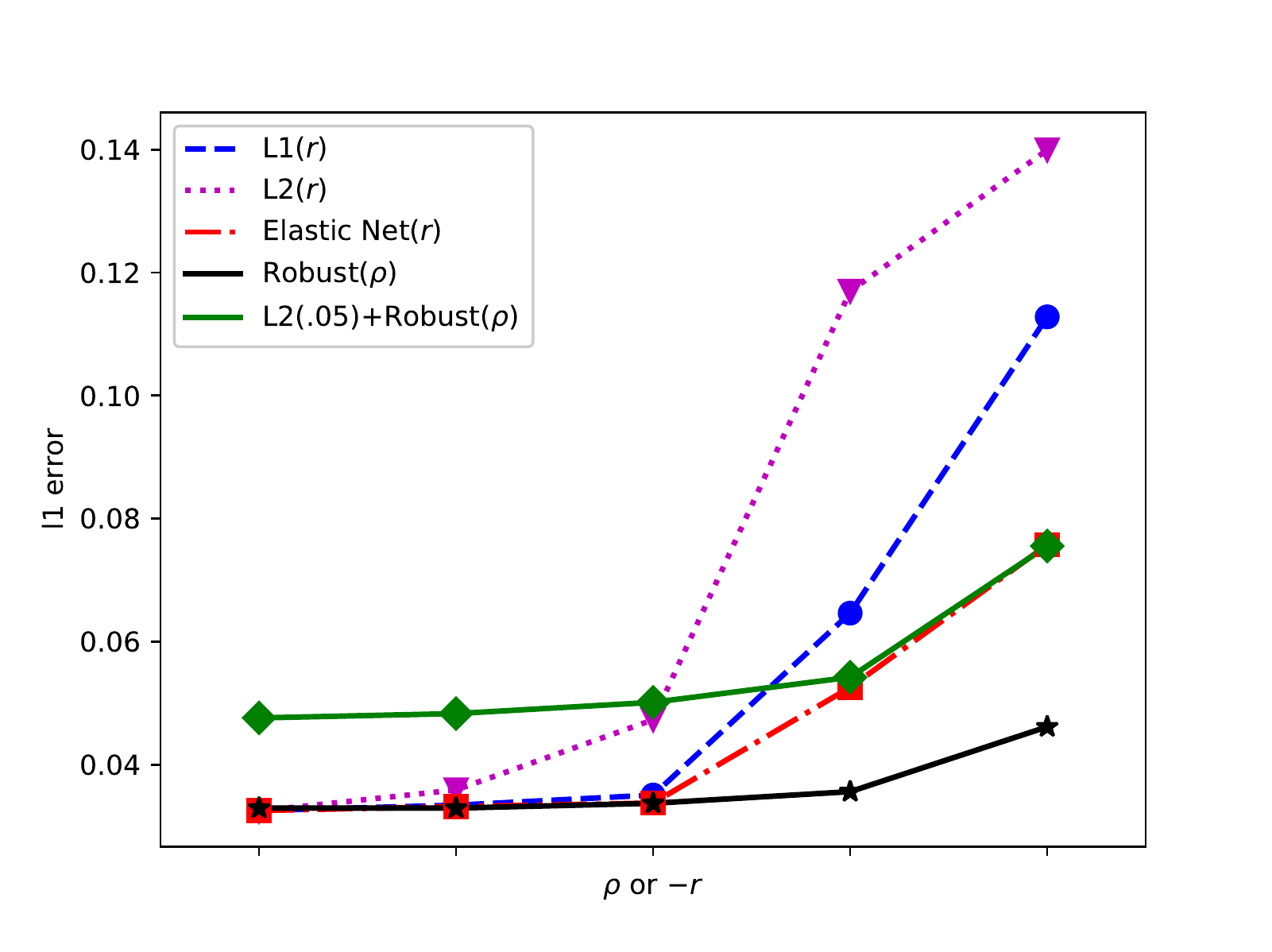}
       \put(45,3){
         \tikz{\path[draw=white,fill=white] (0, 0) rectangle (1cm,.35cm);}}
       \put(9,69){\scriptsize{$\begin{matrix} r_1 = 5 \\ r_2 = 50
           \end{matrix}$}}
       \put(27.5,69){\scriptsize{$\begin{matrix} r_1 = 1 \\ r_2 = 10
           \end{matrix}$}}
       \put(46,69){\scriptsize{$\begin{matrix} r_1 = .5 \\ r_2 = 5
           \end{matrix}$}}
       \put(64.5,69){\scriptsize{$\begin{matrix} r_1 = .1 \\ r_2 = 1
           \end{matrix}$}}
       \put(81,69){\scriptsize{$\begin{matrix} r_1 = .05 \\ r_2 = .5
           \end{matrix}$}}
       \put(9,3){\scriptsize{$\tol = .001$}}
       \put(27.5,3){\scriptsize{$\tol = .01$}}
       \put(46,3){\scriptsize{$\tol = .1$}}
       \put(64.5,3){\scriptsize{$\tol = 1$}}
       \put(81,3){\scriptsize{$\tol = 10$}}
     \end{overpic}
      \\
      (a) Median training loss
      & (b) Median test loss \\
      \hspace{-.4cm}
      \begin{overpic}[width=.5\columnwidth]{
          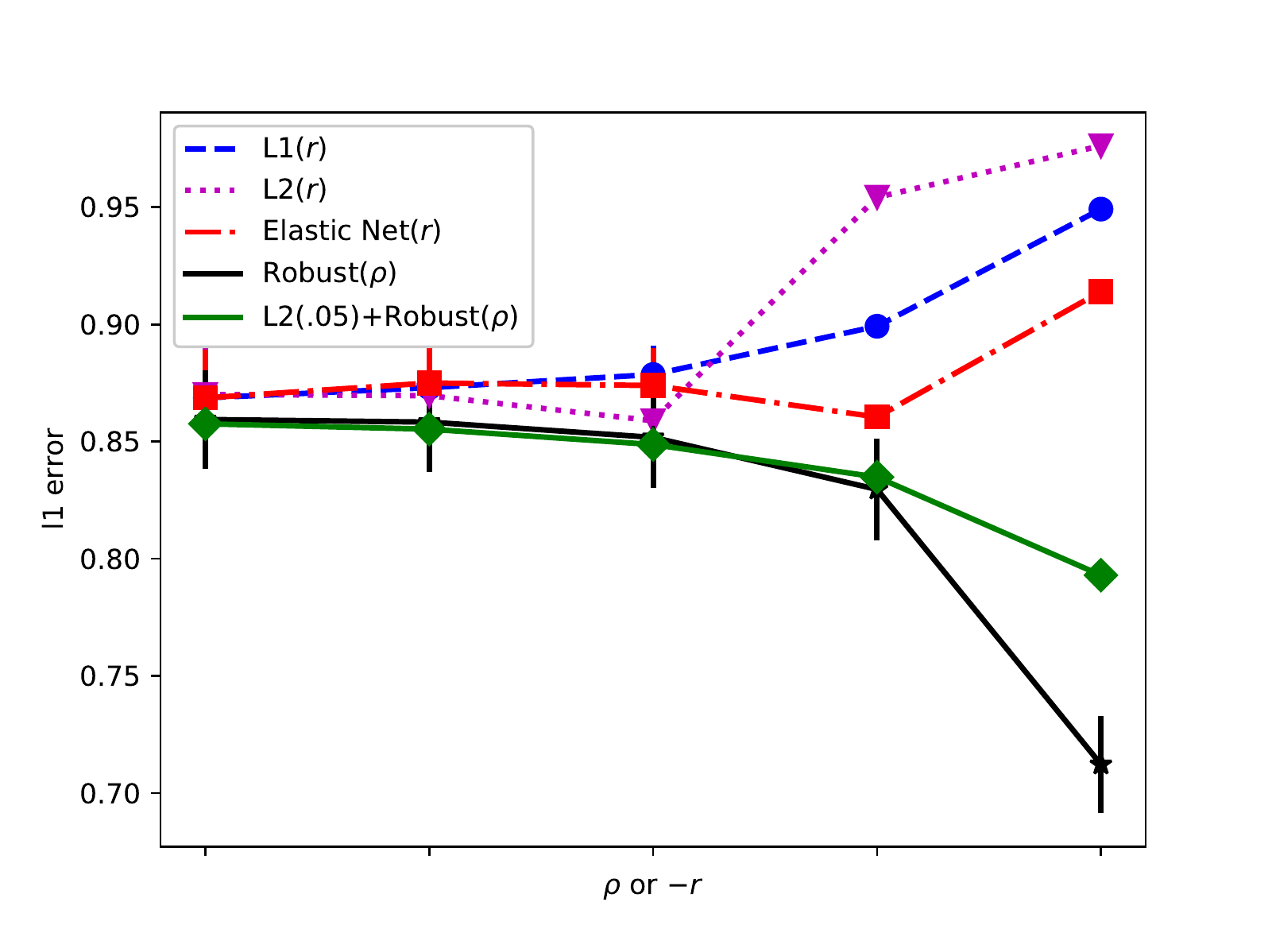}
        \put(45,3){
          \tikz{\path[draw=white,fill=white] (0, 0) rectangle (1cm,.35cm);}}
        \put(11,69){\scriptsize{$\begin{matrix} r_1 = 5 \\ r_2 = 50
            \end{matrix}$}}
        \put(29.5,69){\scriptsize{$\begin{matrix} r_1 = 1 \\ r_2 = 10
            \end{matrix}$}}
        \put(48,69){\scriptsize{$\begin{matrix} r_1 = .5 \\ r_2 = 5
            \end{matrix}$}}
        \put(66.5,69){\scriptsize{$\begin{matrix} r_1 = .1 \\ r_2 = 1
            \end{matrix}$}}
        \put(83,69){\scriptsize{$\begin{matrix} r_1 = .05 \\ r_2 = .5
            \end{matrix}$}}
        \put(11,3){\scriptsize{$\tol = .001$}}
        \put(29.5,3){\scriptsize{$\tol = .01$}}
        \put(48,3){\scriptsize{$\tol = .1$}}
        \put(66.5,3){\scriptsize{$\tol = 1$}}
        \put(83,3){\scriptsize{$\tol = 10$}}
      \end{overpic}
      & 
        \begin{overpic}[width=.5\columnwidth]{
          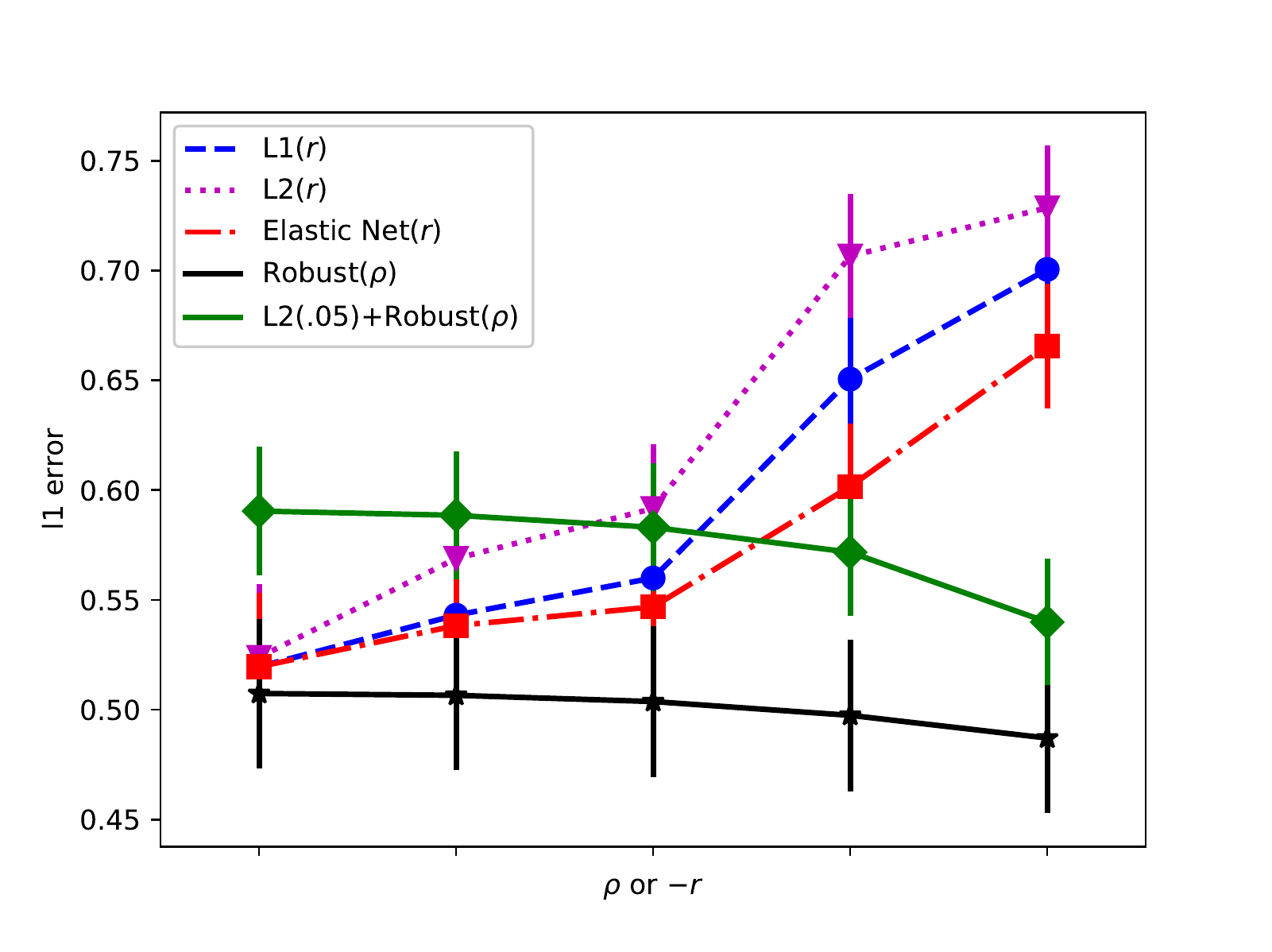}
       \put(45,3){
         \tikz{\path[draw=white,fill=white] (0, 0) rectangle (1cm,.35cm);}}
       \put(9,69){\scriptsize{$\begin{matrix} r_1 = 5 \\ r_2 = 50
           \end{matrix}$}}
       \put(27.5,69){\scriptsize{$\begin{matrix} r_1 = 1 \\ r_2 = 10
           \end{matrix}$}}
       \put(46,69){\scriptsize{$\begin{matrix} r_1 = .5 \\ r_2 = 5
           \end{matrix}$}}
       \put(64.5,69){\scriptsize{$\begin{matrix} r_1 = .1 \\ r_2 = 1
           \end{matrix}$}}
       \put(81,69){\scriptsize{$\begin{matrix} r_1 = .05 \\ r_2 = .5
           \end{matrix}$}}
       \put(9,3){\scriptsize{$\tol = .001$}}
       \put(27.5,3){\scriptsize{$\tol = .01$}}
       \put(46,3){\scriptsize{$\tol = .1$}}
       \put(64.5,3){\scriptsize{$\tol = 1$}}
       \put(81,3){\scriptsize{$\tol = 10$}}
      \end{overpic}
      \\
      (c) Maximal training loss
      & (d) Maximal test loss
    \end{tabular}
    \caption{ \label{fig:crime} Median and maximal loss $|Y - Z^\top \theta|$
      evaluated on training and test datasets. Values of the $x$-axis
      corresponds to different indices for the values of $\rho$ and $r$, so
      that ``$x$-axis = 1'' for the $\ell_1$-constrained problem corresponds
      to $r = 5$, and for the distributionally robust
      method~\eqref{eqn:plug-in} it corresponds to $\tol = .001$. Error bars
      correspond to standard error.}
  \end{center}
\end{figure}
\fi

\subsection{Tail performance in a regression problem}
\label{section:tail-performance}
We consider a linear regression problem using the \texttt{communities and
  crime} dataset~\cite{RedmondBa02,AsuncionNe07}, studying the performance of
distributionally robust methods on tail losses. Given a $122$-dimensional
attribute vector $X$ describing a community, the goal is to predict per
capita violent crimes $Y$ (see~\cite{RedmondBa02}). We use the absolute loss
$\loss(\theta; (x, y)) = |\theta^\top x - y|$ and compare
method~\eqref{eqn:plug-in} with constrained forms of lasso, ridge, and
elastic net regularization~\cite{ZouHa05}, taking constraints
of the form
\begin{equation*}
  \Theta
  = \left\{ \theta \in \R^d: a_1 \lone{\theta} + a_2 \ltwo{\theta} \le r
  \right\}.
\end{equation*}
We vary $a_1$, $a_2$, and $r$: for $\ell_1$-constraints we take $a_1 = 1,
a_2 = 0$ and vary $r_1 \in \{ .05, .1, .5, 1, 5 \}$; for
$\ell_2$-constraints we take $a_1 = 0, a_2 = 1$ and vary $r_2 \in \{ .5, 1,
5, 10, 50\}$; for elastic net we take $a_1 = 1, a_2 = 10$ and set $r = r_1 +
r_2$. We compare these regularizers with the distributionally robust
procedure~\eqref{eqn:plug-in} with $k = 2$, and the same procedure coupled
with the $\ell_2$-constraint ($a_1 = 1, a_2 = 0$) with $r = .05$, where we
vary $\tol \in \{ .001, .01, .1, 1, 10 \}$.

In Figure~\ref{fig:crime}, we plot the quantiles of the training and test
losses with respect to different values of regularization or $\tol$. The
horizontal axis in each figure indexes our choice of regularization value.
We observe that $\robsol$ shows very different behavior than other
regularizers; $\robsol$ attains median losses similar or slightly higher
than the regularized ERM solutions, and achieves much smaller loss on the
tails of the inputs. As $\tol$ grows, the robust solution exhibits
\red{increasing median loss---though slowly---and decreasing maximal loss}.  To
validate our experiments, we made $50$ independent random partitions of our
dataset with $n = 2118$ samples. For each random partition, we divide the
dataset into training set with $n_{\rm train} = 1800$ and a test set with
$n_{\rm test} = 318$. 



\subsection{Fine-grained recognition and challenging sub-groups}
\label{section:fine-grained}


\begin{figure}[t]
  \begin{center}
    \begin{tabular}{cc}
      \hspace{-.4cm}
      \includegraphics[width=.4\columnwidth]{%
        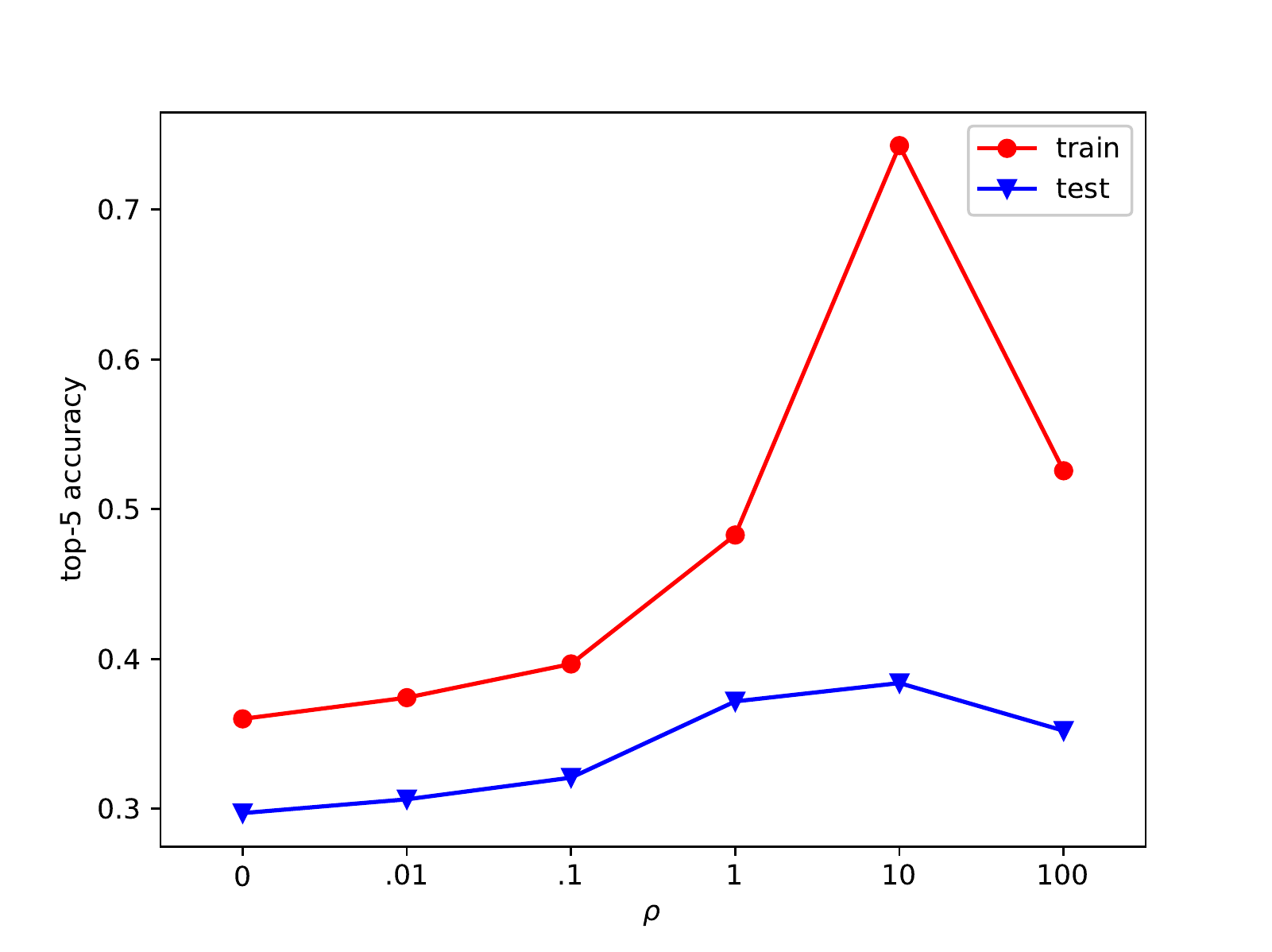} 
      & 
      \includegraphics[width=.4\columnwidth]{%
        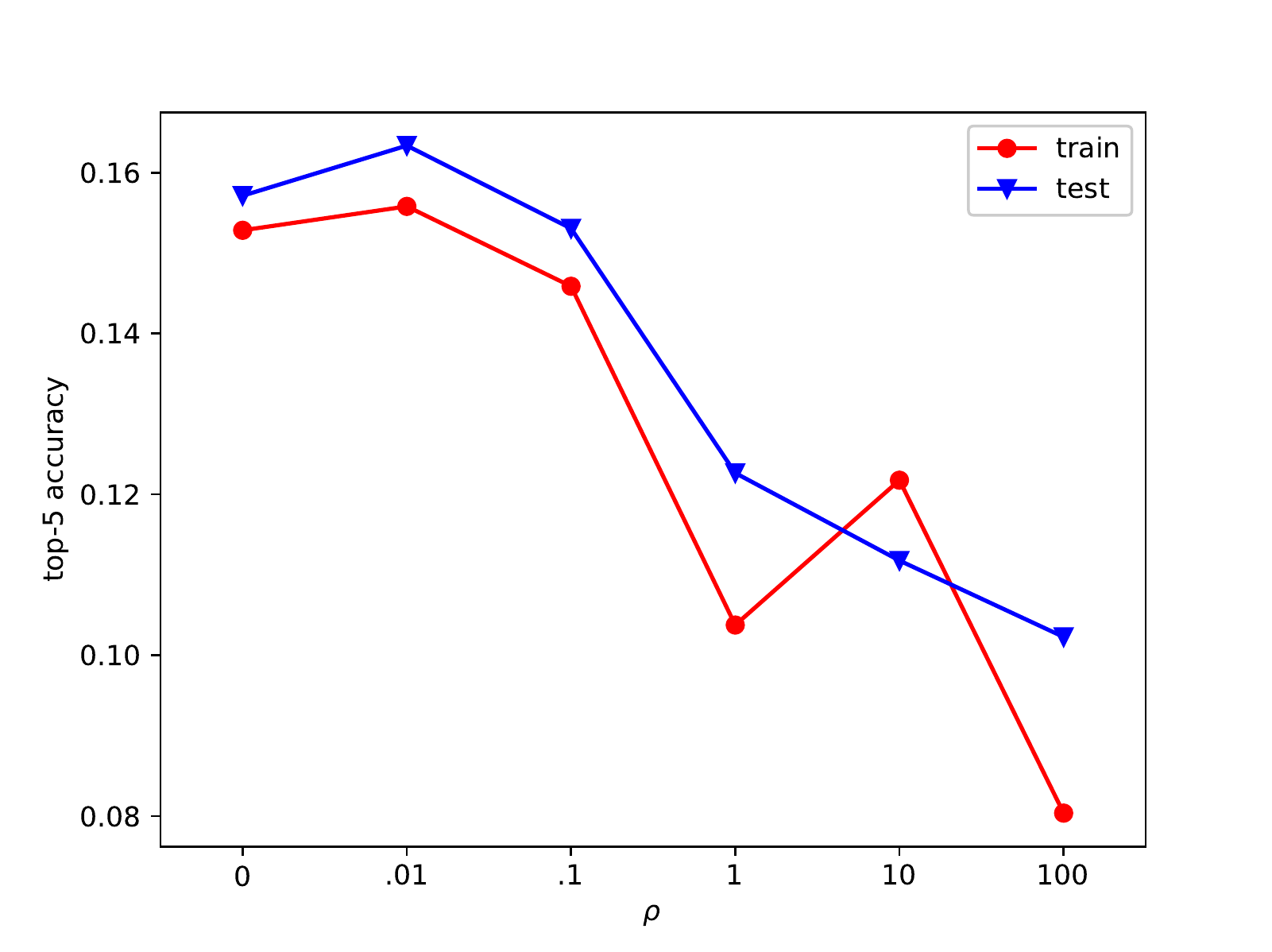}  \\
      \small{(a) Overall top-$5$ accuracy}
      & \small{(b) Standard deviation of top-$5$ accuracy} \\
      \hspace{-.4cm}
      \includegraphics[width=.4\columnwidth]{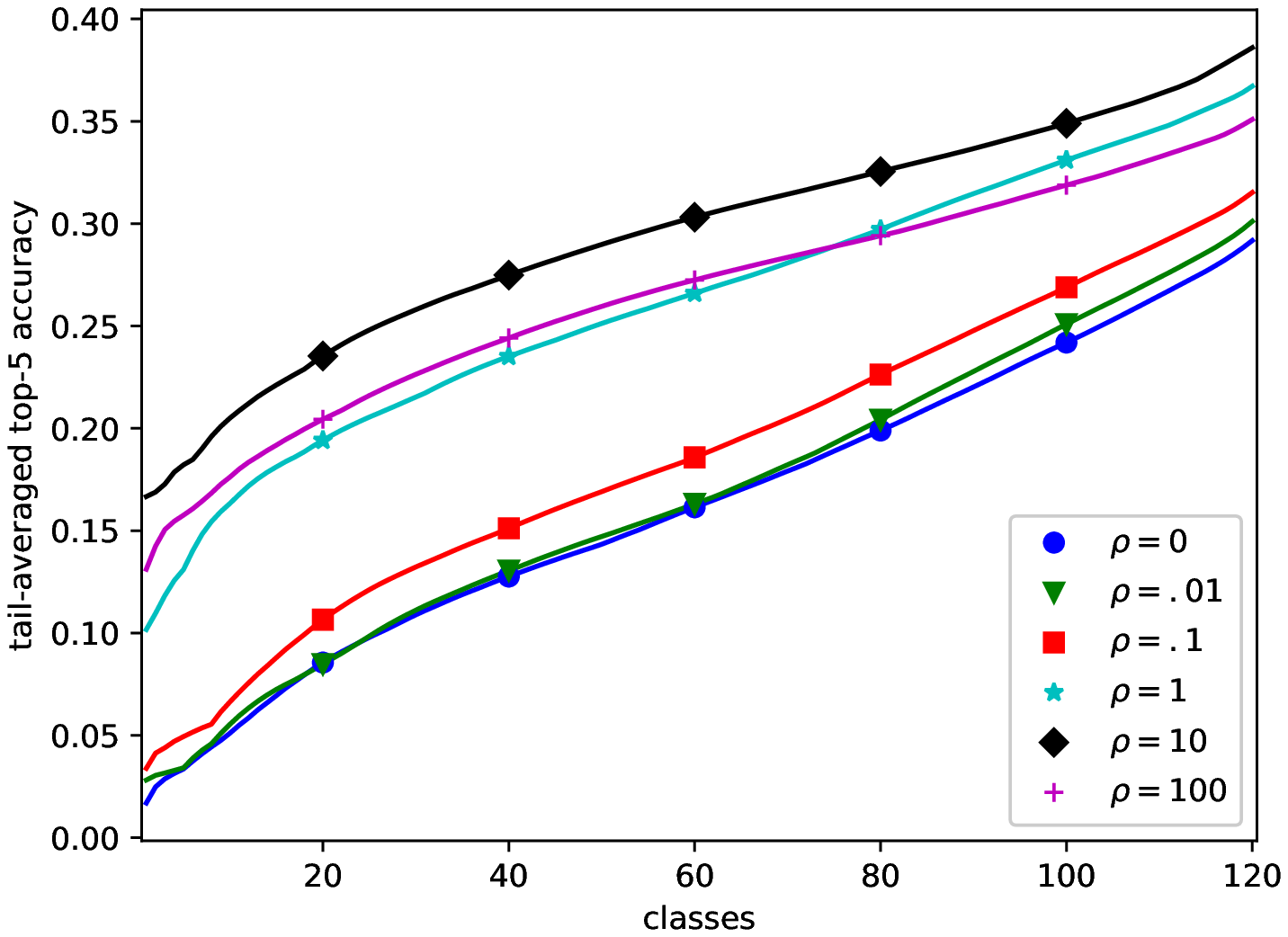} 
      & 
      \includegraphics[width=.4\columnwidth]{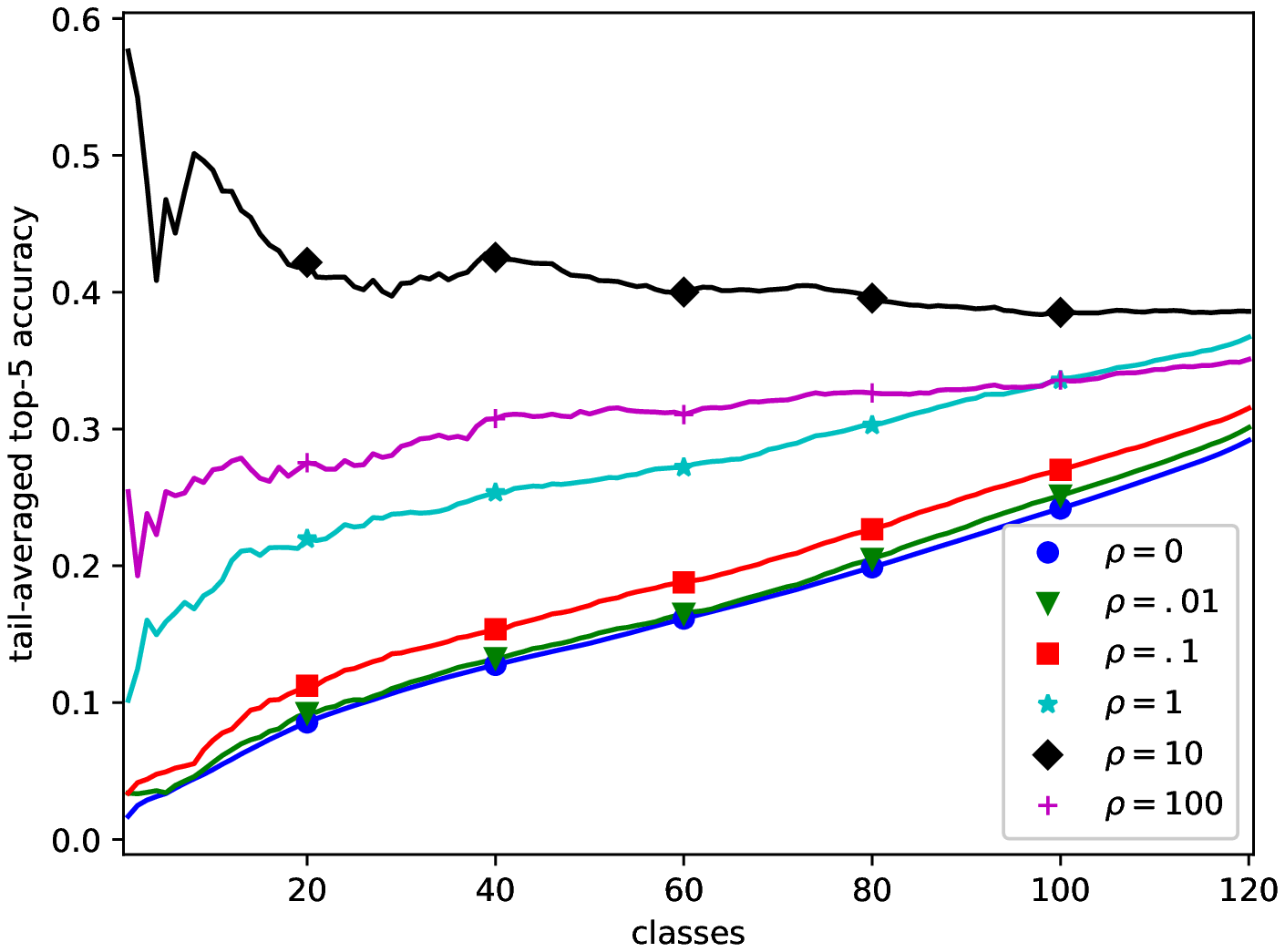}  \\
      \small{(c) Test top-$5$ on worst $c$ classes}
      & \small{(d) Test top-$5$ accuracy on worst ERM classes}
    \end{tabular}
    \caption{ \label{fig:dogs} (a) Top-$5$ error against $\tol$ on train and
      test. (b) Standard deviation of top-$5$ accuracy across $120$
      different classes against $\tol$.  (c) Test top-$5$ accuracy on the
      worst-$c$ classes under each model, i.e.\ $c$ classes with lowest
      accuracy under each model. (d) Test top-$5$ accuracy on the worst-$c$
      classes ordered by accuracy of \emph{ERM} model ($\rho = 0$).}
  \end{center}
  \ifdefined\useaosstyle
  \vspace{-10pt}
  \else
  \vspace{-15pt}
  \fi
\end{figure}

Finally, we consider the fine-grained recognition task of the
\texttt{Stanford Dogs} dataset~\cite{KhoslaJaYaLi11}, where the goal is to
classify an image of a dog into one of $120$ different breeds. There are
20,580 images, $n_{\rm train} =$ 12,000 training examples,
with $100$ training examples for each class. We use the default histogram of
SIFT features in the dataset~\cite{UsuiKo09}, resulting
in vectors $x
\in \R^d$ with $d = $12,000.

We train $120$ one-versus-rest classifiers, one each class, and combine their
predictions by taking the $k$ predictions with largest scores for a given
example $x$. For each binary classification problem, we use the binary
logistic loss, regularized with lasso (in constrained form) so that
  $\Theta_{\rm one-vs-rest}
  = \left\{ \theta \in \R^d: \lone{\theta} \le r \right\}$.
  Thus, for each class $i$, we represent a pair $(x, y)$ by $y = 1$ if $x$ is
  of breed $i$, and $-1$ otherwise, fitting a binary classifier $\theta_i$ for
  each class.  We use $r = 1.0$ for all of our methods based on
  cross-validation for ERM ($\rho = 0$). As we predict using the $m$ highest
  scores, we measure performance with respect to top-$m$ accuracy, which
  counts the number of test examples in which the true label was among these
  $m$ predictions. \red{As $\tol$ grows larger, we expect better performance on
  challenging classes, sacrificing performance on easier classes}, and due to
  uniform performance, for the variance in the class-wise accuracies to be
  smaller, though we do not necessarily expect that average accuracies should
  improve as $\tol$ increases.

  In Figure~\ref{fig:dogs}, we present top-$5$ accuracies; top-1 and top-3
  accuracies are similar. Overall accuracy \emph{improves} moderately as
  $\tol$ grows (Figure~\ref{fig:dogs}(a)), and the \emph{standard deviation}
  of the top-5 accuracy across the classes decreases as $\tol$ increases
  (Figure~\ref{fig:dogs}(b)), consistent with our hypothesis that the robust
  formulations should yield \red{more uniform performance across different
  subpopulations.} In Figure~\ref{fig:dogs}(c), we plot the accuracy averaged
  over $c$-classes that suffer the lowest accuracy under each model, varying
  $c$ on the horizontal axis; the accuracy at $c = 120$ is simply the average
  top-5 accuracy of the models. For $c$ small, meaning for classes on which
  the respective models perform most poorly, we observe that the ensemble of
  one-vs-rest $\robsol$'s outperform the ensemble of ERM solutions
  $\ermsol$'s. In Figure~\ref{fig:dogs}(d), we plot the accuracy averaged over
  the first $c$-classes that have the lowest accuracy under the ERM model.  We
  see that robust solutions $\robsol$ improve performance on classes that ERM
  does poorly on; such tail-performance improves monotonically with $\rho$ up
  to $\rho = 10$; we conjecture the degradation for higher $\tol$ is a
  consequence of overly conservative estimates. Figure~\ref{fig:dogs}(c) shows
  that the gap between the robust classifier performance and non-robust
  classifier goes from .17 vs.\ .03 (hardest class accuracy) to .38 vs.\ .28
  (overall accuracy), so that relative performance gains of the robust
  approach seem largest on the hardest classes. Although it is hard to draw
  conclusions from this experiment due to improved overall performance when
  increasing $\tol$, we conjecture that is due to the regularization effect
  for relatively small values of $\rho$ described by many previous
  authors~\cite{GotohKiLi15, Lam16, DuchiGlNa16, LamZh17, NamkoongDu17}.




%% file: upper.tex
\section{Convergence Guarantees}
\label{section:upper}

Our empirical experiments in the previous section evidence
the potential statistical benefits of the distributionally robust
estimator~\eqref{eqn:plug-in}.  As a consequence, we view it as important to
develop some of its theoretical properties, so we investigate its performance
under a variety of conditions on the $f$-divergence,
providing finite sample convergence guarantees for $f$-divergences with
$f(t) \asymp t^k$ with $k \in (1, \infty)$. Recalling the
definition~\eqref{eqn:cr-risk} of worst-case risk $\risk_k(\theta; P_0)$
for the Cressie-Read
divergences~\eqref{eqn:cressie-read}, we show that the empirical minimizer
$\robsol$ for the plug-in~\eqref{eqn:plug-in} satisfies
$\risk_f(\what{\theta}_n; P_0) -\inf_{\theta \in \Theta} \risk_f(\theta; P_0)
\le C n^{-\frac{1}{k_* \vee 2}}$ with high probability, where
$k_* = \frac{k}{k-1}$ and $C$ is a problem dependent constant. As we show in
Section~\ref{section:lower}, the $n^{-1/(k_* \vee 2)}$ rate is optimal in $n$.
The departure from parametric rates as the uncertainty set becomes large,
meaning $k \downarrow 1$ or $k_* = \frac{k}{k-1} \uparrow \infty$, is a
consequence of the fact that in the worst case, it is challenging to estimate
$L^{k_*}$-norms of random variables $X$ for $k_* > 2$; that is, the minimax
rate for such estimation is $n^{-1/k_*}$ for $k_* > 2$.

Throughout this section, we assume that for any $\param \in \Theta$ and
$x \in \statdomain$, we have $\loss(\param; x) \in [0, \zbound]$ for some
$\zbound \ge 1$, and restrict attention to the Cressie-Read family of
divergences~\eqref{eqn:cressie-read} with $k \in (1, \infty)$. We first show
pointwise concentration of the finite sample objective $\risk_k(\theta; \emp)$
to its population counterpart $\risk_k(\theta; P_0)$; we use convex
concentration inequalities~\cite{BoucheronLuMa13, Talagrand96b} to show
concentration of $\risk_k(\theta; \emp)$ to $\E[\risk_k(\theta; \emp)]$, and
then carefully bound the bias of $\E[\risk_k(\param; \emp)]$ in estimating the
population risk $\risk_k(\param; P_0)$.
\begin{theorem}
  \label{theorem:concentration}
  Assume that $\loss(\param; x) \in [0, \zbound]$ for all $\param \in \Theta$
  and $x \in \statdomain$, and define
  $c_k(\tol) \defeq (k (k - 1) \tol + 1)^{1/k}$.  For a fixed
  $\param \in \Theta$ and $t > 0$, whenever $n \ge k \vee 3$, with probability
  at least $1- 2e^{-t}$
  \begin{align*}
    \left| \risk_k(\param; \emp) - \risk_k(\param; P_0) \right|
    \le 10 n^{-\frac{1}{k_* \vee 2}} c_k(\tol)^2 \zbound \left( \frac{c_k(\tol)}{c_k(\tol)-1} \vee 2\right)
    \left( \frac{1}{k} + 
    \sqrt{ t + 2\log n} \right).
  \end{align*}
\end{theorem}
\noindent
See Section~\ref{section:proof-of-concentration} for the proof. Relaxing the
boundedness assumption $\loss(\theta; x) \in [0, \zbound]$ to sub-Gaussian or
sub-exponential tails, or providing similar finite-sample guarantees for
general $f$-divergences are topics of future research.

Given the pointwise concentration result
(Theorem~\ref{theorem:concentration}), we can use a simple covering argument
to obtain its uniform counterpart. Our uniform guarantees rely on
covering numbers for the model class
$\{\loss(\theta; \cdot): \theta \in \Theta\}$
(e.g.~\cite{VanDerVaartWe96}).  A
collection $v_1, \ldots, v_N$ is an \emph{$\epsilon$-cover} of a set
$V$ in norm
$\norm{\cdot}$ if for each $v \in \mc{V}$, there exists $v_i$ such that
$\norm{v - v_i} \le \epsilon$. The \emph{covering number}
is
\begin{equation*}
  \covnum(V, \epsilon, \norm{\cdot}) \defeq
  \inf\left\{\covnum \in \N \mid \mbox{there~is~an~}
    \epsilon \mbox{-cover~of~} V
    ~ \mbox{with~respect~to~} \norm{\cdot} \right\}.
\end{equation*}
For $\fclass \defeq \left\{ \loss(\theta, \cdot): \theta \in \Theta \right\}$
equipped with $\sup$-norm
$\linfstatnorm{h} \defeq \sup_{x \in \statdomain}|h(x)|$, a covering argument
gives a uniform concentration result, where we use
\begin{equation*}
  \epsilon_{t, n, k}(\tol)
  \defeq n^{-\frac{1}{k_* \vee 2}} c_k(\tol)^2
  \left( \frac{c_k(\tol)}{c_k-1} \vee 2\right)
  \left( \frac{1}{k} + 
  \sqrt{ t + 2\log n} \right).
\end{equation*}
\begin{corollary}
  \label{corollary:concentration-uniform}
  Let $\loss(\param; x) \in [0, \zbound]$ for all $\param \in \Theta$ and
  $x \in \statdomain$. Then for any $t > 0$, whenever $n \ge k \vee 3$, with
  probability at least
  $1- 2 \covnum(\fclass, \frac{\epsilon_{t, n, k}(\tol)}{3},
  \linfstatnorm{\cdot})e^{-t}$
  \begin{align*}
    \sup_{\param \in \Theta}
    \left| \risk_k(\param; \emp) - \risk_k(\param; P_0) \right|
    \le 30 \zbound \epsilon_{t,n,k}(\tol).
  \end{align*}
\end{corollary}
\noindent See Section~\ref{section:proof-of-concentration-uniform} for the
proof. From Corollary~\ref{corollary:concentration-uniform}, we immediately
get below.
\begin{corollary}
  \label{corollary:concentration-optimization}
  Let $\loss(\param; x) \in [0, \zbound]$ for all $\param \in \Theta$ and
  $x \in \statdomain$. Then for any $t > 0$, whenever $n \ge k \vee 3$, with
  probability at least
  $1- 2 \covnum(\fclass, \frac{\epsilon_{t, n}}{3},
  \linfstatnorm{\cdot})e^{-t}$
  \begin{align*}
    \risk_k(\robsol; P_0)
    \le \inf_{\param \in \Theta} \risk_k(\param; P_0)
    + 60 n^{-\frac{1}{k_* \vee 2}} c_k^2 \zbound \left( \frac{c_k}{c_k-1} \vee 2\right)
    \left( \frac{1}{k} + 
    \sqrt{ t + 2\log n} \right).
  \end{align*}
\end{corollary}

As an example, let $\theta \mapsto \loss(\theta; x)$ be $L$-Lipschitz
for all $x \in \statdomain$, with respect to some norm $\norm{\cdot}$ on
$\Theta$. Assuming
$D \defeq \sup_{\theta, \theta' \in \Theta} \norm{\theta - \theta'} < \infty$,
a standard bound~\cite[Chapter 2.7.4]{VanDerVaartWe96} is
\begin{equation*}
  \covnum\left(\fclass, \epsilon, \linfstatnorm{\cdot}\right)
  \le N\left(\Theta, \frac{\epsilon}{L}, \norm{\cdot}\right)
  \le \left( 1+ \frac{ D L}{\epsilon}
  \right)^{d}.
\end{equation*}
If there exists $\theta_0 \in \Theta$ and $\zbound_0> 0$ such that
$|\loss(\theta_0; x)| \le \zbound_0$ for all $x \in \statdomain$, we have
$|\loss(\theta; X)| \le L D + \zbound_0$, and
Corollary~\ref{corollary:concentration-optimization} implies that
\begin{equation*}
  \risk_k(\robsol; P_0)
  \le \inf_{\param \in \Theta} \risk_k(\param; P_0)
  + 60 n^{-\frac{1}{k_* \vee 2}} c_k^2 (L D + \zbound_0) \left( \frac{c_k}{c_k-1} \vee 2\right)
  \left( \frac{1}{k} + 
  \sqrt{ t + 2d \log (2n)} \right)
\end{equation*}
with probability at least $1-2\exp(- t)$.  Replacing covering numbers in the
above guarantees with Rademacher averages or their localized
variants~\cite{BartlettBoMe05} and leveraging Rademacher contraction
inequalities~\cite{LedouxTa91} remain open.



%% file: lower.tex
\section{Lower Bounds}
\label{section:lower}

To complement our uniform upper bounds, we provide minimax lower bounds
showing they are rate optimal, though developing optimal dimension-dependent
bounds remains open. For a collection $\mc{P}$ of distributions and
$f$-divergence $f$, we define the minimax risk
\begin{equation}
  \label{eqn:minimax}
  \minimax_n(\mc{P}, f, \loss)
  \defeq \inf_{\what{\theta}_n}
  \sup_{P_0 \in \mc{P}} ~~ \E_{P_0^n}\left[
    \risk_f\left(\what{\theta}_n(X_1^n); P_0\right)
    - \inf_{\theta \in \Theta} \risk_f\left(\theta; P_0\right)
  \right]
\end{equation}
where the outer infimum is over all $(X_1, \ldots, X_n)$-measurable functions
and the inner supremum is over probability measures in $\mc{P}$, where the
loss is implicit in the risk $\risk_f$.  Whenever $f(t) \lesssim t^k$ as
$t \uparrow \infty$, we show there exist losses for which
$n^{-1/(k_* \vee 2)}$ is a lower bound on the minimax distributionally robust
risk~\eqref{eqn:minimax} where $k_* = k / (k-1)$. Thus there is a necessary
transition from parametric $\sqrt{n}$-type rates to $n^{1/k_*}$ when $k$ is
small---that is, when we seek protection against large distributional shifts.

It is of interest both to \emph{estimate} the value of the risk
$\risk_f$---see the literature on risk measures we reference in
the introduction---and to minimize it. Consequently,
we divide our lower bounds into estimation rates on the
value $\risk_f(\theta; P_0)$ and on the actual minimax
risk~\eqref{eqn:minimax} for the optimization problem~\eqref{eqn:objective},
which build out of these results (Sections~\ref{section:lower-estimation}
and~\ref{section:lower-risk}, respectively). Within each section, we initially
present our results for the Cressie-Read family~\eqref{eqn:cressie-read} with
$k \in (1, \infty)$, allowing explicit constants, then provide lower bounds
for general $f$-divergences using the same techniques.  The rough intuition
for our approach is as follows: we consider Bernoulli variables $Z
\in \{0, \zbound\}$, where the probability that $Z = \zbound$ is
small, though this probability has substantial influence on the risk
$\risk_f$. This highlights the reason for the potentially slow rates of
convergence: one must sometimes observe rarer events to estimate or optimize
the risk $\risk_f$.

\subsection{Lower bounds on estimation of the robust risk value}
\label{section:lower-estimation}

For the rest of this subsection, we fix any $\theta \in \Theta$, and
consider $Z(x) \defeq \loss(\theta; x)$, abusing notation by writing
$\risk_f(Z) \defeq \sup_{\fdiv{Q}{P_0} \le \rho}  \E_Q[Z]$
and $\risk_k(Z) \defeq \risk_f(Z)$ if $f = f_k$ is a Cressie-Read
divergence~\eqref{eqn:cressie-read}. We are interested here
in the minimax error for
estimating the robust risk $\risk_f(Z)$ itself, rather than any
optimization over $\theta$ (justifying our abuse $Z(x) = \loss(\theta; x)$),
studying
\begin{equation}
  \label{eqn:minimax-estimation}
  \minimax_n(\mc{P}, f)
  \defeq \inf_{\what{R}} \sup_{P_0 \in \mc{P}} 
  ~\E_{P_0^n} \left| \what{R}(Z_1^n) - \risk_f(Z) \right|,
\end{equation}
where $Z \sim P_0$ and $Z_1^n \simiid P_0$, and the outer infimum is over
$\what{R}: \{0, \zbound\}^n \to \R$. Throughout this section, we let $\mc{P}$
be the collection of distributions on $Z \in \{0, \zbound\}$
for a fixed $\zbound > 0$.

We first establish a lower bound for estimating
$\risk_k(Z) = \risk_k(\theta; P_0)$ under the Cressie-Read family
$f_k$~\eqref{eqn:cressie-read}; see
Section~\ref{section:proof-of-lower-estimation} for the proof.
Our proof uses Le Cam's method~\cite{Yu97,LeCamYa00}, by noting that if $Z$
takes two values $z_1 < z_2$, then $\risk_k(Z) = z_2$ holds if and only if
$P_0$ places enough mass on $z_2$; we compute the precise threshold at which
the worst-case region contains a point mass, quantifying the fundamental
difficulty in estimating $\risk_k(Z)$.
\begin{theorem}
  \label{theorem:lower-estimation}
  Let $\tol > 0$ be arbitrary but fixed. Define
  $c_k(\tol) \defeq (1 + k(k - 1) \tol)^{1/k}$,
  $p_k \defeq (1 + k(k-1) \tol)^{-1/(k-1)}$, and
  $\beta_k = \frac{k (k - 1) \tol}{2 (1 + k(k-1) \tol)}$.
  Then
  \ifdefined\useaosstyle
    \begin{align*}
         \minimax_n(\mc{P}, f_k)
       & \ge \zbound \max \Bigg\{ \frac{1}{8k_*p_k} \left( 
      \sqrt{\frac{p_k(1-p_k)}{8n}} \wedge \half(1-p_k) 
    \wedge p_k \right), \\
     & \hspace{65pt} \frac{1}{8} \beta_k^{\frac{1}{k}}
       c_k(\tol) \left( \frac{1}{4n} \wedge
      p_k \wedge (1 - (1-\beta_k)^{1-k_*} p_k)
      \right)^{\frac{1}{k_*}} \Bigg\}.
  \end{align*}
  \else
  \begin{align*}
    \minimax_n(\mc{P}, f_k)
     & \ge \zbound \max \Bigg\{ \frac{1}{8k_*p_k} \left( 
      \sqrt{\frac{p_k(1-p_k)}{8n}} \wedge \half(1-p_k) 
    \wedge p_k \right), \\
     & \hspace{65pt} \frac{1}{8} \beta_k^{\frac{1}{k}}
       c_k(\tol) \left( \frac{1}{4n} \wedge
      p_k \wedge (1 - (1-\beta_k)^{1-k_*} p_k)
      \right)^{\frac{1}{k_*}} \Bigg\}.
  \end{align*}
  \fi
\end{theorem}

For general $f$-divergences we can provide a similar result, showing that
the growth of the function $f$ defining the divergence $D_f$ fundamentally
determines worst-case rates of convergence; when $f(t)$ grows slowly as $t
\uparrow \infty$, the robust formulation~\eqref{eqn:objective} is
conservative, so rates of convergence are slower.
First, we give canonical $\Omega(n^{-1/2})$ lower bounds. We assume that $f$
is strictly convex at $t = 1$, meaning that
$f(\lambda t_0 + (1 - \lambda) t_1) < \lambda f(t_0) + (1 - \lambda) f(t_1)$
whenever $t_0 < 1 < t_1$.  To state our results, we define the binary
divergence
\begin{equation*}
  h_f(q; p) \defeq
  p f\left(\frac{q}{p}\right) + (1-p) f\left( \frac{1-q}{1-p} \right).
\end{equation*}
As $f$ is strictly convex at $t = 1$, for
$q \ge p$ the function $q \mapsto h_f(q; p)$ is strictly increasing
on its domain and continuous, so there exists
a unique
\begin{equation}
  \label{eqn:q-p-def}
  q(p) \defeq \sup_{q \ge p}\{ q: h_f(q; p) \le \tol\}.
\end{equation}
(Moreover, $q$ is nondecreasing and concave in $p$, so it is a.e.\
differentiable.) We then have the following $\Omega(n^{-1/2})$ lower bound.
\begin{proposition}
  \label{proposition:lower-general-estimation-sqrt-new}
  Let $f: (0, \infty) \to \R \cup \{+\infty\}$ be strictly convex at $t = 1$.
  Assume there exists $p \in (0, 1)$ such that $f$ is $\mc{C}^1$ in a
  neighborhood of $\frac{q(p)}{p}$ and $\frac{1 - q(p)}{p}$.  Then for any
  such $p$,
  \begin{equation*}
    \liminf_{n \to \infty}
    \sqrt{n}   \minimax_n(\mc{P}, f)
    \ge \zbound \frac{\sqrt{p(1-p)}}{8}
    \frac{-\partial_p h_f(q(p); p)}{\partial_q h_f(q(p); p)} > 0.
  \end{equation*}
\end{proposition}
\noindent
See Section~\ref{section:proof-of-lower-general-estimation-sqrt-new}
for the proof.
The final ratio is positive, as
the (strict) convexity of $f$ and joint convexity of $h_f$
imply
$\partial_q h_f(q(p); p) > 0 \in \partial_q h_f(p; p)$ and
$\partial_p h_f(q(p); p) < 0 \in \partial_p h_f(q(p); q(p))$.


If the asymptotic growth of $f$ is at most $t^k$, we can give an
$\Omega(n^{-1/k_*})$ lower bound, which we prove in  Section~\ref{section:proof-of-lower-general-estimation-kconj}. Letting
$f^{-1}(s) \defeq \inf\{ t \in [0, 1]: f(t) \le s\}$ and $m > 0$,
define
\begin{equation}
  \label{eqn:C-lower-bound}
  C_{f, \tol, m} \defeq \frac{m}{\tol} \left( 1 \wedge
    \left(\frac{\tol}{2m}\right)^{-k_*}
    \left(1-f^{-1}\left(\frac{\tol}{2}\right)\right)^{k_*}
  \right)^{-1}.
\end{equation}
\begin{proposition}
  \label{proposition:lower-general-estimation-kconj}
  Let $m > 0$ and $k \in (1, \infty)$. If $f(t) \le m t^k$ for
  $t \ge \{ (n \vee C_{f, \tol, m}) \tol m^{-1}\}^{\frac{1}{k}}$,
  then
  \begin{equation*}
    \minimax_n(\mc{P}, f)
    \ge \frac{\zbound}{16} \left(\frac{\tol}{m}\right)^{\frac{1}{k}}
    \left( \frac{1}{n \vee C_{f, \tol, m}} \right)^{\frac{1}{k_*}}.
  \end{equation*}
\end{proposition}

\subsection{Lower bounds on optimization}
\label{section:lower-risk}

Our lower bounds on optimization build on those for estimating $\risk_f$. We
consider linear losses, which makes the situation closest to the estimation of
the risk results in the previous section (as we must still estimate $k$th
norms of random variables), providing analogous lower bounds for optimizing
the worst-case objective $\risk_f(\cdot; P_0)$. Using a standard notion of
distance for proving lower bounds in stochastic
optimization~\cite{AgarwalBaRaWa12, Duchi18}, we construct a reduction from
distributionally robust optimization to hypothesis testing.
Throughout, we let $\mc{P}$ be the set of distributions with
$\statval \in [-1, 1]$ almost surely. We begin by considering the lower bound
for the Cressie-Read family~\eqref{eqn:cressie-read} $f_k$, whose proof we
give in Section~\ref{section:proof-of-lower-optimization}.
\ifdefined\useaosstyle \vspace{-3pt} \fi
\begin{theorem}
  \label{theorem:lower-optimization}
  Let $\loss(\param; \statval) = \param \statval$ where
  $\theta \in \paramdomain = [-\zbound, \zbound]$. Define
  $c_k(\tol) \defeq (1 + k(k - 1) \tol)^{1/k}$,
  $p_k \defeq (1 + k(k-1) \tol)^{-1/(k-1)}$, and
  $\beta_k = \frac{k (k - 1) \tol}{2 (1 + k(k-1) \tol)}$.  Then
  \begin{align*}
    \minimax_n(\mc{P}, f_k, \loss)
    & \ge  \zbound \max \bigg\{
                \frac{1}{16k_*p_k} \left( 
                \sqrt{\frac{p_k(1-p_k)}{n}} \wedge \half(1-p_k) 
      \wedge (1 - 2p_k) \wedge p_k \right), \\
    & \qquad \qquad
      \frac{1}{16} \beta_k^{\frac{1}{k}} c_k(\tol) \left( \frac{1}{4n} 
    \wedge  p_k \wedge (1-p_k)
    \wedge (1 - (1-\beta_k)^{1-k_*} p_k) \right)^{\frac{1}{k_*}}
    \bigg\}.
  \end{align*}
\end{theorem}
  \ifdefined\useaosstyle
\vspace{-3pt}
\fi

For general $f$-divergences, we can show a similar standard $\Omega(n^{-1/2})$
lower bound for optimization. We defer the proof of this result to
Section~\ref{section:proof-of-lower-general-optimization-sqrt-new}.
\ifdefined\useaosstyle
\vspace{-3pt}
\fi
\begin{proposition}
  \label{proposition:lower-general-optimization-sqrt-new}
  Let $\loss(\param; \statval) = \param \statval$ where
  $\theta \in \paramdomain = [-\zbound, \zbound]$ and $\statrv \in [-1,
  1]$. If the conditions on $f$ of
  Proposition~\ref{proposition:lower-general-estimation-sqrt-new} hold,
  \begin{equation*}
    \liminf_{n \to \infty} \sqrt{n} \minimax_n(\mc{P}, f, \loss)
    \ge \zbound \frac{\sqrt{p(1-p)}}{16q(p)}
    \frac{-\partial_p h_f(q(p); p)}{\partial_q h_f(q(p); p)} > 0.
  \end{equation*}
\end{proposition}
\ifdefined\useaosstyle
\vspace{-3pt}
\fi


For $f$-divergences with $f(t) = O(t^k)$ as $t \to \infty$, we can again prove
a $\Omega(n^{-1/k_*})$ lower bound on optimizing $\risk_f(\cdot; P_0)$.
Recalling the definition~\eqref{eqn:C-lower-bound} of $C_{f, \tol, m}$, we
obtain the following result, whose proof we give in
Section~\ref{section:proof-of-lower-general-optimization-kconj}.
\ifdefined\useaosstyle
\vspace{-15pt}
\fi
\begin{proposition}
  \label{proposition:lower-general-optimization-kconj}
  Let $\loss(\param; \statval) = \param \statval$ where
  $\theta \in \paramdomain = [-\zbound, \zbound]$ and
  $\statrv \in [-1, 1]$. If the conditions on $f$ of
  Proposition~\ref{proposition:lower-general-estimation-kconj} hold,
  \begin{align*}
    \minimax_n(\mc{P}, f, \loss)
    & \ge \frac{\zbound}{16} \left(\frac{\tol}{m}\right)^{\frac{1}{k}}
    \left\{ \left( \frac{1}{n \vee C_{f, \tol, m}} \right)^{\frac{1}{k_*}}
    \wedge \left( \frac{\tol}{2m} \right)^{\frac{1}{k_*}}
    \left( \left( \frac{2}{3} \right)^{k-1} \wedge
    \left(\frac{1}{2}\right)^{\frac{1}{k_*}} \frac{2m}{\tol}
    \right)
    \right\}.
  \end{align*}
\end{proposition}
In terms of rates in $n$, there is a tradeoff between convergence rates and
robustness, as measured by the asymptotic growth of the function $f$ defining
the robustness set $\{P : \fdivs{P}{P_0} \le \tol\}$. In this sense, our
finite sample convergence guarantees of Section~\ref{section:upper} are
sharp. All results in this section can be stated in a probabilistic form that
matches our high probability guarantees in the previous section; see the
remark in the beginning of Section~\ref{section:proof-lower}.


%% file: asymptotics.tex
\section{Asymptotics}
\label{section:asymptotics}

\newcommand{\hingedloss}{\hinge{\loss(\param; X) - \eta}}
\newcommand{\hingedlossopt}{B}
\newcommand{\commonfactor}{\frac{c_k}{(k-1)}
  \left( \E\hingedloss^{k_*} \right)^{-\frac{1}{k} - 1}}
\newcommand{\commonfactoremp}{\frac{c_k}{(k-1)} \left(
    \E_{\emp}\hingedloss^{k_*} \right)^{-\frac{1}{k} - 1}}
\newcommand{\gradloss}{\nabla \loss(\param; X)}

In the previous two sections, we studied convergence properties for the robust
formulation~\eqref{eqn:objective} that hold uniformly over collections of data
generating distributions $P_0$, showing that robustness can incur nontrivial
statistical cost.  In this section, by contrast, we turn to pointwise
asymptotic properties of the empirical plug-in~\eqref{eqn:plug-in}, applying
to a \emph{fixed} distribution $P_0$. This allows two contributions. First, we
prove a general consistency result for convex losses. Second, while the
minimax convergence rates in the previous section exhibit a departure from
classical parametric rates, we show that under appropriate regularity
conditions the typical $\sqrt{n}$-rates of convergence and asymptotic
normality guarantees are possible.

\subsection{Consistency}
\label{section:consistency}


In this section, we give a general set of convergence results, relying on the
powerful theory of epi-convergence~\cite{RockafellarWe98, KingWe91}.  Our
first results shows that $\risk_f(\theta; \emp)$ is pointwise consistent
for its population counterpart $\risk_f(\theta; P_0)$. See
Section~\ref{section:proof-of-consistency-estimation} for the proof.
\begin{proposition}
  \label{proposition:consistency-estimation}
  Let $f$ be finite on $(t_0, \infty)$ for some $t_0 < 1$. For any
  $\param \in \Theta$, if $\E[f^*(|\loss(\param; \statrv)|)] < \infty$ then
  $\risk_f(\param; \emp) \cas \risk_f(\param; P_0) < \infty$.
\end{proposition}



We now provide sufficient conditions for parameter consistency in the
distributionally robust estimation problem~\eqref{eqn:plug-in}. The main
assumption is that the loss functions are closed and the non-robust
population risk is coercive. (Weaker sufficient conditions are possible, but
in our view, a bit esoteric.)
\begin{assumption}[Coercivity]
  \label{assumption:coercivity}
  For each $\statval \in \statdomain$, the function $\theta \mapsto
  \loss(\theta; \statval)$ is closed and convex, and $\E_{P_0}[\loss(\param;
    \statrv)] + \bindic{\param \in \Theta}$ is coercive.
\end{assumption}
\noindent It is possible to replace the convexity assumption with a
Glivenko-Cantelli property on the collection $\{f^*(\loss(\param;
\cdot))\}_{\param \in \Theta}$; for example, if $\param \mapsto
\loss(\param; X)$ is continuous and $\Theta$ is compact, then a similar
consistency result holds,
though computation of the plug-in~\eqref{eqn:plug-in} may be
difficult. Coercivity guarantees the existence and
compactness of the set of optima for $\risk_f(\theta; P_0)$.


Define the \emph{inclusion distance}, or the \emph{deviation}, from a set $A$
to $B$ as
\begin{equation*}
  \dinclude(A, B) \defeq \sup_{y \in A} \dist(y, B)
  = \inf_\epsilon \left\{\epsilon \ge 0
    : A \subset \{y : \dist(y, B) \le \epsilon \} \right\}.
\end{equation*}
This is an one-sided notion of the Hausdorff distance
$d_H(A, B) = \max\{ \dinclude(A, B), \dinclude(B, A)\}$.  For any
$\varepsilon \ge 0$ and distribution $P$, define the set of
$\varepsilon$-approximate minimizers
\begin{equation*}
  S_P(\Theta, \varepsilon) \defeq
  \left\{\param \in \Theta \mid \risk_f(\param; P)
  \le \inf_{\param \in \Theta} \risk_f(\param; P) + \varepsilon
  \right\},
\end{equation*}
where we let $S_P(\Theta) = S_P(\Theta, 0)$ for shorthand. The following
consistency result shows that approximate empirical optimizers are eventually
nearly in the population optima $S_{P_0}(\Theta)$; we provide its proof in
Section~\ref{section:proof-of-consistency-soln}.
\begin{proposition}
  \label{proposition:consistency-soln}
  Let $f$ be finite on $(t_0, \infty)$ for some $t_0 < 1$, and assume
  $\E[f^*(|\loss(\theta; \statrv)|)] < \infty$ on a neighborhood of
  $S_{P_0}(\Theta)$. Under Assumption \ref{assumption:coercivity},
  \begin{equation*}
    \inf_{\param \in
      \Theta} \risk_f(\param; \emp) \cas \inf_{\param \in \Theta}
    \risk_f(\param; P_0),
  \end{equation*}
  and for any sequence $\varepsilon_n \downarrow 0$,
  with probability $1$ we have $S_{\emp}(\Theta, \varepsilon_n) \neq
  \varnothing$ eventually and
    $\dinclude\left(S_{\emp}(\Theta, \varepsilon_n),
    S_{P_0}(\Theta)\right) \to 0$.
\end{proposition}


%% file: asymptotics-limit-lam.tex
\subsection{Asymptotic normality}
\label{section:limit-theorems}

\newcommand{\hingedlossindic}{}
\newcommand{\hingedlossoptindic}{}

The worst-case minimax results are sometimes pessimistic, so we provide a
central limit result for the empirical optimizer
$\what{\param}_n \in \argmin_{\param \in \R^d} \risk(\param; \emp)$ to the
population optimizer
$\theta\opt = \argmin_{\param \in \R^d} \risk(\param; \emp)$ under appropriate
smoothness conditions on the risk. Given that in the general formulation of
our problem, the supremum over distributions $P$ near $P_0$ act as nuisance
parameters, it seems challenging to give the most generic conditions under
which asymptotic normality of $\what{\theta}_n$ should hold. Accordingly, we
assume simpler conditions that allow an essentially classical treatment with a
brief proof, based on the dual formulation~\eqref{eqn:dual}.

Throughout this section, we assume that the population optimizer
$\theta\opt = \argmin_{\param \in \R^d} \risk(\param; \emp)$ is unique. We
begin with a smoothness assumption.
\begin{assumption}[Smoothness and growth]
  \label{assumption:smoothness}
  For some $k > 1$, the function $f$ satisfies
  $\liminf_{t \to \infty} f(t) / t^k > 0$. 
  There exists a neighborhood $U$ of $\theta\opt$ s.t.
  \begin{enumerate}[1.]
  \item \label{item:lipschitz-loss} There exists $L : \statdomain \to \R_+$
    such that
    $|\loss(\param_0; \statval) - \loss(\param_1; \statval)| \le L(\statval)
    \ltwo{\param_0 - \param_1}$ for all $\param_i \in U$, where
    $\E[L(\statrv)^{2 k_*}] < \infty$ (we again use $k_* = \frac{k}{k-1}$).
  \item $\E[|\loss(\theta\opt; \statrv)|^{2 k_*}] < \infty$, and
the function $\theta \mapsto \loss(\theta; \statval)$ is
    differentiable on $U$.
  \end{enumerate}
\end{assumption}

Recalling the dual~\eqref{eqn:dual}, for shorthand
define
\begin{equation*}
  g_P(\theta, \lambda, \eta)
  \defeq \lambda \E_P\left[f^*\left(\frac{\loss(\theta; \statrv) - \eta}{
      \lambda}\right)\right] + \tol \lambda + \eta.
\end{equation*}
\begin{assumption}[Strong identifiability]
  \label{assumption:strong-identifiability}
  The objective $g_{P_0}$ is $\mc{C}^2$ near
  $(\theta\opt, \lambda\opt, \eta\opt) = \argmin
  g_{P_0}(\theta, \lambda, \eta)$ with positive definite
  Hessian, and
  $P_0(\loss(\theta\opt; \statrv) - \eta\opt > 0) > 0$.
\end{assumption}
\noindent The second condition of
Assumption~\ref{assumption:strong-identifiability} guarantees
$\lambda\opt > 0$. 
For Cressie-Read divergences~\eqref{eqn:cressie-read}, a sufficient
condition for uniqueness of $(\eta\opt, \lambda\opt)$ follows.
\begin{lemma}
  \label{lemma:uniqueness-eta-per-param}
  Let $f$ be the Cressie-Read divergence~\eqref{eqn:cressie-read} with
  parameter $k \in (1, \infty)$, and $\theta_0 \in \Theta$.  If
  $\loss(\theta_0; \statrv)$ is non-constant under $P$ and
  $\E_P[|\loss(\theta; \statrv)|^{k_*}] < \infty$ near $\theta_0$, then
  $(\lambda_0, \eta_0) = \argmin_{\lambda \ge 0, \eta} g_{P_0}(\theta_0,
  \lambda, \eta)$ is unique.
\end{lemma}
\noindent
See Appendix~\ref{sec:proof-uniqueness-eta-per-param} for a proof.  Sufficient
conditions for differentiability are similar to the classical conditions for
asymptotic normality of quantile estimators~\cite{VanDerVaart98}; for example,
if $\loss(\cdot; \statrv)$ is $\mc{C}^2$ near some $\theta_0$ and
$P(\loss(\theta; \statrv) = \eta) = 0$ for $\theta, \eta$ near
$\theta_0, \eta_0$, then the dual formulation $g_{P_0}$ is $\mc{C}^2$ in a
neighborhood of $(\theta_0, \eta_0, \lambda_0)$ whenever $\lambda_0 > 0$. With
this brief discussion, we now provide an asymptotic normality result.
\begin{theorem}
  \label{theorem:asymptotics-param}
  Let Assumptions~\ref{assumption:smoothness}
  and~\ref{assumption:strong-identifiability} hold. Let $\what{\param}_n$ be
  any sequence of approximate optimizers to the empirical plug-in satisfying
  $\risk_f(\what{\param}_n; \emp) \le \inf_{\param} \risk_f(\param; \emp) +
  o_P(1/n)$. Then
  \begin{equation}
    \label{eqn:asymptotics-param}
    \sqrt{n} \left( \what{\param}_n - \param\opt\right)
    \cd \normal\left(0, V
    \cov\left({f^*}'\left(\frac{\loss(\theta\opt;\statrv) - \eta\opt}{
      \lambda\opt}\right) \nabla \loss(\theta\opt; \statrv)\right)
    V\right)
  \end{equation}
  where $V$ is the first $d$-by-$d$ block of
  $\left( \nabla^2 g_{P_0}(\param\opt, \lambda\opt, \eta\opt)\right)^{-1} \in
  \R^{(d+2) \times (d+2)}$.
\end{theorem}
\noindent See Section~\ref{section:proof-of-asymptotics-param} for the
proof. Under the same assumptions, it is straightforward to see that plug-in
estimators for $V$ and
$\cov({f^*}'(\frac{\loss(\theta\opt;\statrv) - \eta\opt}{
      \lambda\opt}) \nabla \loss(\theta\opt; \statrv))$ are
consistent. Combining these estimators with
Theorem~\ref{theorem:asymptotics-param} gives an asymptotically
pivotal confidence region for $\param\opt$ by Slutsky's lemmas.


We can relax the assumption that $\nabla^2 g_{P_0}(\param\opt, \lambda\opt,
\eta\opt) \succ 0$ in Assumption~\ref{assumption:strong-identifiability}
to positive definiteness of
the Hessian of the map
$(\eta, \theta) \mapsto c_k (\E_{P_0}[\hinge{\loss(\theta; X) -
  \eta}^{k_*}])^\frac{1}{k_*} + \eta$ at $(\theta\opt, \eta\opt)$, which is
the dual objective $g_k$ with $\lambda$ minimized out. We omit the proof with
this relaxed condition for brevity, as it is quite involved. Letting
$\hingedlossopt = \hinge{\loss(\theta\opt; X) - \eta\opt}$, under
Assumption~\ref{assumption:smoothness} and the randomness conditions of
Lemma~\ref{lemma:uniqueness-eta-per-param}, this relaxed condition holds if
\begin{equation}
  \begin{split}
  & (k-1) \E \hingedlossopt^{k_*-2}
    \left( \E \hingedlossopt^{k_*} \E \hingedlossopt^{k_*-2}
    - (\E \hingedlossopt^{k_*-1})^2 \right) 
    \E[ \hingedlossopt^{k_*-1} \nabla^2 \loss(\param\opt; X)] \\
  & \hspace{20pt}
    - \left( \E \hingedlossopt^{k_*-1} \right)^2
    \E[\hingedlossopt^{k_*-2} 
    \nabla \loss(\param\opt; X)]
    \E[\hingedlossopt^{k_*-2} 
    \nabla \loss(\param\opt; X)]^{\top}
    \succ 0,
  \end{split}
    \label{eqn:hessian-pd}
\end{equation}
and $k \in (1, 2)$. For $k = 2$, the relaxed condition holds if in addition to
the bound~\eqref{eqn:hessian-pd}, there is a neighborhood of
$(\theta\opt, \eta\opt)$ such that $\P(\loss(\theta; \statrv) = \eta) = 0$.
Assumption~\ref{assumption:strong-identifiability} also requires
identifiability of nuisance variables $\lambda\opt, \eta\opt$.  Whether
directly analyzing the primal formulation~\eqref{eqn:objective}---rather than
our proof via the dual~\eqref{eqn:dual}---can relax this
assumption remains open.


%% file: discussion.tex
\section{Discussion and further work}
\label{section:discussion}


We have presented a collection of statistical problems that arise out of a
distributionally robust formulation of M-estimation, whose purpose is to
obtain uniformly small loss and protect against rare but large losses. While
our results give convergence guarantees, and our experimental results suggest
the potential of these approaches in a number of prediction problems, numerous
questions remain.

In our view, the most important limitation is guidance in the choices of the
robustness set, that is, $\{Q : \fdiv{Q}{P_0} \le \tol\}$. The analytic
consequences of our choices are nice in that they allow explicit dual
calculations and algorithmic development; in the case in which the radius
$\tol$ is instead shrinking with as $\tol/n$, asymptotic and non-asymptotic
considerations~\cite{NamkoongDu17, DuchiGlNa16, Ben-TalHeWaMeRe13, Lam16,
  LamZh17} show that the robustness provides a type of regularization by
variance of the loss when $f$ is smooth, no matter what choice of $f$. In our
setting, such limiting similarity is not the case, and it may be unrealistic
to assume a user of the approach can justify the appropriate choice of
$f$. Although we provide heuristics for choosing $f$ and $\rho$ in
Section~\ref{section:experiments}, a principled understanding of these
adaptive procedures is an important future direction of research.

The minimax guarantees demonstrate tradeoffs in terms of the
robustness we provide, in the sense that larger robustness sets yield more
difficult estimation and optimization problems. Our upper and lower bounds
match up to rates in $n$ of $n^{-1/k_*}$ (up to logarithmic factors), though
not in dimension dependence, so
our understanding of higher-dimensional robustness is limited.  Obtaining
convergence guarantees (Section~\ref{section:upper}) with scale-sensitive
model complexity terms such as Rademacher complexity and its localized
variants~\cite{BartlettBoMe05} is also a topic of future research.  In our
asymptotic results (Section~\ref{section:asymptotics}), we require an
identifiability assumption on the dual formulation, and it is open whether
this assumption can be relaxed by analyzing the primal problem directly.

The robust formulation~\eqref{eqn:objective} and its empirical
formulation~\eqref{eqn:plug-in} are complementary to traditional robustness
approaches in statistics arising out of Huber's
work~\cite{Huber81,HuberRo09}. In the classical notions of Huber robustness,
one wishes to obtain an estimate of a parameter $\theta$ of a
distribution $P_0$ contaminated by some $Q$; in our case, in contrast,
we wish to obtain a parameter that performs well \emph{for all} contaminations
$Q$, at least contaminations nearby in some $f$-divergence ball. Developing a
deeper understanding of the connections and contrasts between classical
contamination models and distributional robustness approaches will likely
yield fruit.

Two related issues arise when we consider problems with covariates $X$ and a
outcome $Y$. The distributionally robust formulation~\eqref{eqn:objective}
considers shifts in the joint distribution $(X, Y) \sim P_0$. Traditional
domain adaptation approaches, in contrast, take a fixed conditional
distribution $P_{0, Y|X}(y \mid x)$ and consider shifts to the marginal
distribution $P_{0, X}$ (covariate shift). In causal data analyses, one wishes
to perturb only the distribution of the covariates $X$, observing the effect
of such interventions on $Y$.  Connecting these ideas and developing variants
of the formulation~\eqref{eqn:objective} that only protect against covariate
shift or structural shifts on $X$ may be useful in many scenarios.



%% file: formulation-proof.tex
\section{Proof of Duality Results}

\subsection{Proof of Lemma~\ref{lemma:cressie-read-risk}}
\label{section:proof-of-cressie-read-risk}

First, we compute the Fenchel conjugate for Cressie-Read family of divergences
$f_k$.
\begin{lemma}
  \begin{equation}
    \label{eqn:cressie-read-conjugate}
    f_k^*(s) = \frac{1}{k}\hinge{(k-1)s + 1}^{k^*} - \frac{1}{k}
\end{equation}
\end{lemma}
\begin{proof}
  Consider the supremum $f^*(s) = \sup_t \{st - f(t)\}$.
  Then for $t \ge 0$, we have
  \begin{equation*}
    \frac{\partial}{\partial t} \left[st - f_k(t)\right] = s -
    \frac{1}{k-1} (t^{k-1}-1).
  \end{equation*}
  If $s < 0$, then the supremum is attained at $t = 0$, as the
  derivative above is $< 0$ at $t = 0$. If $s \ge -\frac{1}{k-1}$,
  then we solve $\frac{\partial}{\partial t} \left[st - f_k(t)\right]
  = $ to find $t = \left((k-1)s + 1\right)^{1/(k-1)}$, and substituting gives
  \begin{equation*}
    st - f(t) = \frac{1}{k}\left((k-1)s + 1\right)^{\frac{k}{k-1}} - \frac{1}{k}
  \end{equation*}
  which is our desired result as $1 - 1/k = 1/k_*$.
\end{proof}

From the dual formulation~\eqref{eqn:dual}, we have
\ifdefined\useaosstyle
\begin{align*}
  & \sup_{P \ll P_0} \left\{\E_P[Z] ~ \mbox{s.t.~} \fdiv{P}{P_0} \le \tol
  \right\} \\
  & = \inf_{\lambda \ge 0, \eta}
    \left\{ \lambda\E_{P_0} f^*\left(\frac{Z - \eta}{\lambda}\right)
    + \lambda \tol + \eta \right\} \\
  & = \inf_{\lambda \ge 0, \eta}
    \left\{\frac{(k-1)^{k_*}}{k} \lambda^{1-k_*}
    \E_{P_0} \hinge{Z - \eta + \frac{\lambda}{k-1}}^{k_*}
    + \lambda (\tol - \frac{1}{k}) + \eta \right\} \\
  & = \inf_{\lambda \ge 0, \tilde{\eta}}
    \left\{
    (k-1)^{k_*}k^{-1} \E_{P_0}\hinge{Z - \tilde{\eta}}^{k_*}
    \lambda^{1-k_*} 
    + \left(\tol + \frac{1}{k(k-1)}\right)\lambda
    + \tilde{\eta} \right\}
\end{align*}
\else
\begin{align*}
  \sup_{P \ll P_0} \left\{\E_P[Z] ~ \mbox{s.t.~} \fdiv{P}{P_0} \le \tol
  \right\}
  & = \inf_{\lambda \ge 0, \eta}
    \left\{ \lambda\E_{P_0} f^*\left(\frac{Z - \eta}{\lambda}\right)
    + \lambda \tol + \eta \right\} \\
  & = \inf_{\lambda \ge 0, \eta}
    \left\{\frac{(k-1)^{k_*}}{k} \lambda^{1-k_*}
    \E_{P_0} \hinge{Z - \eta + \frac{\lambda}{k-1}}^{k_*}
    + \lambda (\tol - \frac{1}{k}) + \eta \right\} \\
  & = \inf_{\lambda \ge 0, \tilde{\eta}}
    \left\{
    (k-1)^{k_*}k^{-1} \E_{P_0}\hinge{Z - \tilde{\eta}}^{k_*}
    \lambda^{1-k_*} 
    + \left(\tol + \frac{1}{k(k-1)}\right)\lambda
    + \tilde{\eta} \right\}
\end{align*}
\fi
where the last line followed by setting $\tilde{\eta} \defeq \eta - \frac{\lambda}{k-1}$.
Taking derivatives with respect to $\lambda$ to infimize the preceding
expression, we have (noting that $(k_* - 1) / k_* = 1/k$)
\begin{align}
  \lambda = (k-1) (k(k-1)\tol + 1)^{-\frac{1}{k_*}} \left(\E_{P_0}\hinge{Z - \tilde{\eta}}^{k_*}\right)^{\frac{1}{k_*}}
  \label{eqn:lambda-min-cressie}
\end{align}
By substituting into the preceding expression, we find that
the supremum is
\begin{align*}
  \inf_{\tilde{\eta}} \left(k(k-1)\tol + 1\right)^{\frac{1}{k}}
  \left(\E_{P_0}\hinge{Z - \tilde{\eta}}^{k_*}\right)^{1/k_*} + \tilde{\eta}.
\end{align*}

\subsection{Moments and duality}
\label{sec:moments-duality}

\newcommand{\kdual}{k_*}

We discuss the norm-like behavior of the robust risk $\risk_f$ when $f$
behaves asymptotically as $t^k$ for some $k \in (1, \infty)$; we treat $c_j$
and $C_j$ as constants whose values may change from line to line.  For
simplicity we assume that $f$ is differentiable, though subdifferential
calculus~\cite{HiriartUrrutyLe93ab} allows immediate extension to the
non-differentiable case.  Assume that $0 < \liminf_{t \to \infty} f(t) / t^k
\le \limsup_{t \to \infty} f(t) / t^k < \infty$.  Then as $t \mapsto f'(t)$
is non-decreasing, there exist $0 < c_0 \le c_1 < \infty$ such that $c_0
t^{k-1} \le f'(t) \le c_1 t^{k-1}$ for all large enough $t$.  Then for all
large $s$ the $t$ solving $f'(t) = s$ satisfies $c_0 t^{k-1} \le s \le c_1
t^{k-1}$, that is, $(s / c_1)^\frac{1}{k-1} \le (f')^{-1}(s) \le
(s/c_0)^\frac{1}{k-1}$.  Recall that the conjugate $f^*(s) \defeq \sup_{t}
\{s t - f(t)\}$ satisfies the duality $(f')^{-1}(s) = (f^*)'(s)$, and as
$\dom f \subset \R_+$, $f^*$ is
non-decreasing~\cite{HiriartUrrutyLe93ab}. Then there are constants $C_0,
C_1$ such that for all large $s$, we evidently have
\begin{equation*}
  C_0 s^\frac{1}{k-1} \le (f^*)'(s) \le
  C_1 s^\frac{1}{k-1},
\end{equation*}
and so by an integration argument
\begin{equation*}
  C_0 s^{\kdual} \le f^*(s) \le C_1 s^{\kdual}
  ~~ \mbox{for~large~} s.
\end{equation*}
In particular, for some threshold $\tau_f$ depending on $f$,
if we define the shorthand $Z = \loss(\param; \statrv)$ then
the dual~\eqref{eqn:dual} satisfies
\begin{align*}
  \lefteqn{\inf_{\lambda \ge 0, \eta \in \R}
    \left\{C_0 \lambda^{1 - \kdual}
    \E\left[(Z - \eta)^\kdual
      \indic{Z \ge \lambda \tau_f} \right]
    + \lambda \E\left[f^*\left(\frac{Z - \eta}{\lambda}
      \right) \indic{Z < \lambda \tau_f} \right]
    + \eta + \lambda \tol \right\}} \\
  & \le
  \risk_f(\param; P_0) \\
  & \le
  \inf_{\lambda \ge 0, \eta \in \R}
  \left\{C_1 \lambda^{1 - \kdual}
  \E\left[(Z - \eta)^\kdual
    \indic{Z \ge \lambda \tau_f} \right]
  + \lambda \E\left[f^*\left(\frac{Z - \eta}{\lambda}
    \right) \indic{Z < \lambda \tau_f} \right]
  + \eta + \lambda \tol \right\}.
\end{align*}
That $\inf_{\lambda \ge 0} \{\lambda \rho + C \lambda^{1 - \kdual}\}
= (\kdual - 1)^{1/\kdual} (1 + \frac{1}{\kdual})
\rho^{1/k} C^{1/\kdual}$ shows that once again, we have the
dependence of $\risk_f$ on $\kdual$th moments of the loss.

\section{Proofs of Examples}

\subsection{Proof of Example~\ref{example:mean-estimation}}
\label{sec:proof-example-mean}

That $\theta_1 = \sum_v p_v \theta_v$ is immediate. For the second
claim, we begin with a characterization of the Chebyshev center and
a few of its properties.
We have
\begin{equation*}
  \theta\maximin = \sum_v s_v \theta_v
  ~~~ \mbox{where}~~ \ones^T s = 1, s \succeq 0,
\end{equation*}
and by the KKT conditions for optimality, $s_v > 0$ only if $\max_{w
  \in V} \ltwos{\theta\maximin - \theta_w} = \ltwos{\theta_v -
  \theta_w}$.  We recall that a function $h$ is
$c$-strongly
if $\<\nabla h(\theta) - \nabla h(\tau), \theta - \tau\> \ge c
\ltwo{\theta - \tau}^2$, and $\theta\opt$ minimizes $h$ over $\Theta$ if
and only if for some $g \in \partial h(\theta\opt)$ we have $\<g, \theta -
\theta\opt\> \ge 0$ for all $\theta \in \Theta$. Let
$h(\theta) = \half \max_{v \in V} \ltwo{\theta - \theta_v}^2$.
Then
letting $\theta \neq \theta\maximin$, we immediately see that
for any $v \in V$ for which
$\ltwo{\theta - \theta_v} = \max_{w \in V} \ltwo{\theta - \theta_w}$, we have
for $g = \theta - \theta_v \in \partial h(\theta)$ and
$g' \in \partial h(\theta\maximin)$ that
\begin{equation}
  \label{eqn:chebyshev-forced-growth}
  \begin{split}
    \<\theta - \theta_v, \theta - \theta\maximin\>
    & = \<g, \theta - \theta\maximin\> \\
    & \ge \<g', \theta - \theta\maximin\>
    + \ltwos{\theta - \theta\maximin}^2
    \ge \ltwos{\theta - \theta\maximin}^2.
  \end{split}
\end{equation}

Fix $\theta \in \R^d$ with $\theta \neq \theta\maximin$.
For fixed $\eta > 0$, consider the objectives
$h_v(\cdot; \eta) : \R \to \R$,
\begin{equation*}
  h_v(t; \eta) \defeq
  \E\left[\hinge{(1/2) \ltwos{(1 - t) \theta + t \theta\maximin
        - \theta_v - \noise}^2 - \eta}\right].
\end{equation*}
By the continuity of the density of $\noise$, $h_v$ is differentiable in
$t$, and we have
\begin{equation*}
  h_v'(t; \eta) =
  \E\left[\indic{\ltwos{(1 - t) \theta + t \theta\maximin
        - \theta_v - \noise}^2 \ge 2 \eta}
    \<\theta\maximin - \theta,
    (1 - t) \theta + t \theta\maximin - \theta_v - \noise\>
    \right]
\end{equation*}
and
\begin{equation}
  \label{eqn:derivative-at-zero}
  h_v'(0; \eta) =
  \E\left[\indic{\ltwos{\theta - \theta_v - \noise}^2 \ge 2 \eta}
    \<\theta\maximin - \theta, \theta - \theta_v - \noise\> \right].
\end{equation}
For any vector $\mu \in \R^d$, define the constant
$c(\eta; \mu) > 0$ such that
\begin{equation*}
  \E[\indic{\ltwos{\mu + \noise}^2 \ge 2 \eta} (\mu + \noise)]
  = c(\eta; \mu) \mu,
\end{equation*}
which must exist by the rotational symmetry of the Gaussian.
Now we claim that for any $\mu_1, \mu_2$ with $\ltwo{\mu_1} > \ltwo{\mu_2}$,
\begin{equation}
  \label{eqn:constant-claims}
  \lim_{\eta \to \infty} \frac{c(\eta; \mu_1)}{c(\eta; \mu_2)} = \infty.
\end{equation}
Deferring the proof of the claim~\eqref{eqn:constant-claims},
let us see how it yields the theorem.

We shall show that  for $\theta \neq \theta\maximin$,
if we define
\begin{equation*}
  R(t; \eta) \defeq
  \sum_v p_v \E\left[\hinge{(1/2) \ltwo{(1 - t) \theta
        + t \theta\maximin - \theta_v - \noise}^2 - \eta}\right],
\end{equation*}
then for all large $\eta$, we have
$R'(0; \eta) < 0$, so
$\theta$ cannot minimize
$\sum_v p_v \E[\hinge{(1/2) \ltwos{\theta - \theta_v - \noise}^2
    - \eta}]$.
That this gives the theorem is nearly immediate,
because for all $0 < \alpha \le 1$, the $\eta$ minimizing
the CVaR risk
$\alpha^{-1} \E[\hinge{\loss(\theta; Y) - \eta}] + \eta$ is
the $1 - \alpha$ quantile of $\loss(\theta; Y)$, which for
our setting evidently tends to $\infty$ as $\alpha \downarrow 0$.
Thus, if
$\eta(\theta, \alpha) = \argmin_\eta
\{\alpha^{-1} \E[\hinge{\loss(\theta; Y) - \eta}]
+ \eta\}$, we have $\eta(\theta, \alpha) \to \infty$ uniformly
in $\theta$ as $\alpha \downarrow 0$, and so
$R'(0; \eta) < 0$ implies that
$\theta$ cannot minimize
$\cvar{\alpha}(\loss(\theta; Y))$.
To see that $R'(0; \eta) < 0$, simply note that
\begin{equation*}
  R'(0; \eta)
  = \sum_v p_v c(\eta; \ltwo{\theta - \theta_v})
  \<\theta\maximin - \theta, \theta - \theta_v\>
\end{equation*}
by Eq.~\eqref{eqn:derivative-at-zero}. Let $V\opt
= \{v : \ltwo{\theta - \theta_v} = \max_w \ltwo{\theta - \theta_w}\}$.
Then inequality~\eqref{eqn:chebyshev-forced-growth} implies that
\begin{align*}
  R'(0; \eta)
  & \le -\bigg(\sum_{v \in V\opt} p_v c(\eta; \ltwo{\theta - \theta_v})
  \bigg)
  \ltwos{\theta - \theta\maximin}^2 \\
  & \quad ~ + \sum_{v \not \in V\opt} p_v c(\eta; \ltwo{\theta - \theta_v})
  c(\eta; \ltwo{\theta - \theta_v})
  \<\theta\maximin - \theta, \theta - \theta_v\>.
\end{align*}
In particular, for any $v \in V\opt$ we have
\begin{equation*}
  \frac{1}{c(\eta; \ltwo{\theta - \theta_v})}
  R'(0; \eta) \le -\bigg(\sum_{v \in V\opt} p_v \bigg)
  \ltwo{\theta - \theta\maximin}^2 + o(1)
\end{equation*}
as $\eta \to \infty$, giving the theorem.

Finally, we return to prove the claim~\eqref{eqn:constant-claims}, note that
$\ltwo{\mu + \noise}^2$ follows a non-central
$\chi^2$ distribution with
$\P(\ltwo{\mu + \noise}^2 \ge t)
= Q_{d/2}(\ltwo{\mu}, \sqrt{t})$ for $Q$ the
Marcum Q-function. Letting $\Phi$ be the standard Gaussian
CDF, the Marcum Q-function satisfies the asymptotics
(e.g.~\cite[p.~81]{SimonAl00} or~\cite[Eq.~(4)]{SunBaZh10})
that as $t \to \infty$,
\begin{equation*}
  Q_k(a, t) = (1 + o(1))
  \left(\frac{t}{a}\right)^{k - 1/2}
  (1 - \Phi(t - a))
  = (1 + o(1))
  \frac{1}{t - a} \exp\left(-\frac{(t - a)^2}{2}\right)
  \left(\frac{t}{a}\right)^{k - 1/2}.
\end{equation*}
Thus we obtain
\begin{align*}
  \frac{\P(\ltwo{\mu_1 + \noise}^2 \ge t^2)}{
    \P(\ltwo{\mu_2 + \noise}^2 \ge t^2)}
  & = (1 + o(1))
  \left(\frac{\ltwos{\mu_2}}{\ltwos{\mu_1}}\right)^\frac{d - 1}{2}
  \exp\left(-\frac{(t - \ltwos{\mu_1})^2}{2}
  + \frac{(t - \ltwos{\mu_2})^2}{2}\right) \\
  & = (1 + o(1))
  \left(\frac{\ltwos{\mu_2}}{\ltwos{\mu_1}}\right)^\frac{d - 1}{2}
  \exp\left(t (\ltwo{\mu_1} - \ltwo{\mu_2}) +
  \half (\ltwo{\mu_2}^2 - \ltwo{\mu_1}^2)\right)
  \to \infty
\end{align*}
as $t \to \infty$, giving the claim~\eqref{eqn:constant-claims}.

\subsection{Proof of Example~\ref{example:mixture-linear-regression}}
\label{sec:proof-mixture-linear-regression}

That $\theta\ols = \theta_1 = \int \theta_v d\mu(v)$ is immediate.

We begin with a technical lemma on the expectations of Gaussian variables
whose proof we defer to Sec.~\ref{sec:proof-asymptotics-hinge}.
\begin{lemma}
  \label{lemma:asymptotics-hinge}
  Let $Z \sim \normal(0, 1)$. Then
  \begin{equation*}
    \E\left[\hinge{Z^2 - t}\right]
    = \sqrt{\frac{2}{\pi}}
    \left(t - 1 + O(t^{-1})\right) e^{-\half t^2}.
  \end{equation*}
\end{lemma}

Defining the shorthand
$\tau_v^2(\theta) \defeq (\theta - \theta_v)^T \Sigma_v
(\theta - \theta_v) + \sigma_v^2$, for $Z \sim \normal(0, 1)$ we evidently
have
\begin{equation}
  \label{eqn:mixture-cvar}
  \cvar{\alpha}(\loss(\theta; X, Y))
  = \inf_\eta \left\{ \frac{1}{2 \alpha}
  \int \E\left[\hinge{\tau_v^2(\theta) Z^2 - 2 \eta} \right] d\mu(v)
  + \eta
  \right\}.
\end{equation}
For large $\eta$,
Lemma~\ref{lemma:asymptotics-hinge} gives
\begin{align}
  \nonumber
  \E\left[\hinge{\tau_v^2(\theta) Z^2 - 2 \eta}\right]
  & = \tau_v^2(\theta)
  \E\left[\hinge{Z^2 - 2 \eta / \tau_v^2(\theta)}\right] \\
  &
  = C(1 + O(\tau_v^2(\theta)/\eta))
  \eta \exp\left(-\frac{2 \eta^2}{\tau_v^4(\theta)}\right)
  \label{eqn:asymptotics-tau-v}
\end{align}
uniformly in $\tau_v^2(\theta)$, where $C = 2\sqrt{2 / \pi}$
is a fixed constant.

We now compute normalized asymptotics of the mixture
CVaR~\eqref{eqn:mixture-cvar}.
For a measure $\mu$ on $V$ and measurable $g : V \to \R$, 
we define the quantile
$\quant_p(g, \mu) \defeq
\inf\{t \in \R : p \le \mu(g^{-1}(\openleft{-\infty}{t}))\}$,
which gives the following.
\begin{lemma}
  \label{lemma:obvious-essential-suprema}
  Let $\epsilon > 0$ and let
  $\esssup_v g(v) = \inf \{t \mid \mu(\{v : g(v) > t\}) = 0\}$
  be the essential supremum of $g$.
  Then for all $t > 0$,
  \begin{equation*}
    t \, \quant_{1-\epsilon}(g, \mu)
    - \log \frac{1}{\epsilon}
    \le \log \left(\int e^{t g(v)} d\mu(v)\right)
    \le t \esssup_{v \in V} g(v).
  \end{equation*}
\end{lemma}
\begin{proof}
  Clearly
  $\log \int e^{t g(v)} d\mu(v) \le t \esssup_{v \in V} g(v)$.
  For the lower bound, letting $q = \quant_{1 - \epsilon}(g, \mu)$, we have
  $\log \int e^{t g(v)} d\mu(v)
  \ge \log \int e^{t q} \indic{g(v) \ge q}
  d\mu(v)
  \ge \log \epsilon + t q$.
\end{proof}

For $\theta \in \R^d$ and $\eta > 0$ define the normalized logarithmic risk
\begin{equation*}
  R(\theta; \eta)
  \defeq \frac{1}{2 \eta^2}
  \log \int \E_v\left[\hinge{\tau_v^2(\theta) Z^2 - 2 \eta}\right]
  d\mu(v),
\end{equation*}
which by Lemma~\ref{lemma:obvious-essential-suprema}
satisfies
\begin{align*}
  \lefteqn{\quant_{1 - \epsilon}\left(-\frac{1}{\tau_v^4(\theta)}, \mu\right)
    - \frac{\log \frac{1}{\epsilon}}{2 \eta^2}
    + \frac{\log \left(C(1 + O(1/\eta)) \eta\right)}{2 \eta^2}} \\
  & \qquad\qquad ~ \le R(\theta; \eta)
  \le \sup_v \left\{-\frac{1}{\tau_v^4(\theta)}\right\}
  + \frac{\log \left(C(1 + O(1/\eta)) \eta\right)}{2 \eta^2},
\end{align*}
where we have used the boundedness assumptions on $\Sigma_v$,
$\sigma_v$, and $\theta_v$.

Assume now that $\theta \neq \theta\maximin$.
Then $\sup_v \tau_v(\theta) > \sup_v \tau_v(\theta\maximin)$ because
$\Sigma_v \succ 0$ and so $\theta\maximin$ must be unique.
By the assumption that the essential suprema and suprema over $v$ are
equal,
$\quant_{1-\epsilon}(-1/\tau_v^4(\theta)) \to
\sup_v \{-1/\tau_v^4(\theta)\}$ as $\epsilon \to 0$, for
all large enough $\eta$ we have
\begin{equation*}
  R(\theta\maximin; \eta) < R(\theta; \eta).
\end{equation*}
Notably, if $\theta_\eta \in \argmin_\theta R(\theta; \eta)$, then
evidently $\theta_\eta \to \theta\maximin$.


Finally, we consider the quantity
\begin{equation*}
  \eta_\alpha(\theta) \defeq \argmin_\eta
  \left\{\frac{1}{\alpha}\int \E_v\left[\hinge{
      \loss(\theta; X, Y) - \eta}\right] d\mu(v) + \eta \right\}.
\end{equation*}
By definition, we have
\begin{equation*}
  \argmin_\theta \cvar{\alpha}(\loss(\theta; X, Y))
  = \argmin_\theta \int\E_v[\hinge{\loss(\theta; X, Y) - \eta_\alpha(\theta)}]
  d\mu(v),
\end{equation*}
and moreover, $\eta_\alpha(\theta) = \quant_{1 - \alpha}(\loss(\theta;
X, Y))$, where the quantile is computed jointly over
$v \sim \mu$ and $(X, Y)$. As by assumption
$\inf_v \sigma_v^2 > 0$, it is evident
that $\liminf_{\alpha \downarrow 0} \inf_\theta \eta_\alpha(\theta) = \infty$.
In particular, for all small enough $\alpha > 0$, we obtain
$R(\theta\maximin, \eta_\alpha(\theta))
< R(\theta, \eta_\alpha(\theta))$, or
\begin{equation*}
  \int \E_v\left[\hinge{\loss(\theta\maximin; X, Y) - \eta_\alpha(\theta)}
    \right]
  < \int \E_v\left[\hinge{\loss(\theta; X, Y) - \eta_\alpha(\theta)}
    \right],
\end{equation*}
giving the result we claim in
Example~\ref{example:mixture-linear-regression}.

\subsubsection{Proof of Lemma~\ref{lemma:asymptotics-hinge}}
\label{sec:proof-asymptotics-hinge}

Let $\Phi(t) = \P(Z \le t)$ be the standard normal CDF and
$\Gamma(a, x) = \int_x^\infty z^{a - 1} e^{-z} dz$ be the
incomplete Gamma function.
For $t \ge 0$ we have
\begin{align*}
  \E[\hinge{Z^2 - t}]
  & = \sqrt{\frac{2}{\pi}} \int_t^\infty (z^2 - t) e^{-\half z^2} dz \\
  & \stackrel{(i)}{=} \frac{2}{\sqrt{\pi}} \int_{t^2 / 2}^\infty u^{1/2} e^{-u} du
  - 2t (1 - \Phi(t)) \\
  & = \frac{2}{\sqrt{\pi}} \Gamma\left(\frac{3}{2}, \frac{t^2}{2}\right)
  - 2t (1 - \Phi(t))
\end{align*}
where inequality~$(i)$ is via the substitution $u = z^2 / 2$.  Now we use
asymptotics of the normal CDF and incomplete Gamma function to approximate
the preceding display for large $t$. By standard normal
approximations~\cite[Eq.~(7.1.13)]{AbramowitzSt65}, we have for $t \ge 0$
that
\begin{equation*}
  \frac{t}{t^2 + 1} e^{-\half t^2}
  \le \frac{2}{t + \sqrt{t^2 + 4}}
  e^{-\half t^2}
  \le \sqrt{2 \pi} (1 - \Phi(t))
  \le
  \frac{2}{t + \sqrt{t^2 + 8/\pi}} e^{-\half t^2}
  \le \frac{1}{t} e^{-\half t^2}.
\end{equation*}
As $\sqrt{t^2 + c} = t + c / 2t + O(t^{-2})$ for any constant $c$,
we have
\begin{equation*}
  (1 - \Phi(t))
  = \frac{1}{\sqrt{2 \pi}} \left(t^{-1} - O(1)t^{-3}\right) e^{-\half t^2},
\end{equation*}
while we also have the well-known asymptotic expansion
\begin{equation*}
  \Gamma(a + 1, x)
  = \frac{e^{-x} x^{a + 1}}{x - a}
  \left[1 - \frac{a}{(x - a)^2} + \frac{2 a}{(x - a)^3}
    + O\left(\frac{a^2}{(x - a)^4}\right)\right]
\end{equation*}
as $\sqrt{a} / (x - a) \to 0$~\cite[Eq.~(2.12)]{Gautschi97}.

Substituting these above (with $x = t^2 / 2$ and $a = 1/2$) yields for
large $t$ that
\begin{align*}
  \sqrt{\pi / 2} e^{\half t^2}
  \E\left[\hinge{Z^2 - t}\right]
  & = \frac{t^3}{t^2 - 1}
  \left[1 - \frac{2}{(t^2 - 1)^2}
    + O(t^{-6})\right]
  - t \left(\frac{1}{t} - \frac{O(1)}{t^3}\right) \\
  & = t - 1 + O(t^{-1}),
\end{align*}
giving the lemma.


%% file: upper-proof.tex
\section{Proof of Upper Bounds}
\label{section:proof-upper}

\subsection{Proof of Theorem~\ref{theorem:concentration}}
\label{section:proof-of-concentration}

To ease notation, for any fixed $\param \in \Theta$, let
$Z(x) = \loss(\param; x)$ and
\begin{equation*}
  g_k(\eta; P) \defeq c_k \left( \E_{P}[\hinge{Z - \eta}^{k_*}\right)^{\frac{1}{k_*}} + \eta
\end{equation*}
so that $\risk_k(Z; P) = \inf_{\eta} g_k(\eta; P)$ from
Proposition~\ref{proposition:duality}. We begin by showing pointwise
concentration of $g_k(\eta; \emp)$ to $g_k(\eta; P_0)$ for each bounded
$\eta$. First, we begin by recalling a standard convex Lipschitz concentration
inequality for bounded random variables.
\begin{lemma}[Boucheron et al.~\citeyear{BoucheronLuMa13}, Theorem 6.10]
  \label{lemma:convex-concentration}
  Let $h : \R^n \to \R$ be convex or concave and $L$-Lipschitz with respect to
  the $\ell_2$-norm. Let $Z_i$ be independent random variables with
  $Z_i \in [a, b]$. For $t \ge 0$,
  \begin{equation*}
    \P(|h(Z_1^n) - \E[h(Z_1^n)]| \ge t)
    \le 2 \exp\left(-\frac{t^2}{2 L^2(b - a)^2}\right).
  \end{equation*}
\end{lemma}
\noindent To apply Lemma~\ref{lemma:convex-concentration}, we verify that
$g_k(\eta; \emp)$ is Lipschitz in the data vector $Z_1^n$ by using the
following elementary result.
\begin{lemma}
  \label{lemma:lp-norm-lipschitz}
  The map
  $\R^n \ni y \mapsto \left( \frac{1}{n} \sum_{i=1}^n |y_i|^{k_*}
  \right)^{\frac{1}{k_*}}$ is $n^{-\frac{1}{2\vee k_*}}$-Lipschitz with
  respect to the $\ltwo{\cdot}$-norm.
\end{lemma}
\begin{proof-of-lemma}
  We denote
  $\norm{Y}_{L^p(\emp)} = \left( \frac{1}{n} \sum_{i=1}^n |y_i|^{p}
  \right)^{\frac{1}{p}}$ to ease notation.  Noting that
  \begin{equation*}
    \ltwo{\psi_n(y)} = n^{-\half} \left(
      \frac{\norm{Y}_{L^{2(k_*-1)}(\emp)}}{\norm{Y}_{L^{k_*}(\emp)}}
    \right)^{k_*-1},
  \end{equation*}
  we proceed in two cases. If $k_* \le 2$, the result follows from
  $\norm{Y}_{L^{2(k_*-1)}(\emp)} \le \norm{Y}_{L^{k_*}(\emp)}$. If
  $k_* \ge 2$, 
  $\left(\sum_{i=1}^n |y_i|^{2(k_*-1)}\right)^{\frac{1}{2(k_*-1)}} \le
  \left(\sum_{i=1}^n |y_i|^{k_*}\right)^{\frac{1}{k_*}}$ implies
  \begin{equation*}
    \norm{Y}_{L^{2(k_*-1)}(\emp)} \le n^{-\frac{1}{k_*}+\frac{1}{2(k_*-1)}}
    \norm{Y}_{L^{k_*}(\emp)},
  \end{equation*}
  which gives the result.
\end{proof-of-lemma}
\noindent Lemma~\ref{lemma:lp-norm-lipschitz} implies $g_k(\eta; \emp)$ is a
$c_kn^{-\frac{1}{2\vee k_*}}$-Lipschitz function of the data vector $Z_1^n$
with respect to the $\ltwo{\cdot}$-norm.  Applying
Lemma~\ref{lemma:convex-concentration},  for any fixed
$\eta \in [-\frac{1}{c_k-1} \zbound, \zbound]$
\begin{equation}
  \label{eqn:concentration-to-mean}
  |g_k(\eta; \emp) - \E_{P_0}[g_k(\eta; \emp)] | \le
  \sqrt{2t} c_k \left( \frac{c_k}{c_k-1} \vee 2\right) \zbound n^{-\frac{1}{k_* \vee 2}}  
\end{equation}
with probability at least $1-2e^{-t}$.

To establish pointwise concentration of $g_k(\eta; \emp)$ to $g_k(\eta; P_0)$, it
remains to see that $\E_{P_0}[g_k(\eta; \emp)]$ and $g_k(\eta; P_0)$ are close.
We use the following lemma, whose proof we defer to
Section~\ref{section:proof-of-norm-lower-bounds}.
\begin{lemma}
  \label{lemma:norm-lower-bounds}
  Let $k_* \in \openright{1}{\infty}$ and let
  $Y_i$ be an i.i.d.\ sequence of random variables satisfying
  $\E[|Y|^{2k_*}] \le C^{k_*} \E[|Y|^{k_*}]$ for some $C \in \R_+$.
  For any $k_* \in \openright{1}{\infty}$, we have
  \begin{equation}
    \label{eqn:scaled-norm-lower-bound}
    \E\Bigg[\bigg(
    \frac{1}{n}\sum_{i=1}^n |Y_i|^{k_*}\bigg)^{\frac{1}{k_*}}\Bigg]
    \ge \E[|Y|^{k_*}]^{\frac{1}{k_*}}
    - \frac{2}{k} \sqrt{C} n^{-\frac{1}{k_* \vee 2}}
  \end{equation}
\end{lemma}
\noindent Since
$\E\Bigg[\bigg( \frac{1}{n}\sum_{i=1}^n
|Y_i|^{k_*}\bigg)^{\frac{1}{k_*}}\Bigg] \le \E[|Y|^{k_*}]^{\frac{1}{k_*}}$ by
Jensen's inequality, Lemma~\ref{lemma:norm-lower-bounds} implies
\begin{equation*}
  \left| \E_{P_0}[g_k(\eta; \emp)] - g_k(\eta; P_0)\right|
  \le  \frac{2c_k}{k}
  \sqrt{\left( \frac{c_k}{c_k-1} \vee 2 \right) \zbound}
  n^{-\frac{1}{k_* \vee 2}}
\end{equation*}
for any fixed $\eta \in [-\frac{1}{c_k-1} \zbound, \zbound]$.  Combining the
bound with the concentration result~\eqref{eqn:concentration-to-mean}, we
conclude that with probability at least $1-2e^{-2t}$,
\begin{equation}
  \label{eqn:pointwise-concentration}
  |g_k(\eta; \emp) - g_k(\eta; P_0)] | \le n^{-\frac{1}{k_* \vee 2}}  
   \zbound  c_k \left( \frac{c_k}{c_k-1} \vee 2\right)
  \left( \sqrt{2t}  + \frac{2}{k} \right)  \eqdef \epsilon_{t}.
\end{equation}

We now show uniform concentration by using a simple covering argument. The
following lemma restricts the domain of $\eta$ to a compact set, which is
essential to this argument.
\begin{lemma}
  \label{lemma:dual-opt-over-compact-set}
  If $Z \in [0, \zbound]$, then for any distribution $P$
  \begin{equation*}
    \inf_{\eta \in \R} g(\eta; P) = \inf_{\eta } \left\{ g(\eta; P): \eta \in \left[-\frac{1}{c_k-1} \zbound, \zbound\right] \right\}.
  \end{equation*}
\end{lemma}
\begin{proof-of-lemma}
  By definition, $g(\eta; P) = \eta$ for $\eta \ge \zbound$, and
  \begin{equation*}
    g\left(- \frac{1}{c_k-1} \zbound; P\right) \ge  c_k \frac{\zbound}{c_k-1} - \frac{\zbound}{c_k-1}
    = \zbound = g(\zbound; P).
  \end{equation*}
  Since $\eta \mapsto g(\eta; P)$ is convex, this implies the result.
\end{proof-of-lemma}
\noindent Recalling the shorthand $\epsilon_{t, n} \defeq  n^{-\frac{1}{k_* \vee 2}} \zbound 
  c_k \left( \frac{c_k}{c_k-1} \vee 2\right) 
  \left( \sqrt{2t}  + \frac{2}{k} \right)$, define the sequence
\begin{equation*}
  \eta_i \defeq - (c_k-1)^{-1}\zbound +  i \epsilon_{t, n}
\end{equation*}
for nonnegative integers $i \le \frac{c_k}{c_k-1}
\frac{\zbound}{\epsilon_{t, n}}$. Then, for any
$\eta \in [-(c_k-1)^{-1}\zbound, \zbound]$, there exists
$1 \le i(\eta) \le \frac{c_k}{c_k-1} \frac{\zbound}{\epsilon_{t, n}}$ such that
$|\eta - \eta_{i(\eta)}| \le \epsilon_{t, n}$.
\begin{align*}
  & \sup_{\eta \in [-(c_k-1)^{-1}\zbound, \zbound]} | g(\eta; \emp) - g(\eta; P_0)| \\
  & \le \sup_{\eta \in [-(c_k-1)^{-1}\zbound, \zbound]}
    \left\{ | g(\eta; \emp) - g(\eta_{i(\eta)}; \emp) |
    + | g(\eta_{i(\eta)}; \emp) - g(\eta_{i(\eta)}; P_0) |
    + | g(\eta_{i(\eta)}; P_0) - g(\eta; P_0) | \right\} \\
  & \le \max_{1\le i \le \frac{c_k}{c_k-1} \frac{\zbound}{\epsilon_{t, n}}}
    | g(\eta_{i(\eta)}; \emp) - g(\eta_{i(\eta)}; P_0) |
    + 2 (1+c_k) \epsilon_{t, n}
\end{align*}
where we used $(1+c_k)$-Lipschitzness of $\eta\mapsto g(\eta; P_0)$ and
$\eta\mapsto g(\eta; \emp)$ in the last inequality. Taking the union bound
over the pointwise concentration result~\eqref{eqn:pointwise-concentration}
with $\eta = \eta_i$, conclude from
Lemma~\ref{lemma:dual-opt-over-compact-set}
\begin{align*}
  \left| \risk_k(Z; \emp) - \risk_k(Z; P_0)\right|
  & = \left| \inf_{\eta} g_k(\eta; \emp) - \inf_{\eta} g_k(\eta; P_0)\right| \\
  & = \left| \inf_{\eta \in [-(c_k-1)^{-1}\zbound, \zbound]} g_k(\eta; \emp) - \inf_{\eta  \in [-(c_k-1)^{-1}\zbound, \zbound]} g_k(\eta; P_0)\right| \\
  & \le \sup_{\eta \in [-(c_k-1)^{-1}\zbound, \zbound]} | g(\eta; \emp) - g(\eta; P_0)| \\
  & \le (2c_k+3)\epsilon_{t, n}
\end{align*}
with probability at least
$1-2\exp\left(-t + \log\frac{c_k}{c_k-1} \frac{\zbound}{\epsilon_{t, n}}\right)$.
Doing a change of variables
$t(s) = s + \left( \frac{1}{k_* \vee 2} + 1 \right) \log n$, we obtain the
final result.

\subsubsection{Proof of Lemma~\ref{lemma:norm-lower-bounds}}
\label{section:proof-of-norm-lower-bounds}

\newcommand{\normpow}{q}

First, we claim that it suffices to show
\begin{align}
  \E[|Y|^\normpow]^{1/\normpow}
  & \ge \E\left[\bigg(\frac{1}{n}\sum_{i=1}^n |Y_i|^\normpow\bigg)^{1/\normpow}\right]
    \nonumber \\
  \label{eqn:non-scaled-norm-lower-bound}
  & \ge \E[|Y|^\normpow]^{1/\normpow}
    - 2 \frac{\normpow-1}{\normpow} \begin{cases}
      (C^{\normpow/2} \vee 1) \cdot n^{-1/\normpow} & \mbox{if~} \normpow \ge 2 \\
      (C \vee C^{1 - \normpow/2}) \cdot n^{-1/2} & \mbox{if~} \normpow < 2.
    \end{cases}
\end{align}
where the last inequality holds for $n \ge C^\normpow$ when $\normpow \ge
2$. To see how our desired bound~\eqref{eqn:scaled-norm-lower-bound} follows
from~\eqref{eqn:non-scaled-norm-lower-bound}, we
use a quick scaling argument.
Let $\alpha > 0$, and note that
$\E[|\alpha Y|^{2\normpow}] \le (C
\alpha^2)^\normpow \E[|Y|^\normpow]$ by assumption.
Let $\sigma_n \defeq
\E[(\frac{1}{n} \sum_{i=1}^n |Y_i|^\normpow)^{1/\normpow}]$ and $\sigma =
\E[|Y|^\normpow]^{1/\normpow}$ for shorthand. First, if $\normpow \ge 2$, we
have $(\alpha^2 C)^{\normpow/2} \ge 1$ if $\alpha \ge C^{-\half}$, and we
obtain
\begin{equation*}
  \alpha \sigma_n \ge \alpha \sigma - 2 \frac{\normpow - 1}{\normpow}
  \alpha^\normpow C^{\normpow/2}
  n^{-1/\normpow}
  ~~ \mbox{or} ~~
  \sigma_n \ge \sigma - 2 \frac{\normpow - 1}{\normpow} C^{\normpow/2} \alpha^{\normpow - 1} n^{-1/\normpow}.
\end{equation*}
Choosing $\alpha = C^{-\half}$ gives the
result~\eqref{eqn:scaled-norm-lower-bound} when $\normpow \ge 2$.
For $\normpow < 2$, we similarly obtain that
$C \alpha^2 \ge (C \alpha^2)^{1 - \normpow/2}$ for $\alpha \ge C^{-\half}$,
whence we have the lower bound
\begin{equation*}
  \alpha \sigma_n \ge \alpha \sigma - 2 \frac{\normpow - 1}{\normpow}
  C \alpha^2
  n^{-1/2}
  ~~ \mbox{or} ~~
  \sigma_n \ge \sigma - 2 \frac{\normpow - 1}{\normpow}
  C \alpha
  n^{-1/2}
\end{equation*}
for $\alpha \ge C^{-\half}$. Choosing $\alpha = C^{-\half}$ thus gives the
desired result~\eqref{eqn:scaled-norm-lower-bound}.

Now, we proceed to show the bound~\eqref{eqn:non-scaled-norm-lower-bound}. Let
\begin{equation*}
  \gamma_n = \argmin_{\gamma \ge 0} \left\{\frac{1}{(\normpow-1)}
  \frac{\frac{1}{n} \sum_{i=1}^n |Y_i|^\normpow}{\gamma^{\normpow-1}}
  + \gamma \right\}
  = \left(\frac{1}{n} \sum_{i=1}^n |Y_i|^\normpow\right)^{1/\normpow}
\end{equation*}
so that
\begin{equation*}
  \frac{1}{\normpow}
  \frac{\frac{1}{n} \sum_{i=1}^n |Y_i|^\normpow}{\gamma_n^{\normpow-1}}
  + \frac{(\normpow-1)\gamma_n}{\normpow}
  = \left(\frac{1}{\normpow} + \frac{\normpow-1}{\normpow}\right)\left(
  \frac{1}{n} \sum_{i=1}^n |Y_i|^\normpow\right)^{1/\normpow}
  = \left(\frac{1}{n} \sum_{i=1}^n |Y_i|^\normpow\right)^{1/\normpow}.
\end{equation*}
For any $\gamma \ge 0$ we have by the first order inequality for convexity
(as the function $\gamma \mapsto 1 / \gamma^{\normpow-1} + \gamma$ is convex for
$\gamma \ge 0$) that
\begin{align}
  \left(\frac{1}{n} \sum_{i=1}^n |Y_i|^\normpow\right)^{1/\normpow}
  & = \frac{\frac{1}{n} \sum_{i=1}^n |Y_i|^\normpow}{\normpow \gamma_n^{\normpow-1}}
  + \frac{\normpow-1}{\normpow} \gamma_n \nonumber \\
  & \ge \frac{\frac{1}{n} \sum_{i=1}^n |Y_i|^\normpow}{\normpow \gamma^{\normpow-1}}
  + \frac{\normpow-1}{\normpow}\gamma
  + \left(\frac{\normpow-1}{\normpow} - \frac{(\normpow-1) \frac{1}{n} \sum_{i=1}^n |Y_i|^\normpow}{
    \normpow \gamma^\normpow}\right)(\gamma_n - \gamma).
  \label{eqn:convexity-with-ks}
\end{align}
We now show how to provide a bound on magnitude of the final term
in expression~\eqref{eqn:convexity-with-ks}.

Let $\sigma^\normpow = \E[|Y|^\normpow]$, and choose $\gamma^\normpow = \max\{n^{-\alpha},
\sigma^\normpow\}$, where $\alpha \ge 0$ is a power to be chosen. Then
\begin{align*}
  \E\left[\left(\frac{\normpow-1}{\normpow} - \frac{(\normpow-1)\frac{1}{n} \sum_{i=1}^n |Y_i|^\normpow}{
      \normpow \gamma^\normpow}\right)^2\right]
  & = \left(\frac{\normpow-1}{\normpow}\right)^2
  \E\left[\left(1 - \frac{\sigma^\normpow}{\gamma^\normpow}
    + \frac{\sigma^\normpow}{\gamma^\normpow}
    - \frac{\frac{1}{n} \sum_{i=1}^n |Y_i|^\normpow}{
      \gamma^\normpow}\right)^2\right] \\
  & = \left(\frac{\normpow-1}{\normpow}\right)^2
  \left[\left(1 - \sigma^\normpow / \gamma^\normpow\right)^2
    + \frac{1}{\gamma^{2\normpow} n} \var(|Y|^\normpow)\right],
\end{align*}
and noting that $\var(|Y|^\normpow) \le \E[|Y|^{2\normpow}] \le C^\normpow \E[|Y|^\normpow] =
C^\normpow\sigma^\normpow$, we have
\begin{equation*}
  \frac{1}{\gamma^{2\normpow} n} \var(|Y|^\normpow)
  \le \frac{1}{n}\frac{C^\normpow\sigma^\normpow}{\max\{n^{-2\alpha}, \sigma^{2\normpow}\}}
  = C^\normpow \min\left\{\frac{\sigma^\normpow}{n^{1 - 2 \alpha}},
  \frac{1}{n \sigma^\normpow}\right\}.
\end{equation*}
and
\begin{equation*}
  1 - \frac{\sigma^\normpow}{\gamma^\normpow}
  = 1 -
  \min\left\{n^{\alpha} \sigma^\normpow, 1\right\}
  = \hinge{1 - n^\alpha \sigma^\normpow}.
\end{equation*}
Now we provide an upper bound on the $(\gamma_n - \gamma)$
term in the product in inequality~\eqref{eqn:convexity-with-ks}.
By inspection, we have
\begin{align}
  (\gamma_n - \gamma)^2
  & = \left(\frac{1}{n} \sum_{i=1}^n |Y_i|^\normpow\right)^\frac{2}{\normpow}
  \!\! - 2 \gamma \gamma_n
    + \max\left\{n^{-\alpha} , \sigma^\normpow\right\}^\frac{2}{\normpow}
    \nonumber \\
  & \le \left(\frac{1}{n} \sum_{i=1}^n |Y_i|^\normpow\right)^\frac{2}{\normpow}
    + \max\left\{n^{-\alpha} , \sigma^\normpow\right\}^\frac{2}{\normpow}.
      \label{eqn:first-bound-on-mean-square}
\end{align}
We now state a useful intermediate lemma and consequential inequality,
deferring its proof to Section~\ref{sec:proof-one-two-power}.
\begin{lemma}
  \label{lemma:one-two-power}
  Let $\normpow \in [1, 2]$ and $a \in [1, 2]$. Then for any random variable
  $X \ge 0$,
  \begin{equation*}
    \E[X^{a\normpow}] \le
    \E[X^\normpow]^{2 - a} \E[X^{2\normpow}]^{a - 1}.
  \end{equation*}
\end{lemma}
\noindent
As an immediate consequence of Lemma~\ref{lemma:one-two-power},
we see that for $\normpow \in [1, 2]$ and non-negative random variables $X$,
we have that if $\E[X^{2\normpow}] \le C^\normpow \sigma^\normpow$,
where $\E[X^\normpow] = \sigma^\normpow$, then
\begin{equation}
  \label{eqn:funny-two-power}
  \E[X^2] \le C^{2 - \normpow} \sigma^\normpow.
\end{equation}
To see this, substitute $a = 2/\normpow \in [1, 2]$ in
Lemma~\ref{lemma:one-two-power}, which yields
\begin{equation*}
  \E[X^2] = \E[X^{a\normpow}] \le \E[X^\normpow]^{2 - \frac{2}{\normpow}}
  \E[X^{2\normpow}]^{\frac{2}{\normpow} - 1}
  \le \sigma^{2\normpow - 2} (C^\normpow \sigma^\normpow)^{\frac{2}{\normpow} - 1}
  = C^{2 - \normpow} \sigma^\normpow.
\end{equation*}


Returning to our bound on $(\gamma_n - \gamma)$, we find via
inequality~\eqref{eqn:funny-two-power} that
\begin{align*}
  \E[(\gamma_n - \gamma)^2]
  & \le \E[|Y|^2] + \max\{n^{-2 \alpha/\normpow}, \sigma^2\} \\
  & \le \begin{cases}
    \sigma^2 +
    \max\{n^{-2 \alpha/\normpow}, \sigma^2\} & \mbox{if~} \normpow \ge 2 \\
    C^{2 - \normpow} \sigma^\normpow
    + \max\{n^{-2 \alpha/\normpow}, \sigma^2\} & \mbox{if~} \normpow < 2
  \end{cases} \\
  & \le 2 \begin{cases} \max\{n^{-2 \alpha / \normpow}, \sigma^2\}
    & \mbox{if~} \normpow \ge 2 \\
    \max\{C^{2 - \normpow} \sigma^\normpow, n^{-2\alpha/\normpow}\} & \mbox{if~} \normpow < 2,
  \end{cases}
\end{align*}
where we have used that for $\normpow < 2$ we have
\begin{equation*}
  \sigma^2 = \E[Y^\normpow]^{2/\normpow} \le \E[Y^2] \le C^{2 - \normpow} \sigma^\normpow.
\end{equation*}

In particular, we have by H\"older's inequality that
\begin{align}
  \lefteqn{\E\left[\left(1 - \frac{\frac{1}{n} \sum_{i=1}^n |Y_i|^\normpow}{
        \gamma^\normpow}\right)(\gamma_n - \gamma)\right]^2
    \le \E\left[
      \left(1 - \frac{\frac{1}{n} \sum_{i=1}^n |Y_i|^\normpow}{
        \gamma^\normpow}\right)^2\right] \E[(\gamma_n - \gamma)^2]}
  \label{eqn:holder-lambda-z} \\
  & \qquad \qquad ~ \le 2
  \left(\hinge{1 - n^\alpha \sigma^\normpow}^2 +
  C^\normpow \min\left\{\frac{\sigma^\normpow}{n^{1 - 2 \alpha}},
  \frac{1}{n \sigma^\normpow}\right\}\right)
  \cdot \begin{cases}
    \max\{n^{-2 \alpha/\normpow}, \sigma^2\} & \mbox{if~} \normpow \ge 2 \\
    \max\{n^{-2 \alpha/\normpow}, C^{2 - \normpow} \sigma^\normpow\} & \mbox{if~} \normpow < 2.
  \end{cases}
  \nonumber
\end{align}

We now state a lemma, whose proof we defer to
Section~\ref{section:proof-of-annoy-sigma}, which gives us our desired result.
\begin{lemma}
  \label{lemma:annoy-sigma}
  For any $\sigma \ge 0$, we have
  \begin{subequations}
    \begin{equation}
      \hinge{1 - n^\alpha \sigma^\normpow}^2
      \cdot \begin{cases}
        \max\{n^{-2 \alpha/\normpow}, \sigma^2\} & \mbox{if~} \normpow \ge 2 \\
        \max\{n^{-2 \alpha/\normpow}, C^{2 - \normpow} \sigma^\normpow\} & \mbox{if~} \normpow < 2.
      \end{cases}
      \le \begin{cases} n^{-2\alpha/\normpow} & \mbox{if~} \normpow \ge 2 \\
        C^{2 - \normpow} \min\{\sigma^\normpow, n^{-\alpha}\} & \mbox{if~} \normpow < 2.
      \end{cases}
      \label{eqn:annoy-sigma-a}
    \end{equation}
    and
    \begin{equation}
      \label{eqn:annoy-sigma-b}
      C^\normpow \min\left\{\frac{\sigma^\normpow}{n^{1 - 2 \alpha}},
      \frac{1}{n \sigma^\normpow}\right\}
      \cdot \begin{cases}
        \max\{n^{-2 \alpha/\normpow}, \sigma^2\} & \mbox{if~} \normpow \ge 2 \\
        \max\{n^{-2 \alpha/\normpow}, C^{2 - \normpow} \sigma^\normpow\} & \mbox{if~} \normpow < 2.
      \end{cases}
      \le \begin{cases}
        C^\normpow \frac{1}{n^{1 - \alpha + 2 \alpha / \normpow}} & \mbox{if~} \normpow \ge 2 \\
        \max\left\{\frac{C^2}{n},
        \frac{C^\normpow}{n^{1 - \alpha + 2 \alpha / \normpow}}\right\}
        & \mbox{if~} \normpow < 2.
      \end{cases}
    \end{equation}
  \end{subequations}
\end{lemma}

We now use Lemma~\ref{lemma:annoy-sigma} to give the remainder of the
proof. First, consider the case that $\normpow \ge 2$. Then choosing $\alpha = 1$
we have $\gamma^\normpow = \max\{n^{-1}, \sigma^\normpow\}$, and
\begin{equation*}
  \left|\E\left[\left(1 - \frac{\frac{1}{n} \sum_{i=1}^n |Y_i|^\normpow}{
      \gamma^\normpow}\right)(\gamma_n - \gamma)\right]\right|^2
  \le 2 \left[ C^\normpow n^{(1 - 2/\normpow) \alpha - 1} + n^{-(2/\normpow) \alpha}\right]
  = \frac{2(1 + C^\normpow)}{n^{2/\normpow}}
  \le 4 \frac{C^\normpow \vee 1}{n^{2/\normpow}}.
\end{equation*}
When $\normpow < 2$, we similarly choose $\alpha = 1$, which yields
\begin{equation*}
  \left|\E\left[\left(1 - \frac{\frac{1}{n} \sum_{i=1}^n |Y_i|^\normpow}{
      \gamma^\normpow}\right)(\gamma_n - \gamma)\right]\right|^2
  \le 2 \max\left\{\frac{C^2}{n},
  \frac{C^\normpow}{n^{2/\normpow}}\right\}
  + 2 \frac{C^{2 - \normpow}}{n}.
\end{equation*}
(Asymptotically, then, we obtain $4 \max\{C^2, C^{2 - \normpow}\} / n$.)
Referring to inequality~\eqref{eqn:convexity-with-ks}, we thus have
\begin{align*}
  \E\left[\bigg(\frac{1}{n}\sum_{i=1}^n |Y_i|^\normpow\bigg)^{1/\normpow}\right]
  & \ge \E[|Y|^\normpow]^{1/\normpow}
  - 2 \frac{\normpow-1}{\normpow} \begin{cases}
    (C^{\normpow/2} \vee 1) \cdot n^{-1/\normpow} & \mbox{if~} \normpow \ge 2 \\
    (C \vee C^{1 - \normpow/2}) \cdot n^{-1/2} & \mbox{if~} \normpow < 2,
  \end{cases}
\end{align*}
which was the desired result.

\subsubsection{Proof of Lemma~\ref{lemma:one-two-power}}
\label{sec:proof-one-two-power}

For any random variable $X$, we know that for $\gamma \in [0, 1]$
and any conjugates $p, q \ge 1$, that is, $1/p + 1/q = 1$, we have
by H\"older's inequality that
\begin{equation*}
  \E[X] = \E[X^\gamma X^{1 - \gamma}]
  \le \E[X^{\gamma p}]^{1/p} \E[X^{(1 - \gamma) q}]^{1/q}.
\end{equation*}
Now, let $X = Y^{a\normpow}$, and take
$1/p = 2 - a$ and $1/q = a - 1$. Then we have for any $\gamma \in [0, 1]$
that
\begin{equation*}
  \E[Y^{a\normpow}] \le \E[Y^{\frac{\gamma a\normpow}{2 - a}}]^{2 - a}
  \E[Y^\frac{(1 - \gamma) a\normpow}{a - 1}]^{a - 1}.
\end{equation*}
If we take $\gamma = \frac{2 - a}{a} \in [0, 1]$, then we obtain
\begin{align*}
  \frac{\gamma a}{2 - a} = 1
  ~~ \mbox{and} ~~
  (1 - \gamma) \frac{a}{a - 1} =
  \frac{2(a - 1)}{a} \frac{a}{a - 1} = 2.
\end{align*}
This gives the result of the lemma.

\subsubsection{Proof of Lemma~\ref{lemma:annoy-sigma}}
\label{section:proof-of-annoy-sigma}

We begin with inequality~\eqref{eqn:annoy-sigma-a}. If $\sigma^\normpow \ge
n^{-\alpha}$, the result is trivial, as $\hinge{1 - n^\alpha \sigma^\normpow} =
0$. So we assume that $\sigma^\normpow < n^{-\alpha}$, which
implies that $\sigma^2 \ge n^{-2 \alpha/\normpow}$,
and we know that (for $\normpow < 2$) $C^{2 - \normpow} \sigma^\normpow \ge \sigma^2$.
Thus, when $\normpow < 2$,
we have $\max\{n^{-2
  \alpha / \normpow}, C^{2 - \normpow} \sigma^\normpow\} = C^{2 - \normpow} \sigma^\normpow \le
C^{2 - \normpow} n^{-\alpha}$.  If $\normpow \ge 2$ and $\sigma^\normpow \le n^{-\alpha}$, then
$\sigma^2 \le n^{-2 \alpha / \normpow}$, so that $\max\{\sigma^2, n^{-2 \alpha
  /\normpow}\} = n^{-2 \alpha/\normpow}$.

Now we turn to inequality~\eqref{eqn:annoy-sigma-b}. First, let us assume
that $\normpow \ge 2$. In this case, we have that if $\sigma^\normpow \le n^{-\alpha}$,
then the left-hand expression of~\eqref{eqn:annoy-sigma-b} has bound
\begin{equation*}
  C^\normpow \min\left\{\frac{\sigma^\normpow}{n^{1 - 2\alpha}},
    \frac{1}{n \sigma^\normpow} \right\} n^{-2 \alpha / \normpow}
  = C^\normpow \frac{\sigma^\normpow}{n^{1 - 2 \alpha + 2 \alpha/\normpow}}
  \le C^\normpow \frac{1}{n^{1 - \alpha + 2 \alpha/\normpow}}.
\end{equation*}
On the other hand, for $\sigma^\normpow \ge n^{-\alpha}$, we have
\begin{equation*}
  C^\normpow \min\left\{\frac{\sigma^\normpow}{n^{1 - 2\alpha}},
    \frac{1}{n \sigma^\normpow} \right\} \sigma^2
  = C^\normpow \frac{\sigma^2}{n \sigma^\normpow}
  = C^\normpow
  \frac{1}{n \sigma^{\normpow - 2}}
  \le C^\normpow \frac{1}{n^{1 - \alpha + 2 \alpha/\normpow}},
\end{equation*}
as $\normpow \ge 2$ and $\sigma \ge n^{-\alpha/\normpow}$.
In the case that $\normpow < 2$ in inequality~\eqref{eqn:annoy-sigma-b},
we are left bounding
\begin{equation*}
  \min\left\{\frac{\sigma^\normpow}{n^{1 - 2\alpha}},
    \frac{1}{n \sigma^\normpow} \right\} \max\{n^{-2 \alpha/\normpow},
  C^{2 - \normpow} \sigma^\normpow\}.
\end{equation*}
Assume first that $n^{-2 \alpha/\normpow} \ge C^{2 - \normpow} \sigma^\normpow$,
or $\sigma^\normpow \le C^{\normpow - 2} n^{-2 \alpha/\normpow}$. In this case,
the $\sigma$ maximizing the left minimum is $\sigma^\normpow =
\min\{n^{-\alpha}, C^{\normpow - 2} n^{-2 \alpha/\normpow}\}$,
which gives
\begin{equation*}
  \min\left\{\frac{\sigma^\normpow}{n^{1 - 2\alpha}},
    \frac{1}{n \sigma^\normpow} \right\} \max\{n^{-2 \alpha/\normpow},
  C^{2 - \normpow} \sigma^\normpow\}
  \le \frac{1}{n^{1 - \alpha + 2 \alpha / \normpow}}.
\end{equation*}
On the other hand, when $C^{2 - \normpow} \sigma^\normpow \ge n^{-2 \alpha/\normpow}$,
we obtain that we must maximize (over $\sigma$) the quantity
\begin{equation*}
  C^2 \min\left\{\frac{\sigma^{2\normpow}}{n^{1 - 2 \alpha}},
    \frac{1}{n} \right\}
  \le C^2 \frac{1}{n}.
\end{equation*}
This gives the desired result.

\subsection{Proof of Corollary~\ref{corollary:concentration-uniform}}
\label{section:proof-of-concentration-uniform}

Let
$\fclass \defeq \{ \loss(\param; \cdot): \statdomain \to \R
~~\mbox{for}~~ \param \in \Theta \}$ be our function class. Fix
$t > 0$ and let $N = \covnum(\frac{\epsilon_{t, n}}{3}, \fclass, \linfstatnorm{\cdot})$ to
ease notation, so there exists
$\{\param_1, \cdots, \param_N\} \subset \Theta$ such that
$\{ \loss(\param_1; \cdot), \cdots, \loss(\param_N; \cdot) \}$ is a
$\frac{\epsilon_{t, n}}{3}$-cover of $\fclass$. For any
$\param \in \Theta$, let $i(\param)$ be such that
$\linfstatnorm{\loss(\param; \cdot) - \loss(\param_{i(\param)}; \cdot)} \le
\frac{\epsilon_{t, n}}{3}$.
We have
\begin{align*}
  \lefteqn{\sup_{\param \in \Theta}
    \left| \risk_k(\param; \emp) - \risk_k(\param; P_0) \right|} \\
  & \le
  \sup_{\param \in \Theta}
  \left\{
  \left| \risk_k(\param; \emp) - \risk_k(\param_{i(\param)}; \emp) \right| 
  + \left|\risk_k(\param_{i(\param)}; \emp)
  - \risk_k(\param_{i(\param)}; P_0) \right|
  + \left|\risk_k(\param_{i(\param)}; P_0) - \risk_k(\param; P_0) \right|
  \right\} \\
  & \le
  \max_{i = 1, \ldots, N}
  \left|\risk_k(\theta_i; \emp) - \risk_k(\theta_i; P_0)\right|
  + \frac{2\epsilon_{t, n}}{3},
\end{align*}
where we have used that $\{\loss(\theta_i; \cdot)\}_{i=1}^N$ is a $\epsilon_{t, n} /3$
cover of $\mc{F}$.
A union bound now implies
\begin{align*}
  \P\left(\sup_{\param \in \Theta}
  \left| \risk_k(\param; \emp) - \risk_k(\param; P_0) \right| \ge \epsilon_{t, n}\right)
  & \le 
  N \max_{i=1, \ldots, N} \P\left(
  \left| \risk_k(\param_i; \emp) - \risk_k(\param_i; P_0) \right|
  \ge \epsilon_{t, n} / 3\right).
\end{align*}
Applying Theorem~\ref{theorem:concentration} to each $\param_i$, we obtain
the desired result.


  

%% file: lower-proof.tex
\section{Proof of Lower Bounds}
\label{section:proof-lower}

\newcommand{\minimaxest}{\mathfrak{M}_n(\mc{P}, f)}
\newcommand{\minimaxopt}{\mathfrak{M}_n(\mc{P}, f, \ell)}

All results in Section~\ref{section:lower} can be alternatively stated as a
probabilistic lower bound on the estimation or optimization error
\begin{align*}
  P_0\left( |\what{R}(Z_1^n) -
  \risk_f(Z)| \ge n^{-\frac{1}{2\vee k_*}} \right),~\mbox{or}~P_0\left(
    \risk_f\left(\what{\theta}(X_1^n); P_0\right)
    - \inf_{\theta \in \Theta} \risk_f\left(\theta; P_0\right)
    \ge n^{-\frac{1}{2\vee k_*}}
  \right).
\end{align*}
These results follow by using the below identical proofs by noting that
\begin{equation*}
  \inf_{\what{\theta}(X_1^n)}
  \sup_{P_0 \in \mc{P}}
  P_0\left( |\what{R}(Z_1^n) -
  \risk_f(Z)| \ge \delta \right)
  \ge \frac{1}{2} \left(1 - \tvnorm{P_1^n - P_2^n}\right)
\end{equation*}
whenever $|\risk_f(Z_1) - \risk_f(Z_2) | \ge 2\delta$ for $Z_1 \sim P_1$ and
$Z_2 \sim P_2$ (and similarly for the optimization error).

In the coming proofs related to Section~\ref{section:lower-estimation}, we
define
\begin{equation*}
  \minimaxest
  \defeq \inf_{\what{R}} \sup_{P_0 \in \mc{P}}
  \E_{P_0}\left[ \left| \what{R}(Z_1^n) - \risk_k(Z) \right|\right]
\end{equation*}
for shorthand, and use it without comment.

\subsection{Proof of Theorem~\ref{theorem:lower-estimation}}
\label{section:proof-of-lower-estimation}

Consider the canonical
two point hypothesis testing problem between distributions $P_0$ and $P_1$:
nature first chooses $v \in \{0, 1\}$, then
conditioned on $v$ draws $Z_1, \ldots, Z_n \simiid P_v$.
Assuming that
$| \risk_k(P_0) - \risk_k(P_{1})| \ge 2\delta > 0$ for some $\delta$,
Le Cam's classical
reduction from estimation to testing~\cite{LeCamYa00,Yu97} yields that
\begin{equation}
  \label{eqn:reduction}
  \minimaxest \ge \frac{\delta}{2}
  \left(1 - \tvnorm{P_0^n - P_1^n}\right).
\end{equation}
We use the bound~\eqref{eqn:reduction} to give the lower bound by
choosing $P_0$ and $P_1$ so that $\tvnorm{P_0^n - P_1^n} \le \half$
and $\delta$ is as large as possible.

First, we show the $O(n^{-\half})$ lower bound. We begin with a technical
\begin{lemma}
  \label{lemma:two-point-robust}
  Let $c_k = (1 + k(k-1) \tol)^{\frac{1}{k}}$,
  $p_k = c_k^{-k/(k-1)}$, and
  $\beta_k \defeq \half (1 - c_k^{-k})$.  
  For a pair $z_0 \le z_1$, let $Z$ be such that
  \begin{equation*}
    Z = 
    \begin{cases}
      z_0 & \mbox{w.p.~} 1-p \\
      z_1 & \mbox{w.p.~} p.
    \end{cases}
  \end{equation*}
  If $p \ge p_k$, we have $\risk_k(Z) = z_1$, and if $p \le p_k$, we
  have $\risk_k(Z) \le c_k p^{\frac{1}{k_*}} z_1 + (1-c_k p^{\frac{1}{k_*}})z_0$.
  Further, if $p \le p_k \wedge (1 - (1-\beta)^{1-k_*}p_k)$ for some
  $\beta \in (0, 1)$, then
  $\risk_k(Z) \ge \beta^{\frac{1}{k}} c_k p^{\frac{1}{k_*}} z_1 + (1-
  \beta^{\frac{1}{k}} c_k p^{\frac{1}{k_*}})z_0$.
\end{lemma}
\noindent
See Section~\ref{subsection:proof-of-two-point-robust} for a proof.

Now, consider the two distributions $Z_1 \sim P_1$, $Z_2 \sim P_2$
\begin{align*}
  Z_1 = 
  \begin{cases}
    0 & \mbox{w.p.~} 1-p_k - \delta \\
    \zbound & \mbox{w.p.~} p_k + \delta
  \end{cases}
        ,~~~~~~
        Z_2 = 
        \begin{cases}
          0 & \mbox{w.p.~} 1-p_k + \delta \\
          \zbound & \mbox{w.p.~} p_k - \delta
        \end{cases}
\end{align*}
for some $0 < \delta \le p_k \wedge (1-p_k)$ to be chosen later. Note that
$p_k = c_k^{-k_*} < 1$ as $c_k > 1$. We use the version of $Z_1$ and $Z_2$
such that $Z_1(\cdot)$ and $Z_2(\cdot)$ are upper semi-continuous. 

From Lemma~\ref{lemma:two-point-robust}, we have that $\risk_k(Z_1) = \zbound$
and $\risk_k(Z_2) \le \zbound c_k(p_k -
\delta)^{\frac{1}{k_*}}$. Consequently, $P_1$ and $P_2$ are separated in the
robust objective
\begin{equation*}
  \left| \risk_k(Z_1) - \risk_k(Z_2) \right|
  \ge \zbound(1 - c_k(p_k - \delta)^{\frac{1}{k_*}})
  \ge \frac{c_k^{k_*}}{k_*} \zbound \delta
\end{equation*}
where we used Taylor's theorem
\begin{equation*}
  c_k(p_k - \delta)^{\frac{1}{k_*}}
  = c_k (c_k^{-k_*} - \delta)^{\frac{1}{k_*}}
  \le c_k \left(c_k^{-1} - \frac{1}{k_*} c_k^{\frac{k_*}{k}} \delta\right)
  = 1 - \frac{1}{k_*} c_k^{k_*} \delta.
\end{equation*}

It suffices to show that $\tvnorms{P_1^n - P_2^n} \le \half$ for
$\delta = \sqrt{\frac{p_k(1-p_k)}{8n}} \wedge \half(1-p_k) 
\wedge p_k$.  By Pinsker's inequality, we have
$\tvnorms{P_1^n - P_2^n}^2 \le \frac{n}{2} \dkl{P_2}{P_1}$ so it is enough to
show $\dkl{P_2}{P_1} \le \frac{1}{n}$ for the given value of $\delta$. To this
end, we note that for $\delta \le \half (1- p_k)$,
\begin{align*}
  \dkl{P_2}{P_1} 
  = (1-p_k + \delta) 
  \log \frac{ 1-p_k + \delta}{1-p_k - \delta}
  + (p_k - \delta) \log \frac{p_k - \delta}{p_k + \delta} 
  \le \frac{8 \delta^2}{p_k (1-p_k)}.
\end{align*}
Setting $\delta = \sqrt{\frac{p_k(1-p_k)}{8n}} \wedge \half(1-p_k) \wedge
p_k$, we then have that $\dkl{P_2}{P_1} \le \frac{1}{n}$.

For the second $O(n^{-\frac{1}{k_*}})$ bound, consider the random
variables $Z_1 \sim P_1$ and $Z_2 \sim P_2$ with
\begin{align*}
  Z_1 \equiv 0,
  ~~~~~~
  Z_2 = 
  \begin{cases}
    0 & \mbox{w.p.~} 1 - \delta \\
    \zbound & \mbox{w.p.~} \delta
  \end{cases}
\end{align*}
for some $\delta > 0$ to be choosen later.
We have $\risk_k(Z_1) = 0$ trivially, and since
$1 - (1-\beta)^{1-k_*}p_k > 0 \equiv 1 - c_k^{-k} > \beta$ holds for
$\beta_k = \half(1 - c_k^{-k})$, we have
\begin{equation*}
  \risk_k(Z_2) \ge \zbound \beta_k^{\frac{1}{k}} c_k \delta^{\frac{1}{k_*}}
\end{equation*}
for $0 < \delta \le p_k \wedge ( 1 - (1-\beta_k)^{1-k_*} p_k )$
by Lemma~\ref{lemma:two-point-robust}. This gives the the separation
$|\risk_k(P_1) - \risk_k(P_2)| \ge \zbound \beta_k^{\frac{1}{k}} c_k
\delta^{\frac{1}{k_*}}$.

Noting that
\begin{equation*}
  \dkl{P_1}{P_2} = - \log (1 - \delta) \le \frac{\delta}{1-\delta}
  \le 2 \delta
\end{equation*}
for $\delta \le \half$,
we obtain
\begin{align*}
  \minimaxest
  & \ge \frac{1}{4} \zbound \beta_k^{\frac{1}{k}} c_k \delta^{\frac{1}{k_*}}
  \left(1 - \sqrt{\frac{n}{2}
      \dkl{P_1}{P_2}} \right)
  \ge \frac{1}{8} \zbound c_k \beta_k^{\frac{1}{k}} \delta^{\frac{1}{k_*}}
\end{align*}
where in the first inequality we used
the reduction~\eqref{eqn:reduction} and Pinsker's inequality as
before.  The desired result follows by setting
$\delta = \frac{1}{4n} \wedge p_k \wedge (1 - (1-\beta_k)^{1-k_*} p_k)$.

\subsubsection{Proof of Lemma~\ref{lemma:two-point-robust}}
\label{subsection:proof-of-two-point-robust}

Define the objective function in the dual
representation~\eqref{eqn:cressie-read-risk} as
\begin{equation*}
  g(\eta) \defeq c_k \left( (1-p) \hinge{z_0-\eta}^{k_*} +
    p \hinge{z_1 - \eta}^{k_*} \right)^{\frac{1}{k_*}} + \eta.
\end{equation*}
Taking subgradients, we obtain
\begin{align*}
  \partial g(\eta) = \begin{cases}
    1 & \mbox{if~} \eta > z_1 \\
    [1- c_k p^{\frac{1}{k_*}}, 1] & \mbox{if~} \eta = z_1 \\
    1 - c_k p^{\frac{1}{k_*}} & \mbox{if~} z_0 \le \eta < z_1 \\
    1 - c_k \frac{(1-p) (z_0 - \eta)^{\frac{1}{k-1}} + p (z_1 -
      \eta)^{\frac{1}{k-1}}}{\left( (1-p) (z_0-\eta)^{k_*} + p (z_1 -
        \eta)^{k_*} \right)^{\frac{1}{k}}}
    & \mbox{if~} \eta < z_0.
  \end{cases}
\end{align*}
If $c_kp^{\frac{1}{k_*}} \ge 1$ then $\eta^* = \argmin_{\eta} g(\eta)$ is
attained at $z_1$ by convexity, and $R(P) = g(\eta^*) = z_1$. If
$c_kp^{\frac{1}{k_*}} < 1$, we have $\eta^* \le z_0$ so that
\begin{equation*}
  g(\eta^*) \le g(z_0) = c_kp^{\frac{1}{k_*}} z_1 
  + (1 - c_kp^{\frac{1}{k_*}})z_0,
\end{equation*}
which gives the second claim.

For the second inequality, noting that
\begin{equation*}
  \risk_k(Z) = z_0 + (z_1-z_0) \sup\left\{ q \in [0, 1]: 
    (1-p)^{1-k} (1-q)^k  + p^{1-k} q^k \le c_k^k \right\},
\end{equation*}
it suffices to show that $q = \beta^{\frac{1}{k}} c_k p^{\frac{1}{k_*}}$ is
feasible when $p \le 1 - (1-\beta)^{1-k_*} p_k$. Indeed, we have
\begin{equation*}
  (1-p)^{1-k} (1- \beta^{\frac{1}{k}} c_k p^{\frac{1}{k_*}})^k 
  + p^{1-k} (\beta^{\frac{1}{k}} c_k p^{\frac{1}{k_*}})^k
  \le (1-p)^{1-k} + \beta c_k^k
  \le c_k^k
\end{equation*}
where we used $(1-p)^{1-k} \le (1-\beta)c_k^k$ in the last inequality.

\subsection{Proof of
 Proposition~\ref{proposition:lower-general-estimation-sqrt-new}}
\label{section:proof-of-lower-general-estimation-sqrt-new}

We proceed by LeCam's method as in Theorem~\ref{theorem:lower-estimation}. Let
$Z_1 \sim P_1$, $Z_2 \sim P_2$ have distribution
\begin{align*}
  Z_1 = 
  \begin{cases}
    0 & \mbox{w.p.~} 1-p \\
    \zbound & \mbox{w.p.~} p,
  \end{cases}
  ~~~~
  Z_2 = 
  \begin{cases}
    0 & \mbox{w.p.~} 1-p-\delta  \\
    \zbound & \mbox{w.p.~} p+\delta
  \end{cases}
\end{align*}
for some $\delta \in (0, 1)$ to be chosen later. As before, we show
that $\risk_f(Z_1)$ and $\risk_f(Z_2)$ are well-separated but $P_1$ and $P_2$
are close in total variation distance.

By definition, we have
\begin{align*}
  \risk_f(Z_1) = \sup\left\{
  q \zbound : h_f(q; p) \le \tol, q \in [0, 1]
  \right\} = \zbound q(p)
\end{align*}
and similarly, $\risk_f(Z_2) = \zbound q(p+\delta)$.  For $\delta$ small
enough, the implicit function theorem applies to $h_f(q(p), p) = 0$ by our
hypothesis.  Consequently, we $q(\cdot)$ is
continuously differentiable on a neighbhorhood of $p$ with
\begin{equation*}
  q'(p) = -\frac{\partial_p h_f(q(p); p)}{\partial_q h_f(q(p); p)} > 0,
\end{equation*}
where strict positivity follows by the strict convexity the we assume
in the proposition.
Taylor's theorem  implies
\begin{equation*}
  \risk_f(Z_2) - \risk_f(Z_1) = q(p+\delta) - q(p)
  = q'(p) \delta + o(\delta)
\end{equation*}
as $\delta \to 0$.

We now pick $\delta$ such that $\tvnorms{P_1^n - P_2^n} \le \half$.  By
Pinsker's inequality and standard KL vs.\ $\chi^2$-divergence
inequalities~\cite[Lemmas 2.5--2.7]{Tsybakov09}, we have $\tvnorms{P_1^n -
  P_2^n}^2 \le \frac{n}{2} \dkl{P_1}{P_2}$; we will choose $\delta$ such
that $\dkl{P_1}{P_2} \le \frac{1}{n}$. For $\delta \in [0, p]$, Lemma 2.7 of
\cite{Tsybakov09} yields
\begin{align*}
  \dkl{P_1}{P_2}
  \le \frac{\delta^2}{p} + \frac{\delta^2}{1 - p}
  = \frac{\delta^2}{p(1 - p)}.
\end{align*}
Setting $\delta_n = \sqrt{\frac{p(1-p)}{n}}$, we obtain from the reduction
from estimation to hypothesis testing~\eqref{eqn:reduction} that
\begin{equation*}
  \minimaxest
  \ge \frac{\zbound}{8}
  q'(p) \sqrt{\frac{p(1-p)}{n}} + o\left(\frac{1}{\sqrt{n}} \right),
\end{equation*}
which gives the result.

\subsection{Proof of
  Proposition~\ref{proposition:lower-general-estimation-kconj}}
\label{section:proof-of-lower-general-estimation-kconj}

We use LeCam's method and proceed similarly as in the second part of
Section~\ref{section:proof-of-lower-estimation}.  Consider the two
distributions $Z_1 \sim P_1$, $Z_2 \sim P_2$ with
\begin{align*}
  Z_1 \equiv 0,
  ~~~~~~
  Z_2 = 
  \begin{cases}
    0 & \mbox{w.p.~} 1 - \delta \\
    \zbound & \mbox{w.p.~} \delta,
  \end{cases}
\end{align*}
where we set $\delta = \frac{1}{2(n \vee C_{f, \tol, m})}$. Then
$\risk_f(Z_1) = 0$, and to show separation of $\risk_f(Z_2)$, we require
a bit of work, beginning with the following lemma.
\begin{lemma}
  \label{lemma:q-included}
  For $\delta = \frac{1}{2(n \vee C_{f, \tol, m})}$, define $Q$ by
  $Q(Z = \zbound) = \left(\frac{\tol}{2m}\right)^{\frac{1}{k}}
  \delta^{\frac{1}{k_*}}$ and $Q(Z = 0) = 1-Q(Z = \zbound)$. Then
  $\fdiv{Q}{P_2} \le \tol$.
\end{lemma}
\begin{proof}
  We have
  \begin{align*}
    & \delta f\left( \frac{\left(\frac{\tol}{2m}\right)^{\frac{1}{k}}
      \delta^{\frac{1}{k_*}}}{\delta} \right)
    + (1-\delta) f\left( \frac{1-\left(\frac{\tol}{2m}\right)^{\frac{1}{k}}
      \delta^{\frac{1}{k_*}}}{1-\delta} \right) \\
    & \stackrel{(a)}{\le} \delta f\left(\left(\frac{\tol}{2m}\right)^{\frac{1}{k}}
    \delta^{-\frac{1}{k}} \right)
    + (1-\delta) f\left( 1 - \left(\frac{\tol}{2m}\right)^{\frac{1}{k}}
    \delta^{\frac{1}{k_*}} \right)
    \stackrel{(b)}{\le} \delta f\left(\left(\frac{\tol}{2m}\right)^{\frac{1}{k}}
    \delta^{-\frac{1}{k}} \right)
    + \frac{\tol}{2}
  \end{align*}
  where in step $(a)$, we used that $f$ is non-increasing on $(0, 1)$ along
  with
  $\frac{1-\left(\frac{\tol}{2m}\right)^{1/k} \delta^{1/k_*}}{1-\delta}
  \in (0, 1)$, and in step $(b)$, we used the definition of $f^{-1}(s)
  = \inf\{t \in [0, 1] : f(t) \le s\}$.

  Next, note that since
  $\left(\frac{\tol}{2m}\right)^{\frac{1}{k}} \delta^{-\frac{1}{k}} \ge \left\{
  (n \vee C_{f, \tol, m}) \tol m^{-1} \right\}^{\frac{1}{k}}$ for the given
  range of $\delta$, we have
  $f((\frac{\tol}{2m})^{1/k} \delta^{-1/k})
  \le \frac{\tol}{2 \delta}$ by hypothesis. We conclude that
  $\fdiv{Q}{P_2} \le \tol$.
\end{proof}

As a consequence of Lemma~\ref{lemma:q-included}, we have
$\risk_f(Z_2) \ge \zbound ( \frac{\tol}{2m})^{1/k}
\delta^{1 / k_*}$. As $\risk_f(Z_1) = 0$, we have
$|\risk_f(Z_1) - \risk_f(Z_2)| \ge \zbound
(\frac{\tol}{2m})^{1/k} \delta^{1/k_*}$. Proceeding
similarly as in the last paragraph of
Section~\ref{section:proof-of-lower-estimation} we obtain the result.

\subsection{Proof of Theorem~\ref{theorem:lower-optimization}}
\label{section:proof-of-lower-optimization}

Define the optimization distance between two distributions $P_0$ and $P_1$
(cf.~\cite{AgarwalBaRaWa12,Duchi18}) by
\begin{align*}
  \distopt{P_0}{P_{1}}{f} \defeq 
  \sup \left\{ \delta \ge 0:
  \begin{aligned}
    & \risk_f(\param; P_0) \le \risk_f(\param_0^*; P_0) + \delta
    \mbox{~~~implies~}
      \risk_f(\param; P_{1}) \ge \risk_f(\param_{1}^*; P_{1}) + \delta \\
      & \risk_f(\param; P_{1}) \le \risk_f(\param_{1}^*; P_{1}) + \delta
    \mbox{~implies~}
    \risk_f(\param; P_{0}) \ge \risk_f(\param_{0}^*; P_{0}) + \delta
  \end{aligned}
  \right\}
\end{align*}
where
$\param_v \in \argmin_{\param \in \paramdomain} \risk_f(\param; P_v)$.
With this result, we have the following standard lemma, which is
a reduction of optimization to testing.

We have the following reduction from distributionally robust optimization to
hypothesis testing, which is based on Le Cam's two-point hypothesis
testing reduction.
\begin{lemma}[{\citet[Chs.~5.1--5.2]{Duchi18}}]
  \label{lemma:reduction-robust}
  If $P_1, P_2 \in \mc{P}$ are such that
  $\distopt{P_1}{P_2}{f} \ge \delta$, then
  \begin{align*}
    \minimax_n(\mc{P}, f, \loss)
    & \ge \delta \inf_{\what{\theta}_n}
    \sup_{P_0 \in \mc{P}}
    P_0\left( \risk_f\left(\what{\theta}_n(X_1^n); P_0\right)
      - \inf_{\theta \in \Theta} \risk_f\left(\theta; P_0\right)
      \ge \delta \right) \\
    & \ge \frac{\delta}{2} \left(1 - \tvnorm{P_1^n - P_2^n}\right).
  \end{align*}
\end{lemma}

With this inequality in hand,
we proceed by
We first show the $\Omega(n^{-\half})$ lower bound. Consider the two
distributions $X_1 \sim P_1$, $X_2 \sim P_2$ with
\begin{align*}
  X_1 = 
  \begin{cases}
    -1 & \mbox{w.p.~} 1-p_k - \delta \\
    \epsilon & \mbox{w.p.~} p_k + \delta,
  \end{cases}
  ~~~~~~
  X_2 = 
  \begin{cases}
    -1 & \mbox{w.p.~} 1-p_k + \delta \\
    \epsilon & \mbox{w.p.~} p_k - \delta
  \end{cases}
\end{align*}
where $\epsilon = \frac{\delta}{2k_*p_k}$ for some
$0 < \delta \le p_k \wedge (1-p_k)$ to be choosen later.
Note that
\begin{align*}
  \risk_k(\param; P) = 
  \begin{cases}
    \param 
    \sup_{Q \ll P} \left\{\E_Q[X]: \fdiv{Q}{P} \le \rho \right\}
    & \mbox{if~} \param \ge 0 \\
    \param
    \inf_{Q \ll P} \left\{\E_Q[X]: \fdiv{Q}{P} \le \rho \right\} &
    \mbox{if~} \param < 0.
  \end{cases}
\end{align*}
For $\delta \le 1 - 2 p_k$, we from Lemma~\ref{lemma:two-point-robust}
that
$\risk_k(\param; P_1) = -\param \indic{\param < 0} + \epsilon \param
\indic{\param \ge 0}$ and $\risk_k(\param; P_2) = -\param$ when
$\param < 0$. Now, we have $\risk_k(\param; P_2) \le -\epsilon \param$ when
$\param \ge 0$ since
\begin{align}
  \sup_{Q \ll P_2} \left\{\E_Q[X_2]: \fdiv{Q}{P_2} \le \rho \right\}
  & \le \epsilon c_k (p_k - \delta)^{\frac{1}{k_*}} 
    + (c_k (p_k - \delta)^{\frac{1}{k_*}} - 1) \nonumber \\
  & \le \epsilon c_k (p_k - \delta)^{\frac{1}{k_*}} 
    - \frac{\delta}{k_* p_k}
    \le \epsilon - \frac{\delta}{k_* p_k}
    = -\epsilon. \label{eqn:robust-sup-bound}
\end{align}
Here, we used Taylor's theorem
\begin{equation*}
  c_k(p_k - \delta)^{\frac{1}{k_*}}
  = c_k (c_k^{-k_*} - \delta)^{\frac{1}{k_*}}
  \le c_k \left(c_k^{-1} - \frac{1}{k_*} c_k^{\frac{k_*}{k}} \delta\right)
  = 1 - \frac{1}{k_*} c_k^{k_*} \delta.
\end{equation*}
If we let $\param_i\opt \defeq \argmin_{\param \in \Theta} \risk_k(\param; P_i)$
for $i = 1, 2$, we have $\param_1\opt = 0$, $\param_2\opt = \zbound$ and
$\risk_k(\param_1\opt; P_1) = 0$,
$\risk_k(\param_2\opt; P_2) \le -\zbound \epsilon$.
We then have the following lemma.

\begin{lemma}
  Let the above conditions hold. Then
  $\distopt{P_1}{P_2}{f_k} \ge \frac{\epsilon}{2} \zbound$.
\end{lemma}
\begin{proof}
  Let $\param \in [-\zbound, \zbound]$ be such that
  $\risk_k(\param; P_1) \le \risk_k(\param_1\opt; P_1) + \zbound \kappa$ for some
  $\kappa \in [0, \frac{\epsilon}{2}]$. From
  $\risk_k(\param; P_1) - \risk_k(\param_1\opt; P_1) = \risk_k(\param; P_1) =
  -\param \indic{\param < 0} + \epsilon \param \indic{\param > 0} \le \zbound
  \kappa$, we have
  $-\kappa \le \frac{\param}{\zbound} \le \frac{\kappa}{\epsilon}$. Applying
  this bound, we obtain
  \begin{align*}
    \risk_k(\param; P_2) - \risk_k(\param_2\opt; P_2)
    & = \begin{cases}
      (\theta - \zbound)
      \sup_{Q \ll P_2} \left\{\E_Q[X_2]: \fdiv{Q}{P_2} \le \rho \right\}
      & ~~\mbox{if}~~\theta \ge 0 \\
      - \theta - \zbound
      \sup_{Q \ll P_2} \left\{\E_Q[X_2]: \fdiv{Q}{P_2} \le \rho \right\}
      & ~~\mbox{if}~~\theta < 0 
    \end{cases} \\
    & \ge  -\param \indic{\param < 0} - \epsilon \theta \indic{\param \ge 0}
    + \zbound \epsilon\\
    & \ge  -\param \indic{\param < 0} - \zbound \kappa \indic{\param \ge 0}
    + \zbound \epsilon 
    \ge \frac{\zbound \epsilon}{2} \ge \zbound \kappa
  \end{align*}
  where we used the bound~\eqref{eqn:robust-sup-bound} to get the second
  inequality.

  On the other hand, assume
  $\risk_k(\param; P_2) \le \risk_k(\param_2\opt; P_2) + \zbound \kappa$. In
  this case, we claim that $\param \ge 0$ necessarily. Indeed, if
  $\param < 0$, then using the bound~\eqref{eqn:robust-sup-bound}, 
  \begin{equation*}
    \risk_k(\param; P_2) = -\theta
    \le \risk_k(\param_2\opt; P_2) + \zbound \kappa
    \le - \zbound \epsilon + \zbound \kappa
    = - \zbound  (\epsilon - \kappa) < 0
  \end{equation*}
  which yields a contradiction. Now, from $\param \ge 0$ and
  $\risk_k(\param; P_2) \le \risk_k(\param_2\opt; P_2) + \zbound \kappa$, we
  again obtain from the bound~\eqref{eqn:robust-sup-bound}
  \begin{equation*}
    \zbound \kappa \ge (\param - \param_2\opt) 
    \sup_{Q \ll P_2} \left\{\E_Q[X]: \fdiv{Q}{P_2} \le \rho \right\}
    \ge \epsilon (\zbound - \param).
  \end{equation*}
  Hence, we have
  $\param \ge \zbound \left(1 - \frac{\kappa}{\epsilon}\right)$, and
  \begin{equation*}
    \risk_k(\param; P_1) = \epsilon \param \ge \epsilon \zbound
    \left(1- \frac{\kappa}{\epsilon}\right) 
    = \zbound (\epsilon - \kappa)
    \ge \frac{\zbound \epsilon}{2}
    \ge \risk_k(\param_1\opt; P_1) + \zbound \kappa
  \end{equation*}
  for $\kappa \in [0, \frac{\epsilon}{2}]$. We conclude that the claimed
  separation in $d_{\rm opt}$ holds.
\end{proof}

Now, we argue as in the proof of
Theorem~\ref{theorem:lower-estimation}. Noting that $\dkl{P_1}{P_2} \le
\frac{\delta^2}{p_k (1-p_k)}$ (e.g.~\cite[Lemma 2.7]{Tsybakov09}) for $0 \le
\delta \le (1-p_k)$, let $\delta = \sqrt{\frac{p_k(1-p_k)}{2 n}} \wedge
\half(1-p_k)\wedge (1 - 2p_k) \wedge p_k$. Then
Lemma~\ref{lemma:reduction-robust} yields
\begin{align*}
  \minimax_n(\mc{P}, f_k, \loss)
  \ge \frac{\zbound \epsilon}{4}
  \left(1 - \sqrt{\frac{n}{2} \dkl{P'_1}{P'_2}} \right)
  \ge \frac{\zbound \delta}{8 k_* p_k},
\end{align*}
which gives the first result of the theorem.

Next, we show the second $\Omega(n^{-\frac{1}{k_*}})$ lower
bound. Consider the distributions $X_1 \sim P_1$, $X_2 \sim P_2$
\begin{align*}
  X_1 \equiv -\epsilon,
  ~~~~~~
  X_2 = 
  \begin{cases}
    -\epsilon & \mbox{w.p.~} 1-\delta \\
    1  & \mbox{w.p.~} \delta
  \end{cases}
\end{align*}
where $\epsilon \defeq \half \beta_k^{\frac{1}{k}} c_k
\delta^{\frac{1}{k_*}}$ for some $0 < \delta \le p_k \wedge (1-p_k) \wedge
(1 - (1-\beta_k)^{1-k_*} p_k)$ to be choosen later. Now, we again show that
$\distopt{P_1}{P_2}{f_k} \ge \frac{\epsilon}{2}$. To this end, first observe
that $\risk_k(\param; P_1) = - \epsilon \param$. From the first part of
Lemma~\ref{lemma:two-point-robust}, we have $\risk_k(\param; P_2) =
-\epsilon \param \ge 0$ when $\param < 0$.  For $\param \ge 0$, the last
inequality in Lemma~\ref{lemma:two-point-robust} gives
\begin{equation*}
  \risk_k(\param; P_2) \ge \beta_k^{\frac{1}{k}} c_k \delta^{\frac{1}{k_*}} \param 
  - (1- \beta_k^{\frac{1}{k}} c_k \delta^{\frac{1}{k_*}}) \epsilon \param
  = \left((1+\epsilon) \beta_k^{\frac{1}{k}} c_k \delta^{\frac{1}{k_*}} 
    - \epsilon\right) \param \ge \epsilon \param
\end{equation*}
since $\epsilon = \half \beta_k^{\frac{1}{k}} c_k \delta^{\frac{1}{k_*}}$.
Denoting $\param_i\opt \defeq \argmin_{\param \in \Theta} \risk_k(\param; P_i)$
again, we consequently obtain $\param_1\opt = \zbound$, $\param_2\opt = 0$ with
$\risk_k(\param_1\opt; P_1) = -\zbound \epsilon$, $\risk_k(\param_2\opt; P_2) = 0$.

Next, we show $\distopt{P_1}{P_2}{f_k} \ge \frac{\zbound \epsilon}{2}$. Assume
that $\param \in [-\zbound, \zbound]$ satisfies
$\risk_k(\param; P_1) \le \risk_k(\param_1\opt; P_1) + \zbound \kappa =
-\zbound\epsilon + \zbound \kappa ~\equiv ~ \param \ge \zbound \left(1 -
  \frac{\kappa}{\epsilon}\right)$ for some
$\kappa \in [0, \frac{\epsilon}{2}]$. This implies
\begin{equation*}
  \risk_k(\param; P_2) 
  \ge \zbound \epsilon \left( 1 - \frac{\kappa}{\epsilon}\right)
  \ge \zbound \frac{\epsilon}{2}
  = \risk_k(\param_2\opt; P_2) + \frac{\zbound \epsilon}{2} 
  \ge \risk_k(\param_2\opt; P_2) + \zbound \kappa.
\end{equation*}
On the other hand, if
$\risk_k(\param; P_2) \le \risk_k(\param_2\opt; P_2) + \zbound \kappa$ then
$\epsilon |\param| \le \risk_k(\param; P_2) \le \zbound \kappa$ so that
$|\param| \le \frac{\zbound \kappa}{\epsilon}$. Consequently, we have
\begin{align*}
  \risk_k(\param; P_1) & = -\epsilon \param
  \ge -\zbound \kappa = \zbound (-\epsilon + \epsilon - \kappa) \\
  & \ge \zbound \left(-\epsilon + \frac{\epsilon}{2} \right) 
  = \risk_k(\param_1\opt; P_1) + \frac{\zbound \epsilon}{2}
  \ge \risk_k(\param_1\opt; P_1) + \zbound \kappa
\end{align*}
and we conclude $\distopt{P_1}{P_2}{f_k} \ge \frac{\zbound \epsilon}{2}$.

Proceeding as in the proof of the second part of
Theorem~\ref{theorem:lower-estimation}, we note that $\dkl{P_1}{P_2} \le
2\delta$ when $\delta \le \half$. Setting $\delta = \frac{1}{4n} \wedge p_k
\wedge (1 - (1-\beta_k)^{1-k_*} p_k)$, we conclude
\begin{equation*}
  \minimaxopt \ge \frac{\zbound}{16} \beta_k^{\frac{1}{k}} c_k \delta^{\frac{1}{k_*}}
  =  \frac{\zbound}{16} \beta_k^{\frac{1}{k}} c_k \left(\frac{1}{4n} 
    \wedge  p_k 
    \wedge (1 - (1-\beta_k)^{1-k_*} p_k )\right)^{\frac{1}{k_*}}.
\end{equation*}

\subsection{Proof of Proposition~\ref{proposition:lower-general-optimization-sqrt-new}}
\label{section:proof-of-lower-general-optimization-sqrt-new}

\newcommand{\rem}{r}

For $p$ given by hypothesis, recall the definition~\eqref{eqn:q-p-def} of
$q(p)$.  Following the same logic as in the proof of
Proposition~\ref{proposition:lower-general-estimation-sqrt-new}, the implicit
function theorem implies that $q(\cdot)$ is continuously
differentiable near $p$ with
\begin{equation*}
  q'(p) = \frac{-\partial_p h_f(q(p); p)}{\partial_q h_f(q(p); p)} > 0,
\end{equation*}
where $ h_f(q; p) = p f(\frac{q}{p}) + (1-p) f( \frac{1-q}{1-p} )$ as
before. From Taylor's theorem, we then have
\begin{equation*}
  q(p+ \delta) = q(p) + q'(p) \delta + \rem(\delta)
\end{equation*}
for a remainder $\rem(\delta) = o(\delta)$ as $\delta \to 0$. For
small $\delta > 0$, define
\begin{equation*}
  \epsilon_{\delta} \defeq
  \left( q(p) + \half\left( q'(p) \delta + \rem(\delta)\right) \right)^{-1}-1
  > 0.
\end{equation*}

We use the reduction from robust optimization to testing of
Lemma~\ref{lemma:reduction-robust}. For
some $\delta \in (0, q(p) - p)$ to be choosen later, consider the two
distributions $X_1 \sim P_1$, $X_2 \sim P_2$ with
\begin{align*}
  X_1 = 
  \begin{cases}
    -1 & \mbox{w.p.~} 1-p \\
    \epsilon_{\delta}
    & \mbox{w.p.~} p,
  \end{cases}
  ~~~~~~
  X_2 = 
  \begin{cases}
    -1 & \mbox{w.p.~} 1-p - \delta \\
    \epsilon_{\delta} & \mbox{w.p.~} p + \delta.
  \end{cases}
\end{align*}
For $\loss(\theta; X) = \theta X$, we show that
$\theta \mapsto \risk_f(\theta; P_1)$ and
$\theta \mapsto \risk_f(\theta; P_2)$ are well-separated in the distance
$\mbox{d}_{\rm opt}(\cdot, \cdot)$, but $P_1$ and $P_2$ are close in total
variation distance. By definition
\begin{align*}
  \risk_f(\theta; P_1) = 
  \begin{cases}
    -\theta (1-(1+\epsilon_{\delta})q(p)) & \mbox{if}~~ \theta \ge 0 \\
    -\theta (-\epsilon_{\delta}+ (1+\epsilon_{\delta})q(1-p)) & \mbox{otherwise} ,
  \end{cases}
\end{align*}
and similarly,
\begin{align*}
  \risk_f(\theta; P_2) = 
  \begin{cases}
    -\theta (1-(1+\epsilon_{\delta})q(p+\delta)) & \mbox{if}~~ \theta \ge 0 \\
    -\theta (-\epsilon_{\delta}+ (1+\epsilon_{\delta})q(1-p-\delta)) & \mbox{otherwise.} 
  \end{cases}
\end{align*}
By our choice of $\epsilon_{\delta}$, observe
\begin{equation*}
  1-(1+\epsilon_{\delta})q(p) > 0, ~~~~~~\mbox{but}~~~~~~
  1-(1+\epsilon_{\delta})q(p+\delta) \le 0,
\end{equation*}
and $q(p) > p$ so that $q(p) > p + \delta$ for small
$\delta$, and similarly $q(1 - p - \delta) + \delta > 1 - p$. Consequently,
$1 + \epsilon_{\delta} < \frac{1}{q(p)} < \frac{1}{p+\delta} <
\frac{1}{1-q(1-p-\delta)}$, and so
\begin{equation*}
  -\epsilon_{\delta}+ (1+\epsilon_{\delta})q(1-p)
  \ge -\epsilon_{\delta}+ (1+\epsilon_{\delta})q(1-p-\delta)
  >  0.
\end{equation*}
Thus, we have $\risk_f'(\theta; P_1) < 0$ for all $\theta$,
while $\risk_f'(\theta; P_2) > 0$ for $\theta > 0$ and
$\risk_f'(\theta; P_2) < 0$ for $\theta < 0$.
We conclude that
$\theta_i\opt \defeq \argmin_{\theta \in [-\zbound, \zbound]} \risk_f(\theta;
P_i)$ satisfies $\theta_1\opt = \zbound$ and $\theta_2\opt = 0$.

We now show $\mbox{d}_{\rm opt}(P_1, P_2) \ge \zbound \Delta_{\delta}$, where
\begin{equation*}
  \Delta_{\delta} \defeq
  \frac{q'(p) \delta + \rem(\delta)}{4 \left(q(p) + \half(q'(p) \delta +
      \rem(\delta)\right)}
  = \frac{1}{4} (1+\epsilon_{\delta}) \left( q'(p) \delta + \rem(\delta)\right).
\end{equation*}
In the sequel, we use the following identities to simplify computation:
\begin{equation*}
  2 \Delta_{\delta} = 1 - (1+\epsilon_{\delta})q(p),~~~\mbox{and}~~~
  - 2 \Delta_{\delta} = 1 - (1+\epsilon_{\delta}) q(p+\delta).
\end{equation*}
First, for any $\kappa \in [0, \Delta_{\delta}]$, consider $\theta$ such that
\begin{equation*}
  \risk_f(\theta; P_1) \le \risk_f(\theta_1\opt; P_1) + \zbound \kappa.
\end{equation*}
Assume for contradiction that $\theta < 0$: the above bound implies
\begin{align*}
  \theta & \ge \frac{\zbound }{-\epsilon_{\delta}+ (1+\epsilon_{\delta})q(1-p)}
  \left(1 - (1+\epsilon_{\delta})q(p) - \kappa\right) \\
  & = \frac{\zbound }{-\epsilon_{\delta}+ (1+\epsilon_{\delta})q(1-p)}
  \left( 2\Delta_{\delta} - \kappa\right) \ge 0.
\end{align*}
For $\theta \ge 0$, the optimality bound implies
\begin{equation*}
  \theta \ge \zbound \left(1 - \frac{\kappa}{1-(1+\epsilon_{\delta})q(p)}\right)
  = \zbound \left(1- \frac{\kappa}{2\Delta_{\delta}}\right).
\end{equation*}
Using this bound, we obtain
\begin{align*}
  \risk_f(\theta; P_2) - \risk_f(\theta_2\opt; P_2)
  & = \risk_f(\theta; P_2) = - (1-(1+\epsilon_{\delta})q(p+\delta))\theta
  =  2\zbound \Delta_{\delta}
  \left( 1 - \frac{\kappa}{2 \Delta_{\delta}}\right) \\
  & \ge \zbound \left( 2\Delta_{\delta} - \kappa\right)
    \ge \zbound \kappa.
\end{align*}

Next, for any $\kappa \in [0, \Delta_{\delta}]$, consider $\theta$ such that
\begin{equation*}
  \risk_f(\theta; P_2) \le \risk_f(\theta_2\opt; P_2) + \zbound \kappa
  = \zbound\kappa,
\end{equation*}
which implies
$\theta \le - \frac{\zbound\kappa}{1-(1+\epsilon_{\delta})q(p+\delta)}$ if
$\theta \ge 0$, and
$\theta \ge \frac{\zbound \kappa}{-\epsilon_{\delta}+
  (1+\epsilon_{\delta})q(1-p-\delta)}$ if $\theta < 0$. When $\theta \ge 0$,
we then obtain
\begin{align*}
  \risk_f(\theta; P_1) - \risk_f(\theta_1\opt; P_1)
  = (1-(1+\epsilon_{\delta})q(p)) (\zbound - \theta)
  \ge 2\zbound \Delta_{\delta} \left( 1+ \frac{\kappa}{2\Delta_{\delta}}
  \right)
  \ge \zbound \kappa.
\end{align*}
When $\theta < 0$, we get
\begin{align*}
  \risk_f(\theta; P_1) - \risk_f(\theta_1\opt; P_1)
  \ge \zbound \kappa \left(
  \frac{-\epsilon_{\delta}+
  (1+\epsilon_{\delta})q(1-p)}{-\epsilon_{\delta}+
  (1+\epsilon_{\delta})q(1-p-\delta)} + 2
  \right) \ge \zbound \kappa.
\end{align*}
We thus conclude that
$\mbox{d}_{\rm opt}(P_1, P_2) \ge \zbound \Delta_{\delta}$ as claimed.

We now pick $\delta$ such that $\tvnorms{P_1^n - P_2^n} \le \half$.  By
Pinsker's inequality, we have
$\tvnorms{P_1^n - P_2^n}^2 \le \frac{n}{2} \dkl{P_1}{P_2}$, and letting
$\delta_n = \sqrt{\frac{p(1-p)}{n}}$, we get
$\dkl{P_1}{P_2} \le \frac{1}{n}$ as for $\delta \in [0, p]$,
we have as usual that $\dkl{P_1}{P_2} \le \frac{\delta^2}{p(1 - p)}$.
From the reduction from distributionally robust optimization to hypothesis
testing (Lemma~\ref{lemma:reduction-robust}), we conclude
\begin{align*}
  \minimaxopt
  \ge  \frac{\zbound}{4} \Delta_{\delta_n}.
\end{align*}
Multiplying both sides by $\sqrt{n}$ and taking $n \to\infty$, we obtain the
result.

\subsection{Proof of
  Proposition~\ref{proposition:lower-general-optimization-kconj}}
\label{section:proof-of-lower-general-optimization-kconj}

We proceed as in the second part of
Section~\ref{section:proof-of-lower-optimization}. We use
Lemma~\ref{lemma:reduction-robust} on the
distributions $X_1 \sim P_1$, $X_2 \sim P_2$
\begin{align*}
  X_1 \equiv -\epsilon,
  ~~~~~~
  X_2 = 
  \begin{cases}
    -\epsilon & \mbox{w.p.~} 1-\delta \\
    1  & \mbox{w.p.~} \delta
  \end{cases}
\end{align*}
where
$\epsilon \defeq \left( \frac{\tol}{2m} \right)^{\frac{1}{k}}
\delta^{\frac{1}{k_*}}$ for some 
\begin{equation*}
  0 < \delta \le \frac{1}{2C_{f, \tol, m}} \wedge
  \frac{\tol}{2m} \left(
    \left(\frac{2}{3}\right)^k \wedge
    \half \left( \frac{\tol}{2m} \right)^{-k_*}
    \right)
\end{equation*}
to be choosen later.  Now, we again show that
$\distopt{P_1}{P_2}{f} \ge \frac{\epsilon}{2}$. To this end, first observe that
$\risk_f(\param; P_1) = - \epsilon \param$. When $\param < 0$, we have
$\risk_f(\param; P_2) \ge \theta \E[X_2] \ge - \param (\epsilon (1-\delta) -
\delta) \ge 0$ as $\epsilon \le \half$ in the given range of $\delta$. When
$\param \ge 0$, recall that $Q$ such that
$Q(Z = \zbound) = \left(\frac{\tol}{2m}\right)^{\frac{1}{k}}
\delta^{\frac{1}{k_*}}$ and $Q(Z = 0) = 1-Q(Z = \zbound)$, satisfies
$\fdiv{Q}{P_2} \le \tol$ by Lemma~\ref{lemma:q-included}. Hence, we have for
$\theta \ge 0$
\begin{equation*}
  \risk_f(\param; P_2) \ge \epsilon \theta.
\end{equation*}
Denoting $\param_i\opt \defeq \argmin_{\param \in \Theta} \risk_f(\param; P_i)$
again, we consequently obtain $\param_1\opt = \zbound$, $\param_2\opt = 0$ with
$\risk_f(\param_1\opt; P_1) = -\zbound \epsilon$, $\risk_f(\param_1\opt; P_2) = 0$.

Using an identical argument as in the second part of
Section~\ref{section:proof-of-lower-optimization}, we can show
$\distopt{P_1}{P_2}{f} \ge \frac{\zbound \epsilon}{2}$. Setting
\begin{equation*}
  \delta = \frac{1}{2 (n \vee C_{f, \tol, m})} 
  \wedge   \frac{\tol}{2m} \left(
    \left(\frac{2}{3}\right)^k \wedge
    \half \left( \frac{\tol}{2m} \right)^{-k_*}
    \right)
\end{equation*}
and using the same argument as in
Section~\ref{section:proof-of-lower-optimization}, we obtain the result.


%% file: consistency-proof.tex
\section{Proofs of Consistency}
\label{section:proof-of-consistency}

We begin this section with a brief review of the theory of
epi-convergence~\cite{KingWe91, RockafellarWe98}, which governs convergence
of solutions to optimization problems, so we consequently use its tools to
develop our consistency results.

We begin with some necessary set-valued analysis.
\begin{definition}
  \label{definition:set-convergence}
  Let $\{A_n\}$ be a sequence of subsets of $\R^d$. The \emph{limit supremum}
  (or \emph{limit exterior} or \emph{outer limit}) and \emph{limit infimum}
  (\emph{limit interior} or \emph{inner limit}) of the sequence $\{A_n\}$
  are
  \begin{align*}
    \limsup_n A_n & \defeq \left\{v \in \R^d
    \mid \liminf_{n \to \infty} \dist(v, A_n) = 0 \right\} ~~~ \mbox{and} \\
    \liminf_n A_n & \defeq \left\{v \in \R^d
    \mid \limsup_{n \to \infty} \dist(v, A_n) = 0 \right\}.
  \end{align*}
\end{definition}
\noindent
Recall that the epigraph of a function $h : \R^d \to \R \cup \{+\infty\} $ is
\begin{equation*}
  \epi h \defeq \{(x, t) \in \R^d \times \R \mid h(x) \le t \}.
\end{equation*}
Based on Definition~\ref{definition:set-convergence}
of limits of sets, we say that $\lim_n A = A_\infty$ if
$\limsup_n A_n = \liminf_n A_n = A_\infty \subset \R^d$, and
we have the following notion of convergence of functions in terms of their
epigraphs.
\begin{definition}
  \label{def:epi-conv}
  A sequence of functions \emph{$h_n$ epi-converges to a function $h$},
  denoted $h_n \cepi h$, if
  \begin{equation}
    \label{eqn:epi-convergence}
    \epi h = \liminf_{n \to \infty} \epi h_n = \limsup_{n \to \infty}
    \epi h_n.
  \end{equation}
\end{definition}
\noindent
If $\dom h \neq \varnothing$, meaning that $h$ is proper, epigraphical
convergence~\eqref{eqn:epi-convergence} for closed convex
functions has the following equivalent characterizations.
\begin{lemma}[Theorem~7.17,~\citet{RockafellarWe98}]
  \label{lemma:epi}
  Let $h_n : \R^d \to \wb{\R}, h : \R^d \to \wb{\R}$ be closed convex and
  proper. Then $h_n \cepi h$ is equivalent to either of the following
  two conditions.
  \begin{enumerate}[(i)]
  \item There exists a dense set $A \subset \R^d$ such that
    $h_n(v) \to h(v)$ for all $v \in A$.
  \item For all compact $C \subset \dom h$ not containing a boundary point
    of $\dom h$,
    \begin{equation*}
      \lim_{n \to \infty} \sup_{v \in C} |h_n(v) - h(v)| = 0.
    \end{equation*}
  \end{enumerate}
\end{lemma}
\noindent
Importantly for our development,
epigraphical convergence implies the infimal value convergence, and
under additional conditions, convergence of solution sets.
\begin{lemma}[Theorem 7.31, \citet{RockafellarWe98}]
  \label{lemma:epi-solution-consistency}
  Let $h_n : \R^d \to \wb{\R}, h : \R^d \to \wb{\R}$ satisfy $h_n \cepi h$
  and $-\infty < \inf h < \infty$. Let $S_n(\varepsilon) = \{\theta \mid
  h_n(\theta) \le \inf h_n + \varepsilon\}$ and $S(\varepsilon) = \{\theta
  \mid h(\theta) \le \inf h + \varepsilon\}$.  Then $\limsup_n
  S_n(\varepsilon) \subset S(\varepsilon)$ for all $\varepsilon \ge 0$, and
  $\limsup_n S_n(\varepsilon_n) \subset S(0)$ whenever $\varepsilon_n
  \downarrow 0$.
\end{lemma}
\begin{lemma}[Proposition~7.33,~\citet{RockafellarWe98}]
  \label{lemma:epi-consistency}
  Let $h_n : \R^d \to \wb{\R}, h : \R^d \to \wb{\R}$ be closed and
  proper. If $h_n$ has bounded sublevel sets and $h_n \cepi h$, then
  $\inf_{v} h_n(v) \to \inf_{v} h(v)$.
\end{lemma}

\subsection{Proof of Proposition~\ref{proposition:consistency-estimation}}
\label{section:proof-of-consistency-estimation}

To ease notation, we fix $\param \in \Theta$ and denote $Z(x) \defeq
\loss(\param; x)$, and we typically omit the dependence of $\risk$ on
$\theta$ (as it is fixed), writing $\risk_f(P)$ and $\risk_k(P)$.
The proof builds out of the epi-convergence theory we outline
in the beginning of Section~\ref{section:proof-of-consistency}.

By Proposition~\ref{proposition:duality}, strong duality~\eqref{eqn:dual}
holds for both $P = P_0$ and $P = \emp$.  For a probability measure $P$,
define the dual objective
\begin{align*}
  g_{f,P}(\lambda, \eta)
  & \defeq
  \begin{cases}
    \E_{P}\left[\lambda f^*\left(\frac{Z - \eta}{\lambda}\right)\right]
    + \tol
    \lambda + \eta & ~~\mbox{if}~~\lambda \ge 0 \\
    \infty & ~~\mbox{otherwise},
  \end{cases}
\end{align*}
where by convention we use the \emph{closure} of the perspective $(\lambda,
t) \mapsto \lambda f^*(t / \lambda)$ (cf.~\cite[Sec.~3.2]{Shapiro17}
and~\cite[Prop.~IV.2.2.2]{HiriartUrrutyLe93ab}).
Using that $f^*(s) \ge 0$ for $s \ge 0$ and our assumption
that $\E[f^*(|Z|)] < \infty$, the strong law of large numbers implies
that
\begin{equation*}
  \event \defeq \left\{
  \lim_{n \to \infty} g_{f,\emp}(\lambda, \eta) =
  g_{f,P_0}(\lambda, \eta)~~\mbox{for all}~~ \lambda \in \Q, \eta \in \Q \right\}
\end{equation*}
has $P_0$-measure $1$.  We now show that the functions $g_f$ are both
closed.  To that end, note that standard conjugacy
calculations~\cite[Prop.~I.6.1.2]{HiriartUrrutyLe93ab} imply $1 \in \partial
f^*(0) = \argmax_t \{-f(t)\}$, as $f(1) = 0$, $t = 1$ minimizes $f$,
and $f^*(0) = 0$. Thus we have
$f^*(s) \ge f^*(0) + s$ for all $s$,
so that
\begin{equation*}
  \lambda f^*\left(\frac{z - \eta}{\lambda}\right) - (z - \eta) \ge 0.
\end{equation*}
Fatou's lemma then implies that
for $v = (\eta,\lambda)$ and $v_0 = (\eta_0, \lambda_0)$ we have
\begin{align*}
  & \liminf_{v \to v_0} \left\{ \E_{P}\left[\lambda f^*\left(\frac{Z -
      \eta}{\lambda}\right)
    - (Z - \eta) \right] + \tol \lambda + \eta \right\} \\
  & \ge \E_P\left[
    \liminf_{v \to v_0}
    \left\{\lambda f^*\left(\frac{Z - \eta}{\lambda}\right)
    - (Z - \eta) \right\} \right]
  + \tol \lambda_0 + \eta_0 \\
  & \ge \E_P\left[
    \lambda_0  f^*\left(\frac{Z - \eta_0}{\lambda_0}\right)
    - (Z - \eta_0) \right]
  + \tol \lambda_0 + \eta_0,
\end{align*}
where the last inequality follows by the lower semicontinuity of the
closure of the perspective.
Using Lebesgue's dominated convergence theorem on $(Z - \eta)$, using
the dominating function $|Z| + |\eta|$, we have thus
shown that both
$g_{f,\emp}$ and $g_{f,P_0}$ are lower semicontinuous.
Lemma~\ref{lemma:epi} implies that $g_{f,\emp} \cepi g_{f,P_0}$ with
probability 1.

Finally, we would like to apply
Lemma~\ref{lemma:epi-consistency}; to do so, we must show that
$g_{f,\emp}$ is (eventually) coercive. For this, we note that
$\lambda f^*(\frac{Z - \eta}{\lambda}) - Z + \eta \ge 0$ as above,
so that $g_{f,P}(\eta, \lambda)
\ge \tol \lambda + \E_{P}[Z]$, and thus
for any $P$ for which $\E_P[Z]$ exists,
$\lim_{\lambda \to \infty}
\inf_\eta g_{f,P}(\eta, \lambda) = \infty$.
To show coercivity of $g_{f,P}$ as $\norm{(\eta, \lambda)} \to \infty$,
we thus need only consider limits taken as $\lambda$ remains bounded.
Now, we claim that under the conditions of the lemma,
\begin{equation}
  \label{eqn:liminf-f-conjugate}
  \limsup_{s \to -\infty} \frac{f^*(s)}{s} = \epsilon < 1
  ~~ \mbox{and} ~~
  \liminf_{s \to \infty} \frac{f^*(s)}{s} = \infty.
\end{equation}
Deferring the proof of the claims~\eqref{eqn:liminf-f-conjugate},
let us show how they imply that $g_{f,P_0}$ is coercive.
Assume that $0 \le \lambda \le \Lambda < \infty$.
For any constant $K < \infty$, $K > \Lambda$, there exist
$b, c < \infty$ such that $|z| \le b$ and $\eta < -c$ imply that
$f^*(\frac{z - \eta}{\lambda})
\ge K |\eta| / \Lambda$, and similarly,
$\eta > c$ implies
$\lambda f^*(\frac{z - \eta}{\lambda}) \ge -\frac{1 + \epsilon}{2} \eta$.
For $\eta < -c$, then, we have
\begin{equation*}
  g_{f,P}(\eta, \lambda)
  \ge P(|Z| \le b)
  \left[\frac{K |\eta|}{\Lambda} + \tol \lambda + \eta\right]
  + P(|Z| > b) \tol \lambda + \E_P[\indic{|Z| > b} Z],
\end{equation*}
and for $\eta > c$ we similarly have
\begin{equation*}
  g_{f,P}(\eta, \lambda)
  \ge P(|Z| \le b)
  \left[\tol \lambda + \frac{\epsilon \eta}{2} \right]
  + P(|Z| > b) \tol \lambda + \E_P[\indic{|Z| > b} Z].
\end{equation*}
Whenever $\E_P[|Z|] < \infty$, we see that
$\lim_{|\eta| \to \infty} \inf_{\lambda \in [0, \Lambda]}
g_{f,P}(\eta, \lambda) = \infty$, so that $g_{f,P}$ is coercive.
Consequently, the claim~\eqref{eqn:liminf-f-conjugate}, coupled
with our assumption that $\E_{P_0}[|Z|] < \infty$, implies that
$g_{f,P_0}$ is coercive. Because $g_{f,\emp} \cepi g_{f,P_0}$, we have
uniform convergence of $g_{f,\emp}$ to $g_{f,P_0}$ on compacta
(Lemma~\ref{lemma:epi}), and thus $g_{f,\emp}$ is eventually coercive.
Lemma~\ref{lemma:epi-consistency} thus implies the result.

Finally, we return to the claim~\eqref{eqn:liminf-f-conjugate}.
For the first claim, we have for $s < 0$ that
\begin{align*}
  \frac{1}{s} \sup_{t \ge 0} \{ st - f(t) \}
  & = \inf_{t \ge 0} \left\{t + \frac{f(t)}{|s|}\right\},
\end{align*}
which is decreasing as $s \downarrow -\infty$, and letting $t_0
< 1$ be any value for which $f(t_0) < \infty$ (as
$f$ is finite near $t = 1$), we have
$\limsup_{s \to -\infty} \frac{1}{s} f^*(s) \le t_0 < 1$ as desired.
For the second claim of inequalities~\eqref{eqn:liminf-f-conjugate},
use that $f(t) < \infty$ for all $t \ge 1$; for each $n \in \N$, then,
there exists $s < \infty$ such that $f(n) / s \le 2$, so that
$\frac{1}{s} f^*(s) = \sup_{t \ge 0} \{t - f(t) / s\}
\ge n - 2$. Taking $n \to \infty$ gives the claim.


\subsection{Proof of Proposition~\ref{proposition:consistency-soln}}
\label{section:proof-of-consistency-soln}

The epi-convergence theory of the beginning of
Section~\ref{section:proof-of-consistency}, combined
with Proposition~\ref{proposition:consistency-estimation}, gives most of the
results.  First, we know that $\risk_f(\theta; \emp)$ and $\risk_f(\theta;
P_0)$ are lower semicontinuous in $\theta$, as each is the supremum of
closed convex functions $\theta \mapsto \int \loss(\theta; \statval)
dP(\statval)$. Combined with
Proposition~\ref{proposition:consistency-estimation}, we have
that $\risk_f(\cdot; \emp) \cepi \risk_f(\cdot; P_0)$ with $P_0$-probability
1. Using the coercivity
of $\risk_f(\cdot; P_0)$ and that $\risk_f(\theta; P_0) < \infty$
on an open set containing
$S_{P_0}(\Theta,0)$, we take any compact set $C \subset \R^d$ containing
$S_{P_0}(\Theta,0)$ with $\risk_f(\theta; P_0) < \infty$ on $C$, and we obtain
$\sup_{\theta \in C} |\risk_f(\theta; P_0) - \risk_f(\theta; \emp)| \cas 0$
by Lemma~\ref{lemma:epi}.
The convexity of $\risk_f(\cdot; \emp)$ then implies that
$\risk_f(\cdot; \emp)$ is coercive eventually, so that it has bounded
sublevel sets, and
Lemma~\ref{lemma:epi-consistency} implies that
$\inf_{\theta \in \Theta} \risk_f(\theta; \emp) \cas
\inf_{\theta \in \Theta} \risk_f(\theta; P_0)$.

For the second result, we use that for any sequence $\varepsilon_n \ge 0$,
eventually the set $S_{\emp}(\Theta, \varepsilon_n)$ is non-empty by
coercivity, and then Lemma~\ref{lemma:epi-solution-consistency} implies
that
\begin{equation*}
  \limsup_n S_{\emp}(\Theta, \varepsilon_n) \subset S_{P_0}(\Theta, 0).
\end{equation*}
In turn, this yields that $\lim_n \dinclude(S_{\emp}(\Theta, \varepsilon_n)) =
0$ as $S_{P_0}(\Theta, 0)$ is compact by the coercivity assumption.

%% file: asymptotics-proof-lam.tex
\section{Proof of Limit Theorems}
\label{section:proof-of-limit-theorems}

\subsection{Proof of Lemma~\ref{lemma:uniqueness-eta-per-param}}
\label{sec:proof-uniqueness-eta-per-param}

To ease notation, let $Z = \loss(\theta_0; \statrv)$, and
recall from Lemma~\ref{lemma:cressie-read-risk}
(and its proof in Section~\ref{section:proof-of-cressie-read-risk})
that we may rewrite the dual as
\begin{equation*}
  g_P(\theta, \lambda, \eta)
  = \frac{1}{\lambda^{k_* - 1}} \frac{(k - 1)^{k_*}}{k}
  \E_{P}\left[\hinge{Z - \eta}^{k_*}\right]
  + \left(\tol + \frac{1}{k(k - 1)}\right) \lambda
  + \eta.
\end{equation*}
In this case, it is clear that the minimizing $\lambda$
is unique as in Eq.~\eqref{eqn:lambda-min-cressie}, with
\begin{equation*}
  g_P(\eta) \defeq \inf_{\lambda \ge 0}
  g_P(\theta, \lambda, \eta)
  = c_k \E_P\left[\hinge{Z - \eta}^{k_*}\right]^{1/k_*} + \eta,
\end{equation*}
where $c_k = (k(k - 1) \tol + 1)^{1/k} > 1$.
It is evident that $g_P$ is convex and coercive in $\eta$.
Now, for all $\eta \ge \esssup Z$ we have $g_P(\eta) = \eta$,
so that $g_P$ is strictly increasing in $\eta \ge \esssup Z$. On
the set $(-\infty, \esssup Z)$, we claim that $g_P$ is strictly convex in
$\eta$.
Indeed, for $\eta_1 \neq \eta_2 \in (-\infty, \esssup Z)$ and $\alpha
\in (0, 1)$, we
have
\begin{align*}
  \lefteqn{g_P(\alpha \eta_1 + (1-\alpha) \eta_2)} \\
  & \stackrel{(i)}{\le}
  c_k \norm{\alpha \hinge{Z - \eta_1} + (1-\alpha) \hinge{Z - \eta_2}}_{k_*, P}
  + \alpha \eta_1 + (1-\alpha)\eta_2 \\
  & \stackrel{(ii)}{<} c_k \alpha \norm{\hinge{Z - \eta_1}}_{k_*, P}
  + c_k (1-\alpha) \norm{\hinge{Z - \eta_2}}_{k_*, P}
  + \alpha \eta_1 + (1-\alpha)\eta_2 \\
  & = \alpha g_P(\eta_1) + (1 - \alpha) g_P(\eta_2),
\end{align*}
where step $(i)$ follows by convexity and that the
norm $\norm{\cdot}$ is increasing in positive arguments,
while inequality $(ii)$ follows because equality in Minkowski's
inequality $\norm{Y_1 + Y_2}_{k_*} \le \norm{Y_1}_{k_*} + \norm{Y_2}_{k_*}$
for $k_* \in (1, \infty)$ holds
if and only if there exists $c \in \R_+$ such that
$Y_1 = c Y_2$ with probability one.



\subsection{Proof of Theorem~\ref{theorem:asymptotics-param}}
\label{section:proof-of-asymptotics-param}

We use a powerful result on asymptotic normality
that we show applies in our setting. To state the result, we require
a bit of (temporary) notation.
First, recall the definition of bracketing numbers for a collection
of functions.
\begin{definition}
  Let $\norm{\cdot}$ be a (semi-)norm on $\mc{H}$.  For functions
  $l, u : \statdomain \to \R$ with $l \le u$, the \emph{bracket} $[l, u]$ is
  the set of functions $h : \statdomain \to \R$ such that $l \le h \le u$, and
  $[l, u]$ is an \emph{$\epsilon$-bracket} if $\norm{l - u} \le \epsilon$.
  Brackets $\{[l_i, u_i]\}_{i = 1}^m$ \emph{cover} $\hclass$ if for all
  $h \in \mc{H}$, there is some bracket $i$ such that $h \in [l_i, u_i]$. The
  \emph{bracketing number}
  $N_{[\hspace{1pt}]}(\epsilon, \mc{H}, \norm{\cdot})$ is the minimum number
  of $\epsilon$-brackets needed to cover $\hclass$.
\end{definition}

Now, let $\mc{V} \subset \R^d$ be a convex set and $H : \mc{V} \times
\statdomain \to \R$ be a collection of criterion functions, where
$\what{v}_n = \argmin_{v \in \mc{V}} \E_{\emp}[H(v; \statrv)]$.  Assume that
$v\opt = \argmin_{v \in \mc{V}} \E_{P_0}[H(v; \statrv)]$ exists and is
unique, and for $\epsilon > 0$, define the localized function
classes
\begin{equation*}
  \mc{H}_\epsilon \defeq \left\{\statval \mapsto H(v; \statval)
  - H(v\opt; \statval)
  : \norm{v - v\opt} \le \epsilon \right\}.
\end{equation*}
We say that $M_\epsilon : \statdomain \to \R_+$ is an envelope
for $\mc{H}_\epsilon$ if
$h \in \mc{H}_\epsilon$ implies
$|h(\statval)| \le M_\epsilon(\statval)$; without further mention
we take $M_\epsilon(\statval) \defeq \sup_{\norm{v - v\opt} \le \epsilon}
|H(v;\statval) - H(v\opt;\statval)|$.
With these definitions, we have the following result.
\begin{lemma}[{\cite[Theorem 3.2.10]{VanDerVaartWe96}}]
  \label{lemma:m-estimator}
  Let the conditions above hold, and assume that $\mc{H}_\epsilon$ has
  envelope $M_\epsilon$ with $\E[M_\epsilon^2] < \infty$. Assume
  additionally that
  \begin{enumerate}[(i)]
  \item The function $v \mapsto R(v) \defeq \E[H(v; \statrv)]$ is
    $\mc{C}^2$ near $v\opt$ and $\nabla^2 R(v\opt) \succ 0$.
  \item The bracketing integral of $\mc{H}_{\epsilon}$ is uniformly bounded as
    $\epsilon \to 0$: for some $\epsilon_0 > 0$,
    \begin{equation}
      \label{eqn:bracketing-cond}
      \int_0^{\infty} \sup_{\epsilon < \epsilon_0} \sqrt{
        \log N_{[\hspace{1pt}]} \left(
          \delta \norm{M_{\epsilon}}_{P_0, 2},
          \mc{H}_{\epsilon}, L_2(P_0)
        \right) } d\delta
      < \infty.
    \end{equation}
  \item There exists $C < \infty$ such that
    $\E[M_\epsilon(X)^2] \le C \epsilon^2$ for
    all small $\epsilon$.
  \item There exists a centered Gaussian process $G$ on $\R^d$ where
    $G(v) = G(v')$ $P_0$-almost surely only if $v = v'$ such that for
    every $c, K > 0$,
    \begin{subequations}
      \label{eqn:limits-check}
      \begin{align}
        & \lim_{\epsilon \to 0} \epsilon^{-2}
          \E[M_{\epsilon}(X)^2 \indic{M_\epsilon(X) > c}] = 0,
          \label{eqn:weak-2norm} \\
        & \lim_{\epsilon \to 0} \limsup_{\delta \to 0}
          \sup_{\norm{u_1 - u_2} < \epsilon, \norm{u_1} \vee \norm{u_2} \le K}
          \delta^{-2}
          \E[ \left(H(v\opt + \delta u_1; \statrv)
          - H(v\opt + \delta u_2; \statrv)\right)^2] = 0
          \label{eqn:modulus-zero} \\
        & \lim_{\delta \to 0}
          \delta^{-2}
          \E[ \left(H(v\opt + \delta u_1; \statrv)
          - H(v\opt + \delta u_2; \statrv)\right)^2]
          = \E[(G(u_1) - G(u_2))^2].
          \label{eqn:covariance}
      \end{align}
    \end{subequations}
  \end{enumerate}
  Then, there exists a version of $G$ with bounded, uniformly continuous
  sample paths on compacta.  Further, if $\what{v}_n \in \mc{V}$ satisfies
  $\E_{\emp}[H(\what{v}_n; \statrv)] \le \inf_{v \in \mc{V}} \E_{\emp}[H(v;
    \statrv)] + O_P(1/n)$ and $\what{v}_n \cas v\opt$, then
  $\sqrt{n}(\what{v}_n - v\opt)$ converges in distribution to the unique
  maximizer of the process
  \begin{equation*}
    u \mapsto G(u) + \half u^T \nabla^2 R(v\opt) u.
  \end{equation*}
\end{lemma}

We now show how under the conditions specified in
Theorem~\ref{theorem:asymptotics-param}, our problem satisfies the
conditions of Lemma~\ref{lemma:m-estimator}.
We first provide notation and a few additional definitions for shorthand.
Define
\begin{equation*}
  H(\theta, \lambda, \eta; \statrv)
  \defeq \lambda f^*\left(\frac{\loss(\theta; \statrv) - \eta}{\lambda}
  \right) + \tol \lambda + \eta,
\end{equation*}
so that $g_P(\theta, \lambda, \eta) = \E_{P}[H(\theta, \lambda, \eta;
  \statrv)]$. Let $(\what{\theta}_n, \what{\lambda}_n, \what{\eta}_n)$ be
the empirical minimizer
\begin{align*}
  (\what{\theta}_n, \what{\lambda}_n, \what{\eta}_n)
  \in
  \argmin_{\theta, \lambda \ge 0, \eta} \E_{\emp}[  H(\theta, \lambda, \eta; X) ].
\end{align*}
For $\epsilon > 0$, define the collection
\begin{equation}
  \label{eqn:our-localized-set}
  \mc{H}_\epsilon \defeq
  \left\{ x \mapsto H(\theta, \lambda, \eta; \statval) - H(\theta\opt,
  \lambda\opt, \eta\opt; \statval) :
    \norm{\theta - \theta\opt} + |\lambda - \lambda\opt|
    + |\eta-\eta\opt| \le \epsilon
  \right\}.
\end{equation}

We claim that the envelope $M_\epsilon$ exists for the
set~\eqref{eqn:our-localized-set}. First, we note that $\nabla H$ exists
with probability 1: by our
Assumption~\ref{assumption:strong-identifiability} that $g_{P_0}$ is
$\mc{C}^2$ near $(\theta\opt, \lambda\opt, \eta\opt)$, we know that
$g_{P_0}$ is continuously differentiable. Then For $h(t, \statval)$ an
arbitrary function, convex in $t$, $\int h(t, \statval) dP(\statval)$ is
differentiable at some $t_0$ if and only if $t \mapsto h(t, \statval)$ is
differentiable at $t_0$ for $P$-almost all $\statval$~\cite{Bertsekas73}.
Consequently, for $P_0$-almost all $\statval$ we have $\nabla H(\cdot;
\statval)$ exists in a neighborhood of $(\theta\opt, \lambda\opt,
\eta\opt)$, and
\begin{equation}
  \label{eqn:H-derivative}
  \nabla H(\theta, \lambda, \eta; \statval)
  = \left[
    \begin{matrix}
      {f^*}'\left(\frac{\loss(\theta; \statval) - \eta}{\lambda}\right)
      \nabla \loss(\theta; \statval) \\
      -{f^*}'\left(\frac{\loss(\theta; \statval) - \eta}{\lambda}\right)
      + 1 \\
      f^*\left(\frac{\loss(\theta; \statval) - \eta}{\lambda}\right)
      - \frac{1}{\lambda} {f^*}'\left(\frac{\loss(\theta; \statval) - \eta}{
        \lambda}\right) (\loss(\theta; \statval) - \eta) + \tol
    \end{matrix}
    \right]
\end{equation}
for $(\theta, \lambda, \eta)$ near $(\theta\opt, \lambda\opt, \eta\opt)$.
We begin with a simple technical lemma.
\begin{lemma}
  \label{lemma:f-dual-growth}
  Let $f$ satisfy the conditions of Theorem~\ref{theorem:asymptotics-param}
  and $k_* = \frac{k}{k-1}$.
  Then $\limsup_{s \to \infty} f^*(s) / s^{k_*} < \infty$, and
  for any $t(s) \in \partial f^*(s)$, $t(s) \ge 0$ and
  $\limsup_{s \to \infty} t(s) / s^\frac{1}{k-1} < \infty$.
\end{lemma}
\begin{proof}
  We begin with the first claim, recalling the assumption that
  $\liminf_{t \to \infty} f(t) / t^k  > 0$, so that
  for some $t_0 < \infty$ there exists $c > 0$ such that
  $f(t) \ge c t^k$ for all $t \ge t_0$. Thus for $s \ge 0$, we have
  \begin{equation*}
    f^*(s) = \sup_{t \ge 0} \{st - f(t)\}
    \le \sup_{t \in [0, t_0]} \{st - f(t)\}
    \vee \sup_{t \ge t_0} \{st - f(t)\}
    \le s t_0 \vee \sup_{t \ge t_0} \{st - c t^k\}
    \le s t_0 \vee C s^{k_*}.
  \end{equation*}

  Now we show the second claim.  To see this, recall the standard conjugacy
  result~\cite{HiriartUrrutyLe93ab} that $t(s) \in \argmax \{st - f(t)\}$, so
  that $t(s) \ge 0$ always, and let $\hat{t} = (s / k
  c)^{\frac{1}{k-1}}$. Assume that $s$ is large enough that $f(t) \ge c t^k$
  for $t > \hat{t}$.  Then for $t > \hat{t}$, we have
  \begin{equation*}
    s t - f(t) \le st - c t^k
    < s \hat{t} - c \hat{t}^k,
  \end{equation*}
  as $\hat{t}$ uniquely maximizes $s t - c t^k$. Thus $t$ cannot
  belong to $\partial f^*(s)$, giving the result.
\end{proof}

With Lemma~\ref{lemma:f-dual-growth} in hand, the next lemma follows.
\begin{lemma}
  \label{lemma:envelope-functions}
  There exists a constant $C < \infty$ and a neighborhood $U$ of
  $(\theta\opt, \lambda\opt, \eta\opt)$ such that
  $M(\statval) \defeq \sup_{(\theta,\lambda,\eta) \in U}
  \norm{\nabla H(\theta, \lambda, \eta; \statval)}$ satisfies
  \begin{equation*}
    M(\statval) \le C\left[
      \frac{|\loss(\theta\opt; \statval)|^{k_*}
      + |\eta\opt|^{k_*}}{{\lambda\opt}^{k_*}}
      + L(\statval)^{k_*}\right],
  \end{equation*}
  and $M_\epsilon(\statval) \defeq M(\statval) \cdot \epsilon$
  is an envelope for $\mc{H}_\epsilon$.
\end{lemma}
\begin{proof}
  The result is a standard algebraic exercise, coupled with the fact that a
  convex function $h$ is Lipschitz in an $\epsilon$-neighborhood of a point
  $t_0$ with constant $\sup_t \{\ltwo{\partial h(t)} \mid \norm{t - t_0}\}$
  (cf.~\cite{HiriartUrrutyLe93ab}). Thus, we bound the components of $\nabla
  H$ from Eq.~\eqref{eqn:H-derivative}; we only bound
  $\nabla_\theta H$ as the others are completely similar.
  For $(\theta, \lambda,
  \eta)$ in a neighborhood $U$ of $(\theta\opt, \lambda\opt, \eta\opt)$,
  we have for constants $C < \infty$ that may change from line to line
  \begin{align*}
    \norm{\nabla_\theta H(\theta, \lambda, \eta; \statval)}
    & = {f^*}'\left(\frac{\loss(\theta; \statval) - \eta}{\lambda}\right)
    \norm{\nabla \loss(\theta; \statval)} \\
    & \stackrel{(i)}{\le} C \left|\frac{\loss(\theta; \statval) - \eta}{\lambda}
    \right|^\frac{1}{k-1} \norm{\nabla \loss(\theta; \statval)}  \\
    & \stackrel{(ii)}{\le}
    C \left|\frac{\loss(\theta; \statval) - \eta}{\lambda}\right|^{
      \frac{k}{k-1}} + C \norm{\nabla \loss(\theta; \statval)}^{k_*} \\
    & \stackrel{(iii)}{\le} C \frac{|\eta|^{k_*}}{\lambda^{k_*}}
    + C \frac{|\loss(\theta\opt; \statval)|^{k_*}}{\lambda^{k_*}}
    + C L(\statval)^{k_*},
  \end{align*}
  where inequality~$(i)$ follows from
  Lemma~\ref{lemma:f-dual-growth},
  $(ii)$ follows by the Fenchel-Young inequality that $ab
  \le (1/k) |a|^{k} + (1/k_*) |b|^{k_*}$, while inequality~$(iii)$ is a
  consequence of
  Assumption~\ref{assumption:smoothness}.\ref{item:lipschitz-loss}.
  The remainder of the derivation follows from straightforward algebra
  once we note that $\lambda / \lambda\opt$ is bounded for $\lambda$ near
  $\lambda\opt$.
\end{proof}
Finally, we show that each of the conditions of
Lemma~\ref{lemma:m-estimator} holds for our problem.  That
$\E[M_\epsilon(\statrv)^2] < \infty$ is immediate by
Assumption~\ref{assumption:smoothness} on the moments of $\loss$ and $\nabla
\loss$.  For condition (i), we have
Assumption~\ref{assumption:strong-identifiability}.  For the bracketing
integral condition~\eqref{eqn:bracketing-cond}, From a standard bound on
bracketing numbers for Lipschitz functions~\cite[Theorem
  2.7.11]{VanDerVaartWe96}, we have
\begin{equation*}
  \log N_{[\hspace{1pt}]}\left(\delta \norm{M_{\epsilon}},
    \mc{H}_{\epsilon}, L_2(P_0) \right)
  \le (d+2) \log \left( 1 + \frac{2}{\delta} \right)
\end{equation*}
for $\epsilon$ small enough, so that the bracketing integral is bounded.
Each of the quantities~\eqref{eqn:limits-check} follows by Lebesgue's
dominated convergence theorem. For
condition~\eqref{eqn:weak-2norm}, we have
$M_\epsilon(\statval)^2 \indic{M_\epsilon(\statval) > c} / \epsilon^2
= M(\statval)^2 \indic{M(\statval) > c / \epsilon} \to 0$ as $\epsilon \to 0$,
and it is dominated by $M(\statval)$. For
condition~\eqref{eqn:modulus-zero},
we have for $v\opt = (\theta\opt, \lambda\opt, \eta\opt)$ that
\begin{equation*}
  |H(v\opt + \delta u_1; \statval) - H(v\opt + \delta u_2; \statval)|
  \le \sup_{v~{\rm near}~v\opt}
  \norm{\nabla H(v; \statval)} \delta \norm{u_1 - u_2}
  \le M(\statval) \delta \norm{u_1 - u_2}
\end{equation*}
by Lemma~\ref{lemma:envelope-functions}. Thus the dominated convergence
theorem again implies the convergence~\eqref{eqn:modulus-zero}.  For the
covariance condition~\eqref{eqn:covariance}, we use the differentiability of
$H$ as in Eq.~\eqref{eqn:H-derivative} to see that with $v\opt$ as above,
$\frac{1}{\delta} (H(v\opt + \delta u_1; \statval) - H(v\opt + \delta u_2;
\statval)) \to \<\nabla H(v\opt; \statval), u_1 - u_2\>$ and it is
dominated by $M(x) \norm{u_1 - u_2}$. Thus, we may take
\begin{equation*}
  G(u) \defeq \<W, u\>
  ~~ \mbox{for}~~ W \sim \normal\left(0, \cov(\nabla H(\theta\opt,
  \lambda\opt, \eta\opt; \statrv))\right)
\end{equation*}
as our Gaussian process.  The theorem is then an immediate consequence of
Lemma~\ref{lemma:m-estimator}.
